\documentclass[10pt,twocolumn,letterpaper]{article}

\usepackage[pagenumbers]{cvpr} 

\usepackage{pifont}
\usepackage{multirow}
\newcommand*{\ourmodel}{UniK3D\@\xspace}
\newcommand{\cmark}{\ding{51}}
\newcommand{\xmark}{\ding{55}}
\newcommand{\best}[1]{\mathbf{#1}}
\newcommand{\scnd}[1]{\underline{#1}}

\definecolor{cvprblue}{rgb}{0.21,0.49,0.74}
\usepackage[pagebackref,breaklinks,colorlinks,allcolors=cvprblue]{hyperref}

\usepackage{microtype}
\def\paperID{1346} 
\def\confName{CVPR}
\def\confYear{2025}

\title{UniK3D: Universal Camera Monocular 3D Estimation}

\author{
Luigi Piccinelli\textsuperscript{1} \quad Christos Sakaridis\textsuperscript{1} \quad Mattia Segu\textsuperscript{1} \\[0.2cm] Yung-Hsu Yang\textsuperscript{1} \quad Siyuan Li\textsuperscript{1} \quad  Wim Abbeloos\textsuperscript{2} \quad Luc Van Gool\textsuperscript{1,3}\\[0.4cm]
$^1$ETH Z\"urich \quad $^2$Toyota Motor Europe \quad $^3$INSAIT, Sofia University St. Kliment Ohridski
}

\begin{document}

\maketitle
\begin{abstract}
Monocular 3D estimation is crucial for visual perception.
However, current methods fall short by relying on oversimplified assumptions, such as pinhole camera models or rectified images.
These limitations severely restrict their general applicability, causing poor performance in real-world scenarios with fisheye or panoramic images and resulting in substantial context loss.
To address this, we present \emph{\ourmodel}\footnote{Pronounced ``Unique-3D'', with \textbf{K} denoting the intrinsics matrix.}, 
the first
generalizable method for monocular 3D estimation able to model any camera. 
Our method introduces a spherical 3D representation which allows for better disentanglement of camera and scene geometry and enables accurate metric 3D reconstruction for unconstrained camera models.
Our camera component features a novel, model-independent representation of the pencil of rays, achieved through a learned superposition of spherical harmonics.
We also introduce an angular loss, which, together with the camera module design, prevents the contraction of the 3D outputs for wide-view cameras.
A comprehensive zero-shot evaluation on 13 diverse datasets demonstrates the state-of-the-art performance of \ourmodel across 3D, depth, and camera metrics, with substantial gains in challenging large-field-of-view and panoramic settings, while maintaining top accuracy in conventional pinhole small-field-of-view domains. Code and models are available at \href{https://github.com/lpiccinelli-eth/unik3d}{github.com/lpiccinelli-eth/unik3d}.
\end{abstract}    
\section{Introduction}
\label{sec:intro}

\begin{figure}[ht]
    \centering
    \footnotesize
    \includegraphics[width=1.0\linewidth]{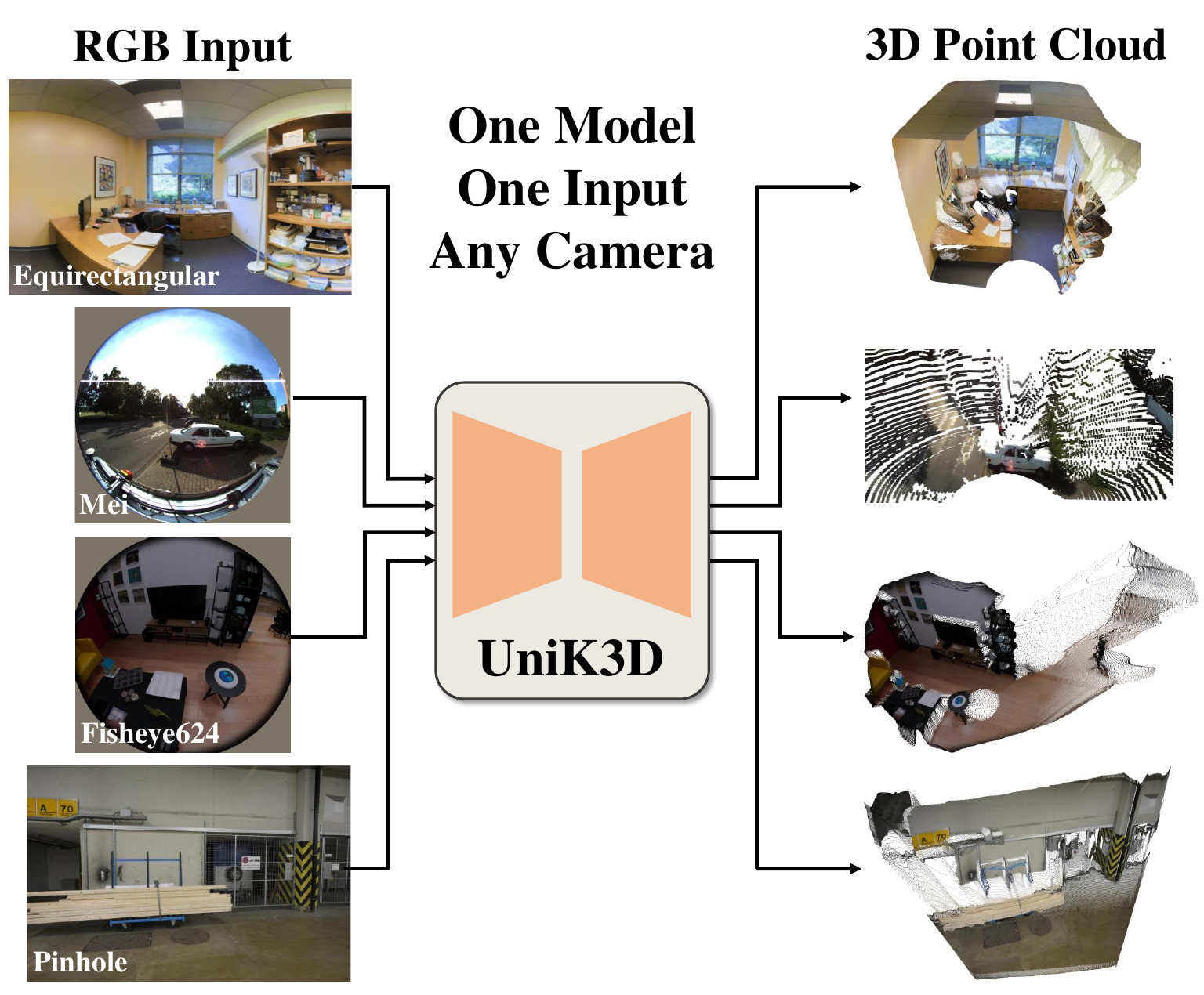}
    \vspace{-10pt}
    \caption{\textbf{\ourmodel} introduces a novel and versatile approach that delivers precise metric 3D geometry estimation from a single image and for any camera type, ranging from pinhole to panoramic, without requiring any camera information.
    By leveraging (i) a flexible and general spherical formulation both for the radial dimension of 3D space and for the two camera-model-dependent orientation dimensions and (ii) advanced conditioning strategies. \ourmodel outperforms traditional models
    without needing camera calibration or domain-specific tuning.
    }
    \label{fig:method:teaser}
\end{figure}

Estimating 3D scene geometry is a fundamental task in computer vision since such 3D information serves as a crucial cue for action planning and execution~\cite{Zhou2019, dong2022depth4robotics}.
The scene's geometry 3D estimation task is vital for a wide range of applications, including autonomous navigation~\cite{wang2019depth4vehicles, park2021dd3d} and 3D modeling~\cite{deng2022nerf}, where accurate spatial understanding is essential.
Recent advances in generalizable monocular depth estimation (MDE)~\cite{ranftl2020midas, yang2024da1, ke2024marigold} deliver impressive performance and visual quality across various domains, but these models are constrained to a relative output scale.
Nonetheless, for practical applications, a consistent and reliable \emph{metric-scaled} monocular depth estimate (MMDE) is crucial, as it enables accurate 3D reconstruction and geometric scene understanding necessary for embodied agents.

Existing methods have made considerable strides in the above direction of metric estimation.
Earlier approaches assumed known camera intrinsics at test time~\cite{guizilini2023zerodepth, yin2023metric3d}, while more recent works have relaxed this assumption~\cite{piccinelli2024unidepth, piccinelli2025unidepthv2, bochkovskii2024depthpro}.
However, these approaches still impose restrictive assumptions about input cameras, such as relying on a basic pinhole camera model~\cite{piccinelli2024unidepth, bochkovskii2024depthpro} or requiring access to ground-truth rectification parameters~\cite{yin2023metric3d}.
These simplifications substantially hinder the applicability and degrade the performance of the above methods in real-world settings, where a wide range of camera projection models with strong non-linear deformations are common, such as fisheye or panoramic lenses.
This limitation is more pronounced when estimating complete metric 3D geometry instead of only depth maps, as the former depends more heavily on the quality of camera estimation.
Due to the restrictive assumptions in existing models, general camera estimation can not be effectively learned, even when models are exposed to images from varied camera types.
Furthermore, the output space of previous state-of-the-art MMDE methods has inherent limitations, \eg both disparity and log-depth prediction become mathematically ill-posed when the field of view (FoV) exceeds 180 degrees.

To address these challenges, we introduce \emph{\ourmodel}, the first framework for monocular metric 3D scene's geometry estimation that generalizes across a wide variety of camera models, from pinhole to fisheye and panoramic configurations, as shown in \cref{fig:method:teaser}.
Our method proposes a novel formulation for monocular 3D estimation which is spherical in two senses.
First, \ourmodel leverages a \emph{fully spherical} output 3D space, modeling the range dimension through \emph{radial distance} instead of perpendicular depth.
This approach is especially beneficial at large angles from the optical axis, effectively resolving the ill-posed nature of traditional methods at extreme fields of view.
Second, while building on the recently proposed decomposition~\cite{piccinelli2024unidepth} of camera prediction from depth estimation, \ourmodel newly presents a general \emph{spherical harmonics basis} as the \emph{direct} output space of the camera module that represents the pencil of rays.
Unlike previous works~\cite{piccinelli2024unidepth, bochkovskii2024depthpro} which predict explicit pinhole camera parameters and then encode~\cite{piccinelli2024unidepth} induced rays using a spherical basis, we remove the camera assumption and directly model the rays.
As a result, \ourmodel spans an unrestricted space of possible camera models, allowing for flexible and accurate depth prediction regardless of camera intrinsics.
Our assumption-free spherical camera representation, with its flexibility, ensures that our model is well-suited for real-world deployment, where capturing scenes with non-standard cameras is common.

Our key contribution is the first camera-universal model for monocular 3D estimation that can accommodate any camera projective geometry.
We achieve this through our unified spherical output representation that supports all inverse projection problems.
By employing a fully spherical framework, our method ensures a complete disentanglement of projective \vs 3D scene geometry, as the dimension of an object projection on the image is a univocal function only \wrt radial distance and not \wrt depth.
This disentanglement allows more consistent 3D reconstructions and enhances the stability of 3D outputs near the $xy$-plane, where depth approaches zero.
Moreover, \ourmodel models the camera rays as a decomposition across a finite spherical harmonics basis.
This choice ensures representation generality and versatility, and at the same time maintains an accurate and compact representation for the resulting pencils of rays, also introducing inductive biases such as continuity and differentiability.
In addition, we propose multiple novel strategies to ensure robust camera conditioning of our \emph{radial module}
such as an asymmetric angular loss based on quantile regression, static encoding, and curriculum learning.

We validate our approach through extensive zero-shot experiments on 13 widely used metric depth datasets, where \ourmodel not only achieves state-of-the-art performance in monocular metric depth and 3D estimation, but also generalizes very well across various camera models, without either preprocessing or specific camera domains during training.

\section{Related Work}
\label{sec:relwork}

\begin{figure*}[ht]
    \centering
    \footnotesize
    \includegraphics[width=0.95\linewidth]{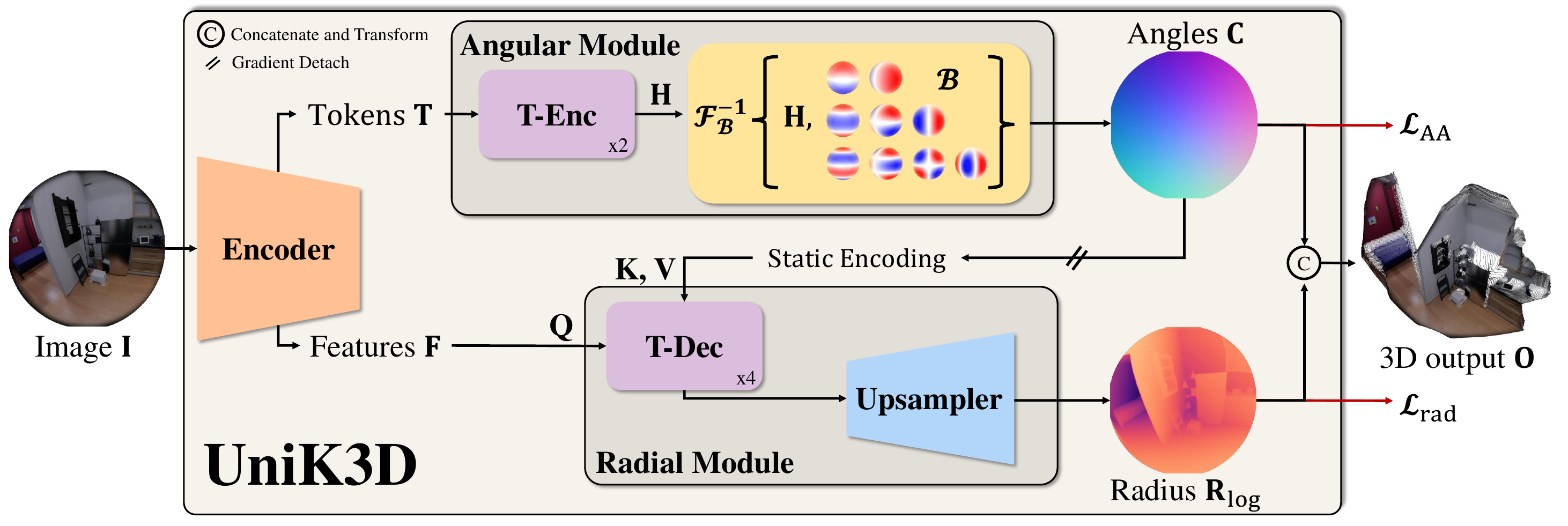}
    \vspace{-10pt}
    \caption{\textbf{Model architecture.}
    \ourmodel utilizes solely the single input image to generate the 3D output point cloud ($\mathbf{O}$) for any camera.
    The projective geometry of the camera is predicted by the Angular Module.
    The camera representation corresponds to azimuth and polar angles ($
    \mathbf{C}$) of the backprojected pencil of rays on the unit sphere $\mathbb{S}^3$.
    The class tokens from the Encoder are processed by 2 Transformer Encoder (T-Enc) layers to obtain the 15 coefficients ($\mathbf{H}$) of the inverse Spherical transform $\mathcal{F}^{-1}_{\mathcal{B}}\{\mathbf{H}\}$ defined by a finite basis ($\mathcal{B}$) of spherical harmonics up to degree 3 with no constant component.
    Stop-gradient is applied to the angular information which conditions the Radial Module, simulating external information flow.
    The ``static encoding'' refers to sinusoidal encoding which matches the internal feature dimensionality.
    The Radial Module is composed of Transformer Decoder (T-Dec) blocks, one for each input resolution, which is utilized to condition the Encoder features on the bootstrapped camera representation.
    This conditioning injects prior knowledge on scene scale and projective geometry.
    The radial output ($\mathbf{R}_{\log}$) is obtained by processing the camera-aware features via a learnable upsampling module.
    The final output is the concatenation of the camera and radial tensors ($\mathbf{C} || \mathbf{R}_{\log}$). A closed-form coordinate transform is applied to obtain the Cartesian 3D output, but supervision is applied directly on angular coordinates, with our asymmetric angular loss $\mathcal{L}_{\text{AA}}$, and radial coordinates.
    }
    \label{fig:method:overview}
    \vspace{-5pt}
\end{figure*}

\noindent{}\textbf{Monocular Depth Estimation.}
The introduction of end-to-end neural networks for MDE, first demonstrated by \cite{Eigen2014}, revolutionized the field by enabling depth prediction through direct optimization, utilizing the Scale-Invariant log loss ($\mathrm{SI}_{\log}$).
Since then, the field has evolved with increasingly sophisticated models, ranging from convolutional architectures~\cite{Fu2018Dorn, Laina2016, Liu2015, Patil2022p3depth} to recent advancements using transformers~\cite{Yang2021, Bhat2020adabins, Yuan2022newcrf, piccinelli2023idisc}.
While these approaches have pushed the boundaries of MDE performance in controlled benchmarks, they often fail when faced with zero-shot scenarios, highlighting a persistent challenge: ensuring robust generalization across varying camera and scene domains and diverse geometric and visual conditions.

\noindent{}\textbf{Generalizable Monocular Depth Estimation.}
To address the limitations of domain-specific models, recent research has focused on developing generalizable and zero-shot MDE techniques.
These methods can be categorized into scale-agnostic approaches~\cite{ranftl2020midas, yang2024da1, yang2024da2, ke2024marigold, wang2024moge}, which aim to mitigate scale ambiguity and emphasize perceptual depth quality, and metric depth models~\cite{bhat2023zoedepth, guizilini2023zerodepth, yin2023metric3d, hu2024metric3dv2, piccinelli2024unidepth, piccinelli2025unidepthv2, bochkovskii2024depthpro}, which prioritize accurate geometric reconstruction.
However, most existing MDE methods fall short of achieving truly zero-shot monocular metric 3D scene estimation.
In particular, scale-agnostic approaches often require additional information to resolve scale ambiguities, while most of the metric-based models depend on a known camera or assume a simplistic pinhole camera configuration.
Even the few models which are designed for zero-shot 3D scene estimation~\cite{piccinelli2024unidepth, bochkovskii2024depthpro, yin2023metric3d} remain constrained: they either explicitly assume a pinhole camera model~\cite{piccinelli2024unidepth, bochkovskii2024depthpro} or necessitate image rectification~\cite{yin2023metric3d}, effectively requiring test-time camera information and limiting their zero-shot generalizability to pinhole cameras.

On the contrary, \ourmodel addresses these limitations by offering a unified solution that can handle any inverse projection problem.
Our model can recover a coherent 3D point cloud from any single image, regardless of camera intrinsics, without any rectification or camera information at test time.
This generality sets \ourmodel apart, enabling robust and universal monocular metric 3D estimation that is required in diverse and challenging real-world applications.

\noindent{}\textbf{Camera Calibration.}
Camera calibration is essential for estimating intrinsic parameters like focal length, principal point, and distortion coefficients to model the mapping from 3D world points to 2D image coordinates.
Traditional parametric models, such as the pinhole model, Kannala-Brandt~\cite{Kannala2006KB}, Mei~\cite{Mei2007mei}, Omnidirection~\cite{Scaramuzza2014Omnidirectional}, Unified Camera Model (UCM)~\cite{Geyer2000ucm}, Enhanced UCM~\cite{Khomutenko2016EUCM}, and Double Sphere~\cite{Usenko2018DS} models are effective for narrow- and wide-angle lenses but require controlled environments for accurate calibration.
As models grow more complex, the risk of errors or divergence increases, especially under varying lighting or sensor noise. Additionally, each model has inherent limitations, \eg UCM cannot represent tangential distortion, and Kannala-Brandt struggles beyond a 210\textsuperscript{$\circ$} FoV.

By contrast, we take a different approach and model the camera backprojection as a linear combination of \emph{spherical basis} functions, \ie via an inverse spherical harmonics transformation, where the model simply infers the scalar expansion coefficients and the spherical domain boundaries.

\section{UniK3D}
\label{sec:method}

Generalizable depth or 3D scene estimation models often face significant challenges when adapting to diverse camera configurations.
Existing methods typically rely on rigid and camera-specific assumptions, such as the pinhole model or equirectangular models, or require extensive preprocessing steps like rectification.
These constraints limit their applicability to real-world scenarios with non-standard camera projective geometries.
By contrast, our model, \ourmodel, introduces a novel framework that enables monocular 3D geometry estimation for any scene and any camera setup.

We begin by introducing the design of our 3D output space and the internal representation of the camera in \cref{ssec:method:repr}.
Our representation is intentionally formulated to be as general as possible, allowing to handle all inverse projection problems.
Through our preliminary studies, we observed a consistent issue: the network predictions contracted to a reduced FoV, even when trained on a diverse set of camera types including large FoVs.
Simple data re-balancing strategies proved insufficient to address this phenomenon.
To overcome this, we have developed a series of architectural and design interventions, detailed in \cref{ssec:method:engineer}, aimed at preventing the backprojection contraction.
In \cref{ssec:method:design}, we describe the architecture of our model, our optimization strategy, and the specific design and loss functions underpinning our approach.
\cref{fig:method:overview} displays an overview of our method.

\subsection{Representations}
\label{ssec:method:repr}


\noindent\textbf{Output Space.} The output representation of \ourmodel is designed to be universally compatible with any scene and camera configuration, providing a direct metric 3D scene estimate for each input image.
Drawing from the disentanglement strategy presented in~\cite{piccinelli2024unidepth}, our approach separates camera parameters from scene geometry.
Specifically, we represent the camera using a dense tensor $\mathbf{C} = \mathbf{\theta} || \mathbf{\phi}$, where $\mathbf{\theta}$ is the polar angle and $\mathbf{\phi}$ is the azimuthal angle, consistent with standard spherical coordinates.
However, we use the Euclidean \emph{radius} (distance from the camera center) as the scene range component within a \emph{fully spherical} framework, instead of relying on traditional perpendicular-depth-based representations.
This design choice ensures that dimensions of projected objects in the image vary \emph{univocally} with radius, a property that does not characterize the depth representation and renders the latter much harder to learn.
Furthermore, the spherical framework enhances numerical stability when handling points near the $xy$-plane, a region where previous methods typically face challenges due to large gradients.
We convert the spherical representation to Cartesian coordinates using a bijective transformation, accurately capturing the 3D geometry of the scene as the output 3D point cloud $\mathbf{O}$.

\noindent\textbf{Camera Internal Space.} In \ourmodel, the dense pencil of rays which represents the viewing directions for the various pixels is expressed through a basis decomposition, providing a flexible and comprehensive angular representation.
As shown in \cref{fig:method:overview}, our Angular Module predicts a tensor of coefficients $\mathbf{H}$, which is derived from the encoder’s class tokens, denoted as $\mathbf{T}$.
These coefficients correspond to a predefined basis: the Spherical Harmonics (SH) basis. We reconstruct the pencil of rays from $\mathbf{H}$ as follows:
\begin{equation}
    \label{eq:sh:inverse}
    \mathbf{C} = \mathcal{F}^{-1}_{\mathcal{B}}\{\mathbf{H}\} = \sum_{l=0}^{L} \sum_{m=-l}^{l} \mathbf{H}_{lm} \mathcal{B}_{lm}(\theta, \phi),
\end{equation}
where $\mathbf{C}$ represents the reconstructed angular field and $\mathcal{F}^{-1}_{\mathcal{B}}$ denotes the inverse transform from the coefficient space to the angular space, using the SH basis $\mathcal{B}$.
$\mathbf{B}_{lm}(\theta, \phi)$ are the SH basis functions, \ie Legendre polynomials, and $\mathbf{H}_{lm}$ are the predicted coefficients.
Here, $l$ and $m$ index the degree and order of the harmonics, respectively.
This inverse transform is implemented as an inner product that maps from $\mathbb{R}^n \times \mathbb{S}^3$ to $\mathbb{S}^3$.
The SH basis domain is defined by 4 parameters: the generalized ``principal point'' of the reference frame, \ie the pole, and the horizontal and vertical FoVs.
This formulation allows us to describe complex ray distributions \emph{compactly} and implicitly, while ensuring important properties of the output, such as continuity and differentiability.

Additionally, the SH basis offers high sparsity, requiring only 15 harmonics for a 3\textsuperscript{rd} degree basis without constant component and an equal number of coefficients to accurately represent intrinsics for most camera types.
By leveraging this SH-based representation and defining the domain through the pole and FoV parameters, \ourmodel achieves a robust and flexible framework that can handle virtually any camera geometry with only 19 parameters.

\begin{figure*}[h!]
    \renewcommand{\arraystretch}{1}
    \centering
    \small
    \begin{tabular}{cc|cccc|c}
        \multirow{1}{*}[1.9cm]{\rotatebox[origin=c]{90}{\parbox{2cm}{\centering Stanford-2D3D\\(Panoramic)}}}
        & \includegraphics[width=0.14\linewidth]{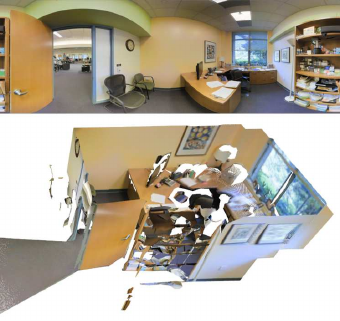}
        & \includegraphics[width=0.14\linewidth]{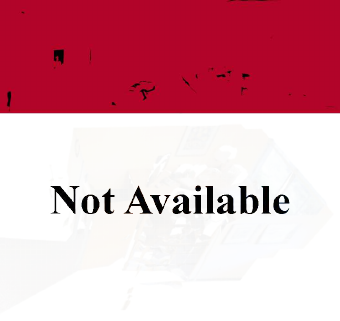}
        & \includegraphics[width=0.14\linewidth]{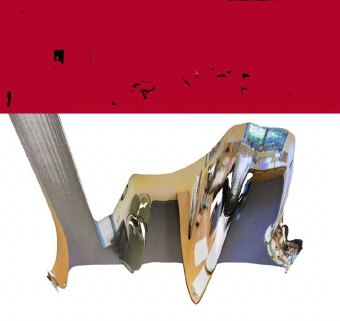}
        & \includegraphics[width=0.14\linewidth]{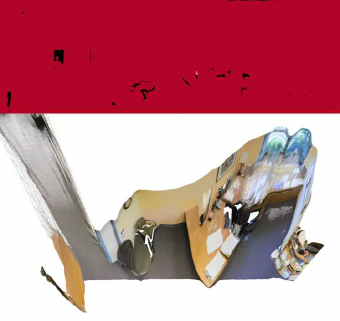}
        & \includegraphics[width=0.14\linewidth]{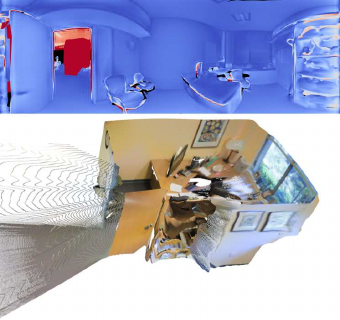}
        & \includegraphics[width=0.04\linewidth]{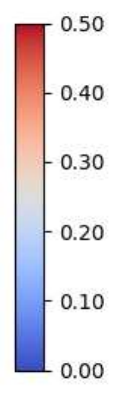} \\
        
        \multirow{1}{*}[2.2cm]{\rotatebox[origin=c]{90}{\parbox{2cm}{\centering KITTI360\\(Mei $180^{\circ}$)} }}
        & \includegraphics[width=0.14\linewidth]{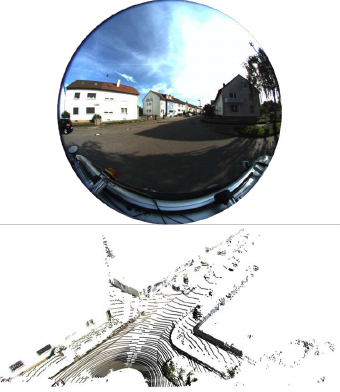}
        & \includegraphics[width=0.14\linewidth]{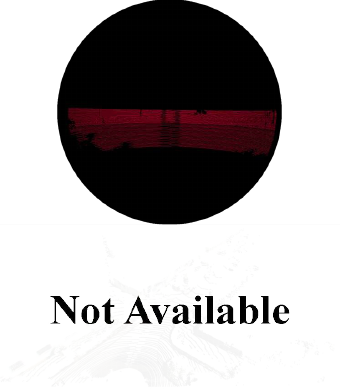}
        & \includegraphics[width=0.14\linewidth]{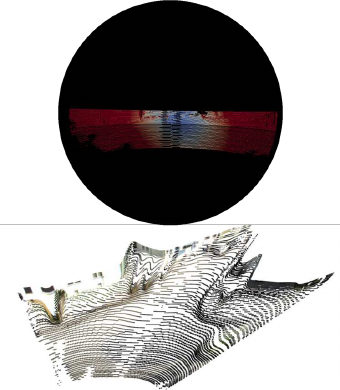}
        & \includegraphics[width=0.14\linewidth]{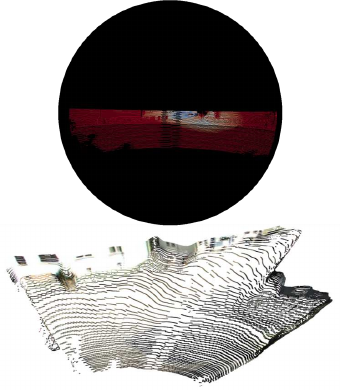}
        & \includegraphics[width=0.14\linewidth]{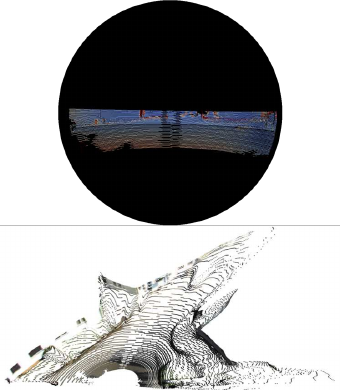}
        & \includegraphics[width=0.04\linewidth]{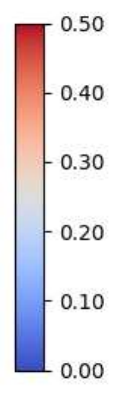} \\
        
        \multirow{1}{*}[2.9cm]{\rotatebox[origin=c]{90}{\parbox{3cm}{\centering Aria Digital Twin\\(Fisheye624)} }}
        & \includegraphics[width=0.14\linewidth]{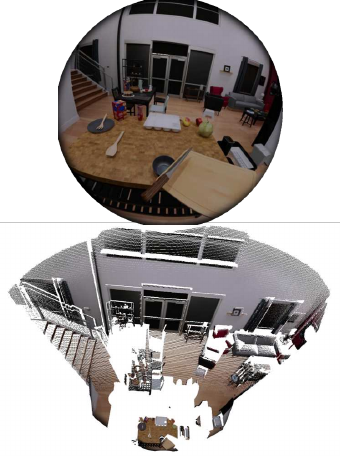}
        & \includegraphics[width=0.14\linewidth]{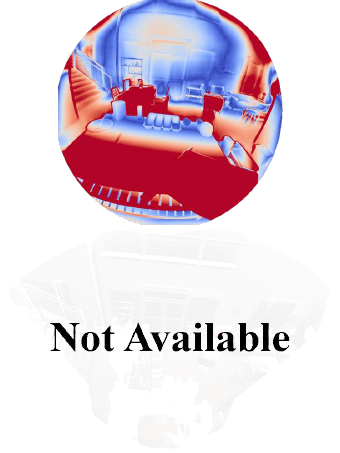}
        & \includegraphics[width=0.14\linewidth]{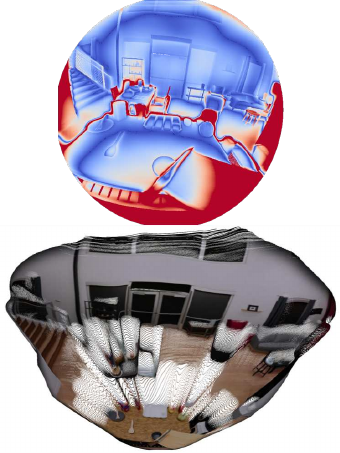}
        & \includegraphics[width=0.14\linewidth]{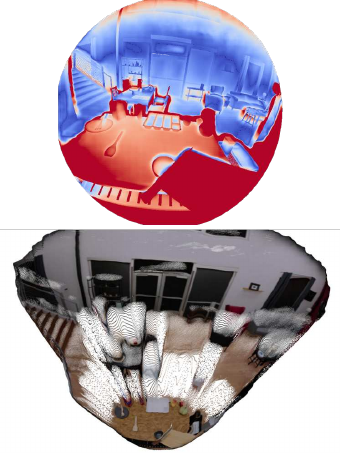}
        & \includegraphics[width=0.14\linewidth]{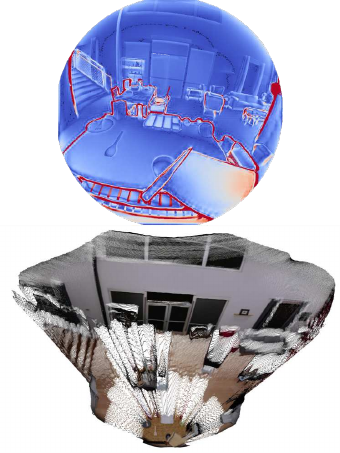}
        & \includegraphics[width=0.04\linewidth]{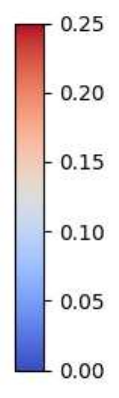} \\

        \multirow{1}{*}[2.4cm]{\rotatebox[origin=c]{90}{\parbox{2cm}{\centering ETH3D\\(Pinhole)}}}
        & \includegraphics[width=0.14\linewidth]{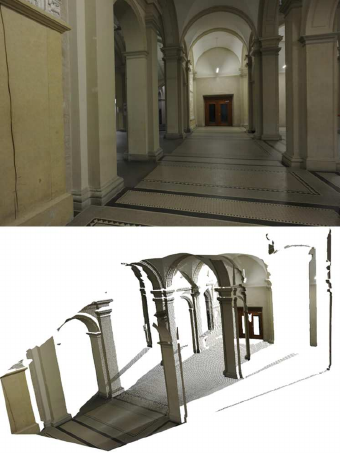}
        & \includegraphics[width=0.14\linewidth]{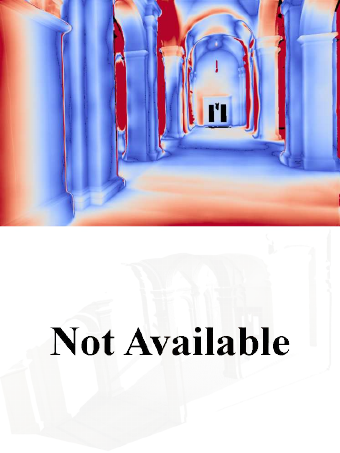}
        & \includegraphics[width=0.14\linewidth]{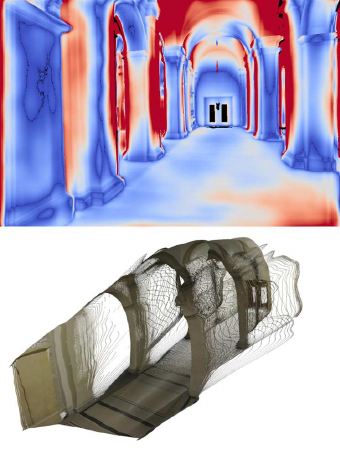}
        & \includegraphics[width=0.14\linewidth]{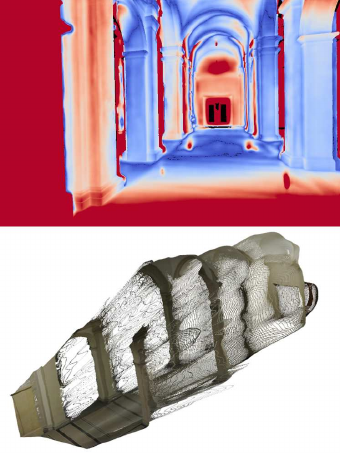}
        & \includegraphics[width=0.14\linewidth]{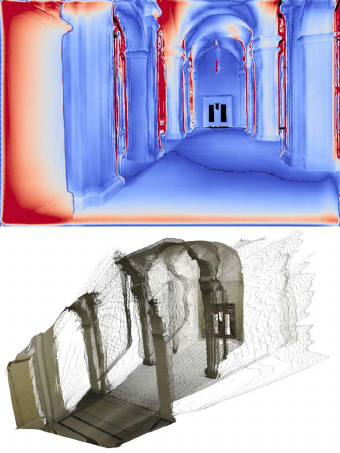}
        & \includegraphics[width=0.04\linewidth]{figures/vis/ADT/adt_cmap.pdf} \\

        & RGB \& GT & DepthAnything~\cite{yang2024da1} & UniDepth~\cite{piccinelli2024unidepth} & MASt3R~\cite{leroy2024master} & \ourmodel & $\mathrm{A.Rel}$ \\
    \end{tabular}
    \vspace{-5pt}
    \caption{\textbf{Qualitative comparisons.} Each pair of consecutive rows represents one test sample. Each odd row displays the input RGB image and the 2D error map, color-coded with the \textit{coolwarm} colormap based on absolute relative error (for panoramic images, the error is computed on distance rather than depth).
    To ensure a fair comparison, errors are calculated on GT-based shifted and scaled outputs for all models.
    Each even row shows the ground truth and predictions of the 3D point cloud.
    The last column displays the specific colormap ranges for absolute relative error.
    Key observations for each rows pair: (1) competing methods are limited to only positive depth and heavily distort the scenes for larger FoV; (2) in the case of representable but large FoV (180\textsuperscript{$\circ$}), \ourmodel output is the only one that does not suffer from pronounced FoV contraction; (3) for moderate-FoV images but with strong boundary distortion, \eg fisheye, \ourmodel can maintain planarity and overall scene structure; (4) our approach also delivers accurate 3D estimates for standard pinhole images.}
    \vspace{-5pt}
    \label{fig:results:main_vis}
\end{figure*}


\subsection{Preventing Distribution Contraction} 
\label{ssec:method:engineer}


\noindent\textbf{Asymmetric Angular Loss.} Neural networks tend to regress towards the most frequent modes in the training data, often neglecting the distribution tails.
In our case, this bias would cause \ourmodel to underrepresent wide-FoV angles in its outputs, since most visual datasets are heavily skewed towards small-FoV pinhole cameras.
This leads to poor performance in scenarios requiring accurate wide-angle predictions.
To overcome this issue, we introduce an asymmetric angular loss based on quantile regression, inspired by robust statistical estimators and decision theory principles, \ie type-I and type-II errors~\cite{neyman1933tests}. Our loss function is defined as:
\begin{equation}
    \mathcal{L}^{\alpha}_{\text{AA}}(\hat{\theta}, \theta^{*}) = \alpha \sum_{\hat{\theta} > \theta^{*}} \left|\hat{\theta} - \theta^{*}\right| + (1 - \alpha) \sum_{\hat{\theta} \leq \theta^{*}} \left|\hat{\theta} - \theta^{*}\right|,
\end{equation}
where $0 \leq \alpha \leq 1$ is the target quantile, $\hat{\theta}$ is the predicted angle, and $\theta^{*}$ is the ground-truth angle.
This formulation adjusts the weighting of over- and underestimations of the polar angle $\theta$.
When $\alpha = 0.5$, the loss degenerates to the standard Mean Absolute Error (MAE), but by tuning $\alpha$, we can emphasize underrepresented angles and balance the regression more effectively.
Unlike naive dataset rebalancing--which would alter the underlying 3D scene diversity and introduce significant complexity, especially across multiple datasets--our loss addresses the angular imbalance directly and efficiently.
By using quantile regression, we minimize the complexity to a simple search over the interval $[0, 1]$ for $\alpha$, making our method well-suited for large-scale and diverse training scenarios.
This quantile-based strategy allows us to address the angular distribution bias without sacrificing simplicity and diversity, making it a robust and scalable solution.

\noindent\textbf{Enhancing Camera Conditioning.} In our initial experiments, we observed that our model struggled to effectively utilize camera conditioning following previous works~\cite{piccinelli2024unidepth}, even when explicitly supplied with ground-truth camera rays during both training and testing.
This issue was subtle for small-FoV pinhole cameras, but it became significant for large-FoV configurations.
The root of the problem lies in weak conditioning: the model fails to disentangle camera parameters from geometric features, causing it to route local aberrations back to the encoder features' space, without integrating essential FoV information.
As a result, even when prompted with accurate camera parameters at test time, the model might ignore, or be misled by, this information.

To address this, we hypothesize that camera data must be clear and explicitly structured from the beginning of training.
To this end, we implement in \ourmodel a static (non-learnable) encoding of camera rays and adopt a curriculum learning strategy, transitioning gradually from feeding GT camera parameters to predicted ones to the Radial Module.
In particular, the GT camera is fed to the Radial Module with probability $1 - \mathrm{tanh}(\frac{s}{10^{5}})$, where $s$ is the current optimization step.
To reinforce external conditioning, we detach gradients from the camera output that is fed to the Radial Module, hence preventing the model from relying on feedback mechanisms that could undermine the conditioning on the camera.
Additionally, we disable learnable gains, such as LayerScale~\cite{touvron2021layerscale}, in the cross-attention layers of the Radial Module’s transformer decoder, to avoid shortcuts
of the conditioning.
These strategies ensure that the model effectively leverages camera information to adjust its encoder features, enhancing the robustness of 3D predictions.

\subsection{Network Design}
\label{ssec:method:design}

\begin{table*}
\centering
\caption{\textbf{Comparison on zero-shot evaluation for diverse camera domains.} Validation sets: \textit{S.FoV} includes NYU, KITTI, IBims-1, ETH-3D, nuScenes, and Diode Indoor; \textit{S.FoV\textsubscript{Dist}} includes IBims-1, ETH-3D, and Diode Indoor with synthetic distortion; \textit{L.FoV} includes ADT, ScanNet++ (DSLR), and KITTI360; \textit{Pano} uses Stanford-2D3D. All models use a ViT-L backbone. Missing values (\textcolor{gray}{-}) indicate the model's inability to produce the respective output. Metric3D and Metric3Dv2 cannot be evaluated on panoramic images as focal lengths are undefined. \dag: Requires ground-truth (GT) camera for 3D reconstruction. \ddag: Requires GT camera for 2D depth map inference.}
\vspace{-10pt}
\label{tab:results:aggregated_comparisons}
\resizebox{\linewidth}{!}{%
\begin{tabular}{l|ccc|ccc|ccc|ccc}
\toprule
\multirow{2}{*}{\textbf{Method}} & \multicolumn{3}{c|}{S.FoV} & \multicolumn{3}{c|}{S.FoV\textsubscript{Dist}} & \multicolumn{3}{c|}{L.FoV} & \multicolumn{3}{c}{Pano} \\
 & $\mathrm{\delta_1^{SSI}}\uparrow$ & $\mathrm{F_A}\uparrow$ & $\mathrm{\rho_A}\uparrow$ & $\mathrm{\delta_1^{SSI}}\uparrow$ & $\mathrm{F_A}\uparrow$ & $\mathrm{\rho_A}\uparrow$ & $\mathrm{\delta_1^{SSI}}\uparrow$ & $\mathrm{F_A}\uparrow$ & $\mathrm{\rho_A}\uparrow$ & $\mathrm{\delta_1^{SSI}}\uparrow$ & $\mathrm{F_A}\uparrow$ & $\mathrm{\rho_A}\uparrow$ \\
\midrule

DepthAnything~\cite{yang2024da1} & $92.2$ & \textcolor{gray}{-} & \textcolor{gray}{-} & $94.3$ & \textcolor{gray}{-} & \textcolor{gray}{-} & $47.5$ & \textcolor{gray}{-} & \textcolor{gray}{-} & $10.4$ & \textcolor{gray}{-} & \textcolor{gray}{-} \\
DepthAnythingv2~\cite{yang2024da2} & $92.4$ & \textcolor{gray}{-} & \textcolor{gray}{-} & $88.9$ & \textcolor{gray}{-} & \textcolor{gray}{-} & $48.7$ & \textcolor{gray}{-} & \textcolor{gray}{-} & $11.3$ & \textcolor{gray}{-} & \textcolor{gray}{-} \\
Metric3D\textsuperscript{\dag \ddag}~\cite{yin2023metric3d} & $86.4$ & $43.1$ & \textcolor{gray}{-} & $88.0$ & $36.7$ & \textcolor{gray}{-} & $58.7$ & $26.0$ & \textcolor{gray}{-} & \textcolor{gray}{-} & \textcolor{gray}{-} & \textcolor{gray}{-} \\
Metric3Dv2\textsuperscript{\dag \ddag}~\cite{hu2024metric3dv2} & $91.1$ & $59.7$ & \textcolor{gray}{-} & $89.4$ & $47.1$ & \textcolor{gray}{-} & $69.2$ & $24.7$ & \textcolor{gray}{-} & \textcolor{gray}{-} & \textcolor{gray}{-} & \textcolor{gray}{-} \\
ZoeDepth\textsuperscript{\dag}~\cite{bhat2023zoedepth} & $88.9$ & $53.3$ & \textcolor{gray}{-} & $90.3$ & $40.1$ & \textcolor{gray}{-} & $65.3$ & $6.4$ & \textcolor{gray}{-} & $32.7$ & $9.9$ & \textcolor{gray}{-} \\
UniDepth~\cite{piccinelli2024unidepth} & $94.9$ & $59.0$ & $85.0$ & $94.0$ & $43.0$ & $70.5$ & $68.6$ & $16.9$ & $19.8$ & $33.0$ & $2.0$ & $1.7$ \\
MASt3R~\cite{leroy2024master} & $88.0$ & $37.8$ & $80.8$ & $89.9$ & $35.2$ & $\scnd{77.1}$ & $67.1$ & $29.7$ & $25.1$ & $32.3$ & $3.7$ & $2.1$ \\
DepthPro~\cite{bochkovskii2024depthpro} & $87.4$ & $56.0$ & $79.6$ & $80.6$ & $29.4$ & $71.7$ & $64.5$ & $26.1$ & $32.1$ & $31.8$ & $1.9$ & $1.9$ \\
\midrule
\ourmodel-Small & $94.3$ & $61.3$ & $85.7$ & $95.1$ & $48.4$ & $73.8$ & $84.5$ & $55.5$ & $70.1$ & $81.3$ & $72.5$ & $\scnd{53.7}$ \\
\ourmodel-Base & $\scnd{95.5}$ & $\scnd{64.9}$ & $\scnd{86.1}$ & $\scnd{96.5}$ & $\scnd{50.2}$ & $75.1$ & $\scnd{87.4}$ & $\scnd{67.7}$ & $\scnd{79.9}$ & $\best{83.6}$ & $\scnd{73.7}$ & $\scnd{53.7}$ \\
\ourmodel-Large & $\best{96.1}$ & $\best{68.1}$ & $\best{89.4}$ & $\best{97.3}$ & $\best{54.5}$ & $\best{78.8}$ & $\best{91.2}$ & $\best{71.6}$ & $\best{81.9}$ & $\scnd{81.4}$ & $\best{80.2}$ & $\best{57.1}$ \\
\bottomrule
\end{tabular}%
}
\vspace{-10pt}
\end{table*}
\begin{table}[ht]
\centering
\caption{\textbf{Zero-shot comparison with equirectangular-specialized methods.} All methods are zero-shot tested on Stanford-2D3D~\cite{armeni20172d3ds}. Competing methods are all trained on equirectangular images. Our training set includes Matterport3D~\cite{chang2017matterport3d} with 2\% sampling.}
\vspace{-10pt}
\label{tab:results:equi_comparisons}
\resizebox{0.75\linewidth}{!}{%
\begin{tabular}{lc|cc}
\toprule
\textbf{Method} & Train & $\mathrm{\delta_1} \uparrow$ & $\mathrm{A.Rel} \downarrow$ \\
\midrule
BiFuse\textsuperscript{\dag}~\cite{wang2020bifuse} & Matterport3D & $86.2$ & $12.0$ \\
BiFuse++\textsuperscript{\dag}~\cite{wang2022bifuse++} & Matterport3D & $\underline{91.4}$ & $10.7$ \\
UniFuse\textsuperscript{\dag}~\cite{jiang2021unifuse} & Matterport3D & $91.3$ & $\underline{9.42}$ \\
\midrule
\ourmodel & Ours & $\mathbf{96.8}$ & $\mathbf{8.01}$\\
\bottomrule
\end{tabular}%
}
\end{table}

\noindent{}\textbf{Architecture.} Our network consists of an Encoder Backbone, an Angular Module, and a Radial Module, as illustrated in \cref{fig:method:overview}.
Our encoder is ViT-based~\cite{Dosovitskiy2020VIT} and we extract dense features $\mathbf{F} \in \mathbb{R}^{h \times w \times C \times 4}$--where $(h, w) = (\frac{H}{14}, \frac{W}{14})$--along with class tokens $\mathbf{T}$.
The Angular Module processes these class tokens, projecting them onto 512-channel representations that are split into 3 domain parameters and 15 spherical coefficient prototypes.
These tokens pass through two layers of a Transformer Encoder (T-Enc) with 8 heads and are then projected onto scalar values.
The values for the 3 domain parameters define the principal point (2) and the horizontal FoV (1), determining the intervals for the harmonics. We assume square pixels and thus do not learn an extra, fourth parameter for the vertical FoV, but rather compute this fourth parameter directly from the horizontal FoV.
The 15 spherical coefficients undergo an inverse SH transformation according to \eqref{eq:sh:inverse}, using a 3-degree SH basis.
The gradient flowing from the Angular Module to the class tokens is multiplied by 0.1, as the magnitude of the camera-induced gradient for the encoder weights was empirically found to be ca.\ 10x higher than the radial-induced gradient.

The Radial Module first processes the dense encoder features $\mathbf{F}$ through a Transformer Decoder (T-Dec) with 4 parallel layers, one for each resolution, and 1 head.
These layers condition $\mathbf{F}$ on the sine-encoded angular representation $\mathbf{C}$ (cf.\ \cref{sec:supp:arch}).
The conditioned features are then projected onto a 512-channel tensor, forming radial features $\mathbf{D} \in \mathbb{R}^{h \times w \times 512}$.
These radial features are afterwards upsampled to the input resolution using residual convolutional blocks and learnable upsampling techniques, \ie bilinear upsampling followed by a single $1 \times 1$ convolution.
The radial log-scale output $\mathbf{R}_{\log} \in \mathbb{R}^{H\times{}W}$ is computed from the upsampled features and transformed to $\mathbf{R}$ via element-wise exponentiation.
The final 3D spherical output $\mathbf{O} = \mathbf{C} || \mathbf{R}$ is converted to a Cartesian point cloud $\mathbf{O} \in \mathbb{R}^{H \times W \times 3}$ using a spherical-to-Cartesian coordinate transformation.
Also, we predict a confidence map ($\mathbf{\Sigma}$) for the radial outputs by including a second projection head fed with upsampled $\mathbf{D}$ features, besides the first head of the Radial Module which computes $\mathbf{R}_{\log}$.

\noindent{}\textbf{Optimization.} The optimization process is defined by three different losses. 
The angular loss $\mathcal{L}_{\text{AA}}$ is applied on $\theta$ and $\phi$ separately, with $\mathcal{L}^{0.7}_{\text{AA}}$ and $\mathcal{L}^{0.5}_{\text{AA}}$ for $\theta$ and $\phi$, respectively.
The final angular loss can be expressed as
\begin{equation}
    \mathcal{L}_{\text{A}}(\hat{\mathbf{C}}, \mathbf{C}^*) = \beta \mathcal{L}^{0.7}_{\text{AA}}(\hat{\theta}, \theta^*) + (1-\beta) \mathcal{L}^{0.5}_{\text{AA}}(\hat{\phi}, \phi^*),
\end{equation}
with $\hat{(\cdot)}$ and $(\cdot)^*$ serving as prediction and GT identifiers, respectively, and $\beta = 0.75$.
It is worth noting that $\mathcal{L}^{0.5}_{\text{AA}}$ corresponds to the standard, symmetric L1-loss, as the azimuthal dimension $\phi$ w.r.t.\ the principal point is \emph{not} affected by prediction contraction.
Our radial loss is the L1-loss between the predicted and GT log-radius obtained by the GT camera and depth: $\mathcal{L}_{\text{rad}} = \left\| \hat{\mathbf{R}}_{\log} - \mathbf{R}^*_{\log} \right\|_1$.
The confidence loss is the L1-loss between the detached radial loss and the inverse predicted confidence, $\mathbf{\Sigma}$: $\mathcal{L}_{\text{conf}} = \left\|  |\hat{\mathbf{R}}_{\log} - \mathbf{R}^*_{\log}| - \mathbf{\Sigma} \right\|_1$.
The loss is a linear combination of the three losses: $\mathcal{L}_{\text{A}} + \eta \mathcal{L}_{\text{rad}} + \gamma \mathcal{L}_{\text{conf}}$, with $\eta = 2$ and $\gamma = 0.1$.

\section{Experiments}
\label{sec:experiments}


\noindent{}\textbf{Training Datasets.}
The training dataset accounts for 26 different sources: A2D2~\cite{geyer2020a2d2}, aiMotive~\cite{matuszka2023aimotive}, Argoverse2~\cite{2021argoverse2}, ARKit-Scenes~\cite{baruch2021arkitscenes}, ASE~\cite{engel2023ase}, BEDLAM~\cite{black2023bedlam}, BlendedMVS~\cite{yao2020blendedmvs}, DL3DV~\cite{ling2024dl3dv}, DrivingStereo~\cite{yang2019drivingstereo}, DynamicReplica~\cite{karaev2023dynamicreplica}, EDEN~\cite{le2021eden}, FutureHouse~\cite{li2022fuutrehouse}, HOI4D~\cite{liu2022hoi4d}, HM3D~\cite{ramakrishnan2021habitat}, Matterport3D~\cite{chang2017matterport3d}, Mapillary-PSD~\cite{Lopez2020mapillary}, MatrixCity~\cite{li2023matrixcity}, MegaDepth~\cite{li2018megadepth}, NianticMapFree~\cite{arnold2022mapfree}, PointOdyssey~\cite{zheng2023pointodyssey}, ScanNet~\cite{dai2017scannet}, ScanNet++~(iPhone)~\cite{yeshwanthliu2023scannetpp}, TartanAir~\cite{wang2020tartanair}, Taskonomy~\cite{zamir2018taskonomy}, Waymo~\cite{sun2020waymo}, and WildRGBD~\cite{xia2024wildrgbd}.
More details are given in the supplement.

\noindent{}\textbf{Zero-shot Testing Datasets.}
We evaluate the generalizability of models by testing them on 13 datasets not seen during training, grouped in 4 different domains which are defined based on their camera type: 1) small FoV (S.FoV), \ie FoV $< 90^\circ$, 2) small FoV with radial and tangential distortions (S.FoV\textsubscript{Dist}), 3) large FoV (L.FoV), \ie FoV $> 120^\circ$, and 4) Panoramic (Pano) with $360^\circ$ viewing angle.
More specifically, the S.FoV group corresponds to the validation splits of NYU-Depth V2~\cite{silberman2012nyu}, KITTI Eigen-split~\cite{Geiger2012kitti} and nuScenes~\cite{nuscenes}, and the full IBims-1~\cite{koch2022ibims}, ETH-3D~\cite{schoeps2017eth3d}, and Diode Indoor~\cite{Vasiljevic2019diode}; the S.FoV\textsubscript{Dist} is composed by images artificially distorted from IBims-1, ETH-3D, and Diode Indoor (more details in the supplement); L.FoV is the mix of ADT~\cite{pan2023adt}, ScanNet++~(DSLR)~\cite{yeshwanthliu2023scannetpp}, and KITTI360~\cite{Liao2022KITTI360}; and Panoramic (Pano) is to the full Stanford-2D3D~\cite{armeni20172d3ds} dataset.

\noindent{}\textbf{Evaluation Details.}
All methods have been re-evaluated with a fair and consistent pipeline.
In particular, we do not exploit any test-time augmentations and utilize the same set of weights for all zero-shot evaluations.
We use the checkpoint corresponding to the zero-shot model for each method, \ie not fine-tuned on KITTI or NYU.
The metrics utilized in the main experiments are $\mathrm{\delta_1^{SSI}}$, $\mathrm{F_{A}}$, and $\mathrm{\rho_{A}}$.
Further metrics are reported in \cref{sec:supp:quant}.
$\mathrm{\delta_1^{SSI}}$ measures scale- and shift-invariant depth estimation performance.
$\mathrm{F_{A}}$ is the area under the curve (AUC) of F1-score~\cite{ornek20222metrics} up to $1/20$ of the datasets' maximum depth and evaluates monocular 3D estimation.
$\mathrm{\rho_{A}}$ evaluates the camera performance and is the AUC of the average angular error of camera rays up to 15$^{\circ}$, 20$^{\circ}$, 30$^{\circ}$ for S.FoV, L.Fov, and Pano, respectively.
We avoid parametric evaluations, such as those based on focal length or FoV, because they lack generality across diverse camera models.
Instead, our chosen metrics ensure applicability to any camera type, preserving fairness and consistency in evaluation.
\Cref{tab:supp:ft_kitti,tab:supp:ft_nyu} show the fine-tuning ability of \ourmodel by training the final checkpoint on KITTI and NYU-Depth V2 and evaluating in-domain, as per standard practice.

\noindent{}\textbf{Implementation Details.}
\ourmodel is implemented in PyTorch~\cite{pytorch} and CUDA~\cite{nickolls2008cuda}.
For training, we use the AdamW~\cite{Loshchilov2017adamw} optimizer ($\beta_1=0.9$, $\beta_2=0.999$) with an initial learning rate of $5\times{}10^{-5}$. The learning rate is divided by a factor of 10 for the backbone weights for every experiment and weight decay is set to $0.1$.
We exploit Cosine Annealing as learning rate scheduler to one-tenth starting from 30\% of the whole training.
We run 250k optimization iterations with a batch size of 128.
The training time amounts to 6 days on 16 NVIDIA 4090.
The dataset sampling procedure follows a weighted sampler, where the weight of each dataset is its number of scenes.
Our augmentations are both geometric and photometric, \ie random resizing and cropping for the former type, and brightness, gamma, saturation, and hue shift for the latter.
We randomly sample the image ratio per batch between 2:1 and 9:16.
Our ViT~\cite{Dosovitskiy2020VIT} backbone is initialized with weights from DINO-pre-trained~\cite{oquab2023dinov2} models.
For the ablations, we run 100k training steps with a ViT-S backbone, with training pipeline as for the main experiments.

\begin{table}[t]
\centering
\caption{
\textbf{Ablation on data.} \textit{Data} indicates whether training images include strongly distorted cameras, either from real data or synthesized from pinhole cameras. Output representation: depth.}
\vspace{-10pt}
\label{tab:results:ablations_data}
\resizebox{\linewidth}{!}{%
\begin{tabular}{ccc|cc|cc|cc|cc}
\toprule
& \multirow{2}{*}{\textbf{Model}} & \multirow{2}{*}{\textbf{Data}} & \multicolumn{2}{c|}{S.FoV} & \multicolumn{2}{c|}{S.FoV\textsubscript{Dist}} & \multicolumn{2}{c|}{L.FoV} & \multicolumn{2}{c}{Pano} \\
 & & & $\mathrm{F_A}\uparrow$ & $\mathrm{\rho_A}\uparrow$ & $\mathrm{F_A}\uparrow$ & $\mathrm{\rho_A}\uparrow$ & $\mathrm{F_A}\uparrow$ & $\mathrm{\rho_A}\uparrow$ & $\mathrm{F_A}\uparrow$ & $\mathrm{\rho_A}\uparrow$ \\
\midrule
1 & Pinhole & \xmark & $55.1$ & $79.2$ & $31.7$ & $60.0$ & $41.2$ & $35.1$ & $8.4$ & $4.2$ \\
2 & Pinhole & \cmark & $56.1$ & $81.1$ & $40.4$ & $58.2$ & $44.9$ & $43.1$ & $5.9$ & $3.0$ \\
3 & SH & \xmark & $56.1$ & $79.1$ & $34.5$ & $60.2$ & $47.1$ & $56.7$ & $11.3$ & $16.1$ \\
4 & SH & \cmark & $56.2$ & $79.4$ & $42.1$ & $62.7$ & $48.5$ & $60.8$ & $10.9$ & $14.8$ \\
\bottomrule
\end{tabular}%
}
\end{table}
\begin{table}[]
\centering
\caption{\textbf{Ablation on camera model.} \textit{Model} corresponds to the type of camera model for output rays and internal conditioning: pinhole, Zernike-polynomial coefficients, SH coefficients, or non-parametric, \ie predicting one ray per pixel. All experiments are with full data, augmentation, model components, and radial output.}
\vspace{-10pt}
\label{tab:results:ablations_basis}
\resizebox{\linewidth}{!}{%
\begin{tabular}{cc|cc|cc|cc|cc}
\toprule
 & \multirow{2}{*}{\textbf{Model}} & \multicolumn{2}{c|}{S.FoV} & \multicolumn{2}{c|}{S.FoV\textsubscript{Dist}} & \multicolumn{2}{c|}{L.FoV} & \multicolumn{2}{c}{Pano} \\
 & & $\mathrm{F_A}\uparrow$ & $\mathrm{\rho_A}\uparrow$ & $\mathrm{F_A}\uparrow$ & $\mathrm{\rho_A}\uparrow$ & $\mathrm{F_A}\uparrow$ & $\mathrm{\rho_A}\uparrow$ & $\mathrm{F_A}\uparrow$ & $\mathrm{\rho_A}\uparrow$ \\
\midrule

1 & Pinhole & $55.5$ & $79.9$ & $52.5$ & $73.8$ & $45.2$ & $47.9$ & $24.6$ & $16.4$ \\
2 & Zernike & $56.6$ & $80.9$ & $39.9$ & $51.3$  & $49.9$ & $54.6$ & $31.8$ & $17.9$ \\
3 & Non-Parametric & $56.4$ & $81.0$ & $45.2$ & $62.8$  & $42.0$ & $42.8$ & $51.7$ & $20.1$ \\
4 & SH & $57.3$ & $79.8$ & $44.6$ & $59.3$ & $53.5$ & $64.8$ & $58.6$ & $26.3$ \\
\bottomrule
\end{tabular}%
}
\end{table}
\begin{table}[t]
\centering
\caption{
\textbf{Ablation on output representation.} \textit{Output} refers to the type of the 3\textsuperscript{rd} dimension of the predicted output space: either Cartesian z-axis depth or spherical radius, \ie distance. All experiments are with full data and augmentation.}
\vspace{-10pt}
\label{tab:results:ablations_rad}
\resizebox{\linewidth}{!}{%
\begin{tabular}{ccc|cc|cc|cc|cc}
\toprule
& \multirow{2}{*}{\textbf{Model}} & \multirow{2}{*}{\textbf{Output}} & \multicolumn{2}{c|}{S.FoV} & \multicolumn{2}{c|}{S.FoV\textsubscript{Dist}} & \multicolumn{2}{c|}{L.FoV} & \multicolumn{2}{c}{Pano} \\
 & & & $\mathrm{F_A}\uparrow$ & $\mathrm{\rho_A}\uparrow$ & $\mathrm{F_A}\uparrow$ & $\mathrm{\rho_A}\uparrow$ & $\mathrm{F_A}\uparrow$ & $\mathrm{\rho_A}\uparrow$ & $\mathrm{F_A}\uparrow$ & $\mathrm{\rho_A}\uparrow$ \\
\midrule
1 & Pinhole & depth & $56.1$ & $81.1$ & $40.4$ & $58.2$ & $44.9$ & $43.1$ & $5.9$ & $3.0$ \\
2 & Pinhole & radius & $56.0$ & $81.1$ & $39.5$ & $57.6$ & $44.4$ & $48.9$ & $10.1$ & $4.9$ \\
3 & SH & depth & $56.2$ & $79.4$ & $42.1$ & $62.7$ & $48.5$ & $60.8$ & $10.9$ & $14.8$ \\
4 & SH & radius & $56.8$ & $76.7$ & $35.0$ & $43.7$ & $51.8$ & $61.1$ & $53.8$ & $22.0$ \\
\bottomrule
\end{tabular}%
}
\end{table}

\subsection{Comparison with The State of The Art}
\label{ssec:experiments:comparison}

\Cref{tab:results:aggregated_comparisons} presents a comprehensive comparison of \ourmodel against existing SotA methods across various FoV and image types.
Our model consistently outperforms prior models, especially in challenging large-FoV and panoramic scenarios.
For instance, in the L.FoV domain, \ourmodel achieves a remarkable $\delta_{\text{SSI}}^1$ of $91.2\%$ and $\mathrm{F_A}$ of $71.6\%$, outperforming the second-best method by more than $20\%$ and $40\%$,
respectively.
This substantial improvement underscores the robustness of our unified spherical framework in handling wide FoVs.
In the Pano category, our model's $\delta_{\text{SSI}}^1$ and $\mathrm{F_A}$ scores of $71.2\%$ and $66.1\%$ also set the new SotA, demonstrating its ability to effectively reconstruct 3D geometry even under extreme camera setups.
These results validate that our design choices, including the SH-based camera model and radial output representation, are crucial for maintaining high performance in complex and diverse camera settings.

In addition, \cref{fig:results:main_vis} clearly shows how \ourmodel can estimate the 3D geometry of scenes from various and distorted cameras.
This is in contrast to other methods that fail when facing unconventional or non-pinhole camera images, as depicted by the 2\textsuperscript{nd}, 3\textsuperscript{rd}, and 4\textsuperscript{th} columns.
It is important to highlight that Metric3D, Metric3Dv2, and ZoeDepth are evaluated using GT camera parameters for the $\mathrm{F_A}$ score, while \ourmodel, UniDepth, MASt3R, and DepthPro rely on their predicted cameras.
Despite this added difficulty, \ourmodel still demonstrates superior 3D reconstruction performance, showcasing its strength in handling real-world conditions where precise camera information is unavailable.
Interestingly, our method does not sacrifice performance in more conventional, small-FoV scenarios.
\ourmodel keeps its top rank, with a $\delta_{\text{SSI}}^1$ of $94.3$ in the S.FoV setting, outperforming previously leading methods.
This balance highlights that our advancements in L.FoV representation do not undermine the model's effectiveness for S.FoV tasks.
$\mathrm{F_A}$ scores remain high in S.FoV and the $\rho_A$ metric shows that our model consistently provides accurate camera parameter estimation.

Moreover, \ourmodel is competitive with specialized methods for equirectangular images, as demonstrated in \Cref{tab:results:equi_comparisons}.
This shows how our model can incorporate different scene and camera domains at training time without compromising any domain-specific performance.

\begin{table}[]
\centering
\caption{
\textbf{Ablation on network components.} \textit{$\mathcal{L}_{\text{AA}}$} indicates if our asymmetric angular loss is used, L1-loss otherwise. \textit{Cond} indicates if our design for enhanced camera conditioning from \cref{ssec:method:engineer} is utilized. All experiments are with full data and augmentations, radial output representation, and an SH-based camera model.}
\vspace{-10pt}
\label{tab:results:ablations_components}
\resizebox{0.9\linewidth}{!}{%
\begin{tabular}{ccc|cc|cc|cc|cc}
\toprule
& \multirow{2}{*}{$\mathbf{\mathcal{L}}_{AA}$} & \multirow{2}{*}{\textbf{Cond}} & \multicolumn{2}{c|}{S.FoV} & \multicolumn{2}{c|}{S.FoV\textsubscript{Dist}} & \multicolumn{2}{c|}{L.FoV} & \multicolumn{2}{c}{Pano} \\
 & & & $\mathrm{F_A}\uparrow$ & $\mathrm{\rho_A}\uparrow$ & $\mathrm{F_A}\uparrow$ & $\mathrm{\rho_A}\uparrow$ & $\mathrm{F_A}\uparrow$ & $\mathrm{\rho_A}\uparrow$ & $\mathrm{F_A}\uparrow$ & $\mathrm{\rho_A}\uparrow$ \\
\midrule
1 & \xmark & \xmark & $56.8$ & $76.7$ & $35.0$ & $43.7$ & $51.8$ & $61.1$ & $53.8$ & $22.0$ \\
2 & \cmark & \xmark & $57.7$ & $80.9$ & $39.5$ & $52.1$ & $52.9$ & $64.2$ & $56.1$ & $24.4$ \\
3 & \cmark & \cmark & $57.3$ & $79.8$ & $44.6$ & $59.3$ & $53.5$ & $64.8$ & $58.6$ & $26.3$ \\
\bottomrule
\end{tabular}%
}
\vspace{-5pt}
\end{table}

\subsection{Ablation Studies}
\label{ssec:experiments:ablations}

\noindent{}\textbf{Data.} \Cref{tab:results:ablations_data} demonstrates the effect of training on datasets with and without large FoV and camera distortions.
Incorporating images with strong camera distortions generally enhances performance across all domains, particularly in challenging cases such as S.FoV with distortion and L.FoV.
This underscores the importance of diverse camera geometries in the training set to achieve better generalization.
However, the improvement on Pano is limited due to the difficulty of representing panoramic images using a log-depth representation.

\noindent{}\textbf{Camera Model.} As shown in \Cref{tab:results:ablations_basis}, employing SH as the basis for camera rays yields the best overall performance, particularly on L.FoV and Pano.
This highlights the effectiveness of our basis function selection in capturing diverse camera models.
By contrast, the non-parametric model underperforms in $\mathrm{F_A}$ and $\mathrm{\rho_A}$.
Since the latter formulation is purely data-driven, we presume that it requires significantly more data to generalize well.
It tends to underrepresent the tails of the data distribution, \ie L.FoV and Pano, while performing adequately on more common domains, \ie S.FoV with or without distortion.
The Zernike-polynomial basis~\cite{Zernike1934optics}, typically used for modeling lens aberrations, struggles to represent spherical or equirectangular camera geometries due to its inherent planar structure.

\noindent{}\textbf{Output Space.} \Cref{tab:results:ablations_rad} compares different output representations for the third dimension of the predicted space: either the Cartesian z-axis (rows 1 and 3) or the spherical radius (rows 2 and 4).
The results show that using the radius representation improves reconstruction metrics in Pano and L.FoV settings, as depth is less effective when dealing with FoVs near or exceeding 180 degrees.
This improvement is realized only when the radial component is paired with a camera model capable of representing a wide range of geometries, \eg our SH-based model (row 4 \vs row 2).
However, the radius-based output space leads to poorer reconstruction for S.FoV with distortion (row 3 \vs row 4).
This degradation occurs because the radius representation is more sensitive to minor angular variations, which disproportionately impacts accuracy in small but highly distorted views.

\begin{figure}[t]
    \centering
    \footnotesize
    \includegraphics[width=0.9\linewidth]{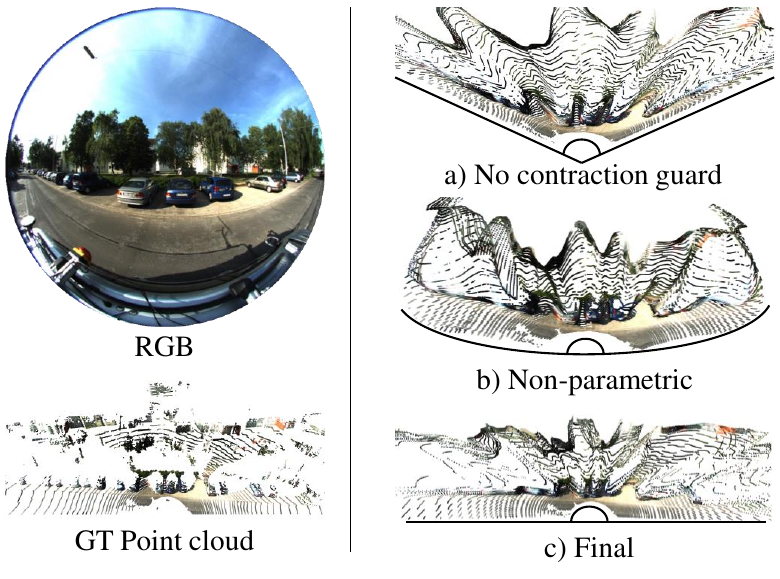}
    \vspace{-10pt}
    \caption{\textbf{FoV effects.}
    The image on the left showcases the challenge of representing the full 180\textsuperscript{$\circ$} FoV, alongside the GT point cloud.
    The effect of FoV contraction occurs when no ``guarding'', \ie asymmetric loss ($\mathcal{L}_{\text{AA}}$) and camera conditioning, is put in force, as shown in a).
    The total absence of any prior may lead to impossible and inconsistent backprojection, as shown in b).
    The final \ourmodel is depicted in c), clearly showing the ability to recover large FoVs with a sensible camera backprojection model.
    }
    \label{fig:experiments:fov}
\end{figure}

\noindent{}\textbf{Components.} \Cref{tab:results:ablations_components} examines the impact of our asymmetric angular loss ($\mathcal{L}_{\text{AA}}$) and our strategies designed to enhance camera conditioning.
Our full model, which leverages both the asymmetric loss and the improved conditioning (row 3), significantly outperforms those that do not, especially in distorted and L.FoV domains.
This demonstrates the efficacy of our combined strategies in preventing contraction in backprojection and improving angular prediction accuracy.
The overall gains are rather due to the synergy of combining these contributions.
Moreover, these strategies aim at mitigating extreme cases, which may not be easily represented in aggregate quantitative results, but are clearly visible in qualitative samples as in~\cref{fig:experiments:fov}.

\section{Conclusion}
\label{sec:conclusions}

We have presented \ourmodel, the first universal framework for monocular 3D estimation that generalizes seamlessly across diverse camera models, from pinhole to fisheye and panoramic.
Our approach introduces strategies to prevent FOV contraction and supports accurate metric 3D estimation through a flexible and robust design for backprojection with any generic camera model.
While expanding the diversity and coverage of training data could even further enhance the robustness and applicability of \ourmodel, the latter already achieves compelling generalization to unseen cameras and 3D scene domains far beyond the capabilities of the previous state of the art, with only a fair quantity of data.

\vfill
\noindent{}\textbf{Acknowledgment.} This work is funded by Toyota Motor Europe via the research project TRACE-Z\"urich.

\newpage

{
    \small
    \bibliographystyle{ieeenat_fullname}
    \bibliography{main}

\begin{thebibliography}{89}
\providecommand{\natexlab}[1]{#1}
\providecommand{\url}[1]{\texttt{#1}}
\expandafter\ifx\csname urlstyle\endcsname\relax
  \providecommand{\doi}[1]{doi: #1}\else
  \providecommand{\doi}{doi: \begingroup \urlstyle{rm}\Url}\fi

\bibitem[Antequera et~al.(2020)Antequera, Gargallo, Hofinger, Bul{\`o}, Kuang, and Kontschieder]{Lopez2020mapillary}
Manuel~Lopez Antequera, Pau Gargallo, Markus Hofinger, Samuel~Rota Bul{\`o}, Yubin Kuang, and Peter Kontschieder.
\newblock Mapillary planet-scale depth dataset.
\newblock In \emph{The European Conference on Computer Vision (ECCV)}, pages 589--604. Springer International Publishing, 2020.

\bibitem[Armeni et~al.(2017)Armeni, Sax, Zamir, and Savarese]{armeni20172d3ds}
Iro Armeni, Sasha Sax, Amir~R Zamir, and Silvio Savarese.
\newblock Joint 2d-3d-semantic data for indoor scene understanding.
\newblock \emph{arXiv preprint arXiv:1702.01105}, 2017.

\bibitem[Arnold et~al.(2022)Arnold, Wynn, Vicente, Garcia-Hernando, Monszpart, Prisacariu, Turmukhambetov, and Brachmann]{arnold2022mapfree}
Eduardo Arnold, Jamie Wynn, Sara Vicente, Guillermo Garcia-Hernando, {\'{A}}ron Monszpart, Victor~Adrian Prisacariu, Daniyar Turmukhambetov, and Eric Brachmann.
\newblock Map-free visual relocalization: Metric pose relative to a single image.
\newblock In \emph{European Conference on Computer Vision (ECCV)}, 2022.

\bibitem[Ba et~al.(2016)Ba, Kiros, and Hinton]{Ba2016layer}
Jimmy~Lei Ba, Jamie~Ryan Kiros, and Geoffrey~E Hinton.
\newblock Layer normalization.
\newblock \emph{arXiv preprint arXiv:1607.06450}, 2016.

\bibitem[Baruch et~al.(2021)Baruch, Chen, Dehghan, Dimry, Feigin, Fu, Gebauer, Joffe, Kurz, Schwartz, and Shulman]{baruch2021arkitscenes}
Gilad Baruch, Zhuoyuan Chen, Afshin Dehghan, Tal Dimry, Yuri Feigin, Peter Fu, Thomas Gebauer, Brandon Joffe, Daniel Kurz, Arik Schwartz, and Elad Shulman.
\newblock {ARK}itscenes - a diverse real-world dataset for 3d indoor scene understanding using mobile {RGB}-d data.
\newblock In \emph{Advances in Neural Information Processing Systems (NIPS)}, 2021.

\bibitem[Bhat et~al.(2020)Bhat, Alhashim, and Wonka]{Bhat2020adabins}
Shariq~Farooq Bhat, Ibraheem Alhashim, and Peter Wonka.
\newblock Adabins: Depth estimation using adaptive bins.
\newblock \emph{Proceedings of the IEEE/CVF Conference on Computer Vision and Pattern Recognition (CVPR)}, pages 4008--4017, 2020.

\bibitem[Bhat et~al.(2023)Bhat, Birkl, Wofk, Wonka, and M{\"u}ller]{bhat2023zoedepth}
Shariq~Farooq Bhat, Reiner Birkl, Diana Wofk, Peter Wonka, and Matthias M{\"u}ller.
\newblock Zoedepth: Zero-shot transfer by combining relative and metric depth.
\newblock \emph{arXiv preprint arXiv:2302.12288}, 2023.

\bibitem[Black et~al.(2023)Black, Patel, Tesch, and Yang]{black2023bedlam}
Michael~J. Black, Priyanka Patel, Joachim Tesch, and Jinlong Yang.
\newblock {BEDLAM}: A synthetic dataset of bodies exhibiting detailed lifelike animated motion.
\newblock In \emph{Proceedings of the IEEE/CVF Conference on Computer Vision and Pattern Recognition (CVPR)}, pages 8726--8737, 2023.

\bibitem[Bochkovskii et~al.(2024)Bochkovskii, Delaunoy, Germain, Santos, Zhou, Richter, and Koltun]{bochkovskii2024depthpro}
Aleksei Bochkovskii, Ama{\"e}l Delaunoy, Hugo Germain, Marcel Santos, Yichao Zhou, Stephan~R Richter, and Vladlen Koltun.
\newblock Depth pro: Sharp monocular metric depth in less than a second.
\newblock \emph{arXiv preprint arXiv:2410.02073}, 2024.

\bibitem[Caesar et~al.(2020)Caesar, Bankiti, Lang, Vora, Liong, Xu, Krishnan, Pan, Baldan, and Beijbom]{nuscenes}
Holger Caesar, Varun Bankiti, Alex~H. Lang, Sourabh Vora, Venice~Erin Liong, Qiang Xu, Anush Krishnan, Yu Pan, Giancarlo Baldan, and Oscar Beijbom.
\newblock nuscenes: A multimodal dataset for autonomous driving.
\newblock In \emph{Proceedings of the IEEE/CVF Conference on Computer Vision and Pattern Recognition (CVPR)}, 2020.

\bibitem[Chang et~al.(2017)Chang, Dai, Funkhouser, Halber, Niessner, Savva, Song, Zeng, and Zhang]{chang2017matterport3d}
Angel Chang, Angela Dai, Thomas Funkhouser, Maciej Halber, Matthias Niessner, Manolis Savva, Shuran Song, Andy Zeng, and Yinda Zhang.
\newblock Matterport3d: Learning from rgb-d data in indoor environments.
\newblock In \emph{Proceedings of the International Conference on 3D Vision (3DV)}, 2017.

\bibitem[Dai et~al.(2017)Dai, Chang, Savva, Halber, Funkhouser, and Nie{\ss}ner]{dai2017scannet}
Angela Dai, Angel~X. Chang, Manolis Savva, Maciej Halber, Thomas Funkhouser, and Matthias Nie{\ss}ner.
\newblock Scannet: Richly-annotated 3d reconstructions of indoor scenes.
\newblock In \emph{Proceedings of the IEEE/CVF Conference on Computer Vision and Pattern Recognition (CVPR)}, 2017.

\bibitem[Deng et~al.(2022)Deng, Liu, Zhu, and Ramanan]{deng2022nerf}
Kangle Deng, Andrew Liu, Jun-Yan Zhu, and Deva Ramanan.
\newblock Depth-supervised nerf: Fewer views and faster training for free.
\newblock In \emph{Proceedings of the IEEE/CVF Conference on Computer Vision and Pattern Recognition}, pages 12882--12891, 2022.

\bibitem[Dong et~al.(2022)Dong, Garratt, Anavatti, and Abbass]{dong2022depth4robotics}
Xingshuai Dong, Matthew~A Garratt, Sreenatha~G Anavatti, and Hussein~A Abbass.
\newblock Towards real-time monocular depth estimation for robotics: A survey.
\newblock \emph{IEEE Transactions on Intelligent Transportation Systems}, 23\penalty0 (10):\penalty0 16940--16961, 2022.

\bibitem[Dosovitskiy et~al.(2021)Dosovitskiy, Beyer, Kolesnikov, Weissenborn, Zhai, Unterthiner, Dehghani, Minderer, Heigold, Gelly, Uszkoreit, and Houlsby]{Dosovitskiy2020VIT}
Alexey Dosovitskiy, Lucas Beyer, Alexander Kolesnikov, Dirk Weissenborn, Xiaohua Zhai, Thomas Unterthiner, Mostafa Dehghani, Matthias Minderer, Georg Heigold, Sylvain Gelly, Jakob Uszkoreit, and Neil Houlsby.
\newblock An image is worth 16x16 words: Transformers for image recognition at scale.
\newblock In \emph{International Conference on Learning Representations (ICLR)}. OpenReview.net, 2021.

\bibitem[Eigen et~al.(2014)Eigen, Puhrsch, and Fergus]{Eigen2014}
David Eigen, Christian Puhrsch, and Rob Fergus.
\newblock Depth map prediction from a single image using a multi-scale deep network.
\newblock In \emph{Advances in Neural Information Processing Systems (NeurIPS)}, pages 2366--2374. Neural information processing systems foundation, 2014.

\bibitem[Engel et~al.(2023)Engel, Somasundaram, Goesele, Sun, Gamino, Turner, Talattof, Yuan, Souti, Meredith, et~al.]{engel2023ase}
Jakob Engel, Kiran Somasundaram, Michael Goesele, Albert Sun, Alexander Gamino, Andrew Turner, Arjang Talattof, Arnie Yuan, Bilal Souti, Brighid Meredith, et~al.
\newblock Project aria: A new tool for egocentric multi-modal ai research.
\newblock \emph{arXiv preprint arXiv:2308.13561}, 2023.

\bibitem[Frits(1934)]{Zernike1934optics}
Zernike Frits.
\newblock Beugungstheorie des schneidenver-fahrens und seiner verbesserten form, der phasenkontrastmethode.
\newblock \emph{Physica: Nonlinear Phenomena}, 1:\penalty0 689--704, 1934.

\bibitem[Fu et~al.(2018)Fu, Gong, Wang, Batmanghelich, and Tao]{Fu2018Dorn}
Huan Fu, Mingming Gong, Chaohui Wang, Kayhan Batmanghelich, and Dacheng Tao.
\newblock Deep ordinal regression network for monocular depth estimation.
\newblock \emph{Proceedings of the IEEE/CVF Conference on Computer Vision and Pattern Recognition (CVPR)}, pages 2002--2011, 2018.

\bibitem[Garg et~al.(2016)Garg, Kumar, Carneiro, and Reid]{Garg2016}
Ravi Garg, B.~G.~Vijay Kumar, Gustavo Carneiro, and Ian Reid.
\newblock Unsupervised cnn for single view depth estimation: Geometry to the rescue.
\newblock \emph{Lecture Notes in Computer Science}, 9912 LNCS:\penalty0 740--756, 2016.

\bibitem[Geiger et~al.(2012)Geiger, Lenz, and Urtasun]{Geiger2012kitti}
Andreas Geiger, Philip Lenz, and Raquel Urtasun.
\newblock Are we ready for autonomous driving? {The} {KITTI} vision benchmark suite.
\newblock In \emph{Proceedings of the IEEE/CVF Conference on Computer Vision and Pattern Recognition (CVPR)}, 2012.

\bibitem[Geyer and Daniilidis(2000)]{Geyer2000ucm}
Christopher Geyer and Kostas Daniilidis.
\newblock A unifying theory for central panoramic systems and practical applications.
\newblock In \emph{The European Conference on Computer Vision (ECCV)}, 2000.

\bibitem[Geyer et~al.(2020)Geyer, Kassahun, Mahmudi, Ricou, Durgesh, Chung, Hauswald, Pham, M{\"u}hlegg, Dorn, Fernandez, J{\"a}nicke, Mirashi, Savani, Sturm, Vorobiov, Oelker, Garreis, and Schuberth]{geyer2020a2d2}
Jakob Geyer, Yohannes Kassahun, Mentar Mahmudi, Xavier Ricou, Rupesh Durgesh, Andrew~S. Chung, Lorenz Hauswald, Viet~Hoang Pham, Maximilian M{\"u}hlegg, Sebastian Dorn, Tiffany Fernandez, Martin J{\"a}nicke, Sudesh Mirashi, Chiragkumar Savani, Martin Sturm, Oleksandr Vorobiov, Martin Oelker, Sebastian Garreis, and Peter Schuberth.
\newblock {A2D2: Audi Autonomous Driving Dataset}.
\newblock \emph{arXiv preprint arXiv:2004.06320}, 2020.

\bibitem[Guizilini et~al.(2023)Guizilini, Vasiljevic, Chen, Ambruș, and Gaidon]{guizilini2023zerodepth}
Vitor Guizilini, Igor Vasiljevic, Dian Chen, Rareș Ambruș, and Adrien Gaidon.
\newblock Towards zero-shot scale-aware monocular depth estimation.
\newblock In \emph{Proceedings of the IEEE/CVF International Conference on Computer Vision (ICCV)}, pages 9233--9243, 2023.

\bibitem[He et~al.(2015)He, Zhang, Ren, and Sun]{He2015}
Kaiming He, Xiangyu Zhang, Shaoqing Ren, and Jian Sun.
\newblock Deep residual learning for image recognition.
\newblock \emph{Proceedings of the IEEE/CVF Conference on Computer Vision and Pattern Recognition (CVPR)}, 2016-December:\penalty0 770--778, 2015.

\bibitem[Hendrycks and Gimpel(2016)]{Hendrycks2016gelu}
Dan Hendrycks and Kevin Gimpel.
\newblock Bridging nonlinearities and stochastic regularizers with gaussian error linear units.
\newblock \emph{CoRR}, abs/1606.08415, 2016.

\bibitem[Hirose and Tahara(2021)]{Hirose2021Depth360}
Noriaki Hirose and Kosuke Tahara.
\newblock Depth360: Self-supervised learning for monocular depth estimation using learnable camera distortion model.
\newblock \emph{2022 IEEE/RSJ International Conference on Intelligent Robots and Systems (IROS)}, pages 317--324, 2021.

\bibitem[Hu et~al.(2024)Hu, Yin, Zhang, Cai, Long, Chen, Wang, Yu, Shen, and Shen]{hu2024metric3dv2}
Mu Hu, Wei Yin, Chi Zhang, Zhipeng Cai, Xiaoxiao Long, Hao Chen, Kaixuan Wang, Gang Yu, Chunhua Shen, and Shaojie Shen.
\newblock Metric3d v2: A versatile monocular geometric foundation model for zero-shot metric depth and surface normal estimation.
\newblock \emph{arXiv preprint arXiv:2404.15506}, 2024.

\bibitem[Jiang et~al.(2021)Jiang, Sheng, Zhu, Dong, and Huang]{jiang2021unifuse}
Hualie Jiang, Zhe Sheng, Siyu Zhu, Zilong Dong, and Rui Huang.
\newblock Unifuse: Unidirectional fusion for 360 panorama depth estimation.
\newblock \emph{IEEE Robotics and Automation Letters (RA-L)}, 6\penalty0 (2):\penalty0 1519--1526, 2021.

\bibitem[Kannala and Brandt(2006)]{Kannala2006KB}
Juho Kannala and Sami~Sebastian Brandt.
\newblock A generic camera model and calibration method for conventional, wide-angle, and fish-eye lenses.
\newblock \emph{IEEE Transactions on Pattern Analysis and Machine Intelligence (T-PAMI}, 28:\penalty0 1335--1340, 2006.

\bibitem[Karaev et~al.(2023)Karaev, Rocco, Graham, Neverova, Vedaldi, and Rupprecht]{karaev2023dynamicreplica}
Nikita Karaev, Ignacio Rocco, Benjamin Graham, Natalia Neverova, Andrea Vedaldi, and Christian Rupprecht.
\newblock Dynamicstereo: Consistent dynamic depth from stereo videos.
\newblock In \emph{Proceedings of the IEEE/CVF Conference on Computer Vision and Pattern Recognition (CVPR)}, 2023.

\bibitem[Ke et~al.(2024)Ke, Obukhov, Huang, Metzger, Daudt, and Schindler]{ke2024marigold}
Bingxin Ke, Anton Obukhov, Shengyu Huang, Nando Metzger, Rodrigo~Caye Daudt, and Konrad Schindler.
\newblock Repurposing diffusion-based image generators for monocular depth estimation.
\newblock In \emph{Proceedings of the IEEE/CVF Conference on Computer Vision and Pattern Recognition (CVPR)}, pages 9492--9502, 2024.

\bibitem[Khomutenko et~al.(2016)Khomutenko, Garcia, and Martinet]{Khomutenko2016EUCM}
Bogdan Khomutenko, Ga{\"e}tan Garcia, and Philippe Martinet.
\newblock An enhanced unified camera model.
\newblock \emph{IEEE Robotics and Automation Letters (RA-L)}, 1:\penalty0 137--144, 2016.

\bibitem[Koch et~al.(2020)Koch, Liebel, Körner, and Fraundorfer]{koch2022ibims}
Tobias Koch, Lukas Liebel, Marco Körner, and Friedrich Fraundorfer.
\newblock Comparison of monocular depth estimation methods using geometrically relevant metrics on the {IBims-1} dataset.
\newblock \emph{Computer Vision and Image Understanding (CVIU)}, 191:\penalty0 102877, 2020.

\bibitem[Kumar et~al.(2020)Kumar, Yogamani, Bach, Witt, Milz, and M{\"a}der]{Kumar2020UnRectDepth}
Varun~Ravi Kumar, Senthil~Kumar Yogamani, Markus Bach, Christian Witt, Stefan Milz, and Patrick M{\"a}der.
\newblock Unrectdepthnet: Self-supervised monocular depth estimation using a generic framework for handling common camera distortion models.
\newblock \emph{2020 IEEE/RSJ International Conference on Intelligent Robots and Systems (IROS)}, pages 8177--8183, 2020.

\bibitem[Laina et~al.(2016)Laina, Rupprecht, Belagiannis, Tombari, and Navab]{Laina2016}
Iro Laina, Christian Rupprecht, Vasileios Belagiannis, Federico Tombari, and Nassir Navab.
\newblock Deeper depth prediction with fully convolutional residual networks.
\newblock \emph{Proceedings of the International Conference on 3D Vision (3DV)}, pages 239--248, 2016.

\bibitem[Le et~al.(2021)Le, Mensink, Das, Karaoglu, and Gevers]{le2021eden}
Hoang-An Le, Thomas Mensink, Partha Das, Sezer Karaoglu, and Theo Gevers.
\newblock Eden: Multimodal synthetic dataset of enclosed garden scenes.
\newblock In \emph{Proceedings of the IEEE/CVF Winter Conference on Applications of Computer Vision (WACV)}, pages 1579--1589, 2021.

\bibitem[Lee et~al.(2019)Lee, Han, Ko, and Suh]{Lee2019bts}
Jin~Han Lee, Myung{-}Kyu Han, Dong~Wook Ko, and Il~Hong Suh.
\newblock From big to small: Multi-scale local planar guidance for monocular depth estimation.
\newblock \emph{CoRR}, abs/1907.10326, 2019.

\bibitem[Leroy et~al.(2024)Leroy, Cabon, and Revaud]{leroy2024master}
Vincent Leroy, Yohann Cabon, and J{\'e}r{\^o}me Revaud.
\newblock Grounding image matching in 3d with mast3r.
\newblock \emph{arXiv preprint arXiv:2406.09756}, 2024.

\bibitem[Li et~al.(2023)Li, Jiang, Xu, Xiangli, Wang, Lin, and Dai]{li2023matrixcity}
Yixuan Li, Lihan Jiang, Linning Xu, Yuanbo Xiangli, Zhenzhi Wang, Dahua Lin, and Bo Dai.
\newblock Matrixcity: A large-scale city dataset for city-scale neural rendering and beyond.
\newblock In \emph{Proceedings of the IEEE/CVF International Conference on Computer Vision (ICCV)}, pages 3205--3215, 2023.

\bibitem[Li and Snavely(2018)]{li2018megadepth}
Zhengqi Li and Noah Snavely.
\newblock Megadepth: Learning single-view depth prediction from internet photos.
\newblock In \emph{Proceedings of the IEEE/CVF Conference on Computer Vision and Pattern Recognition (CVPR)}, pages 2041--2050, 2018.

\bibitem[Li et~al.(2022)Li, Wang, Huang, Pan, and Yang]{li2022fuutrehouse}
Zhen Li, Lingli Wang, Xiang Huang, Cihui Pan, and Jiaqi. Yang.
\newblock Phyir: Physics-based inverse rendering for panoramic indoor images.
\newblock In \emph{Proceedings of the IEEE/CVF Conference on Computer Vision and Pattern Recognition (CVPR)}, 2022.

\bibitem[Liao et~al.(2022)Liao, Xie, and Geiger]{Liao2022KITTI360}
Yiyi Liao, Jun Xie, and Andreas Geiger.
\newblock {KITTI}-360: A novel dataset and benchmarks for urban scene understanding in 2d and 3d.
\newblock \emph{IEEE Transactions on Pattern Analysis and Machine Intelligence (T-PAMI)}, 2022.

\bibitem[Ling et~al.(2024)Ling, Sheng, Tu, Zhao, Xin, Wan, Yu, Guo, Yu, Lu, et~al.]{ling2024dl3dv}
Lu Ling, Yichen Sheng, Zhi Tu, Wentian Zhao, Cheng Xin, Kun Wan, Lantao Yu, Qianyu Guo, Zixun Yu, Yawen Lu, et~al.
\newblock {DL3DV}-10k: A large-scale scene dataset for deep learning-based 3d vision.
\newblock In \emph{Proceedings of the IEEE/CVF Conference on Computer Vision and Pattern Recognition (CVPR)}, pages 22160--22169, 2024.

\bibitem[Liu et~al.(2015)Liu, Shen, Lin, and Reid]{Liu2015}
Fayao Liu, Chunhua Shen, Guosheng Lin, and Ian Reid.
\newblock Learning depth from single monocular images using deep convolutional neural fields.
\newblock \emph{IEEE Transactions on Pattern Analysis and Machine Intelligence (T-PAMI)}, 38:\penalty0 2024--2039, 2015.

\bibitem[Liu et~al.(2022)Liu, Liu, Jiang, Lyu, Wan, Shen, Liang, Fu, Wang, and Yi]{liu2022hoi4d}
Yunze Liu, Yun Liu, Che Jiang, Kangbo Lyu, Weikang Wan, Hao Shen, Boqiang Liang, Zhoujie Fu, He Wang, and Li Yi.
\newblock Hoi4d: A 4d egocentric dataset for category-level human-object interaction.
\newblock In \emph{Proceedings of the IEEE/CVF Conference on Computer Vision and Pattern Recognition (CVPR)}, pages 21013--21022, 2022.

\bibitem[Loshchilov and Hutter(2017)]{Loshchilov2017adamw}
Ilya Loshchilov and Frank Hutter.
\newblock Decoupled weight decay regularization.
\newblock \emph{7th International Conference on Learning Representations, ICLR 2019}, 2017.

\bibitem[Matuszka et~al.(2023)Matuszka, Barton, Butykai, Hajas, Kiss, Kov{\'a}cs, Kuns{\'a}gi-M{\'a}t{\'e}, Lengyel, N{\'e}meth, Pet{\H{o}}, Ribli, Szeghy, Vajna, and Varga]{matuszka2023aimotive}
Tamas Matuszka, Ivan Barton, {\'A}d{\'a}m Butykai, P{\'e}ter Hajas, D{\'a}vid Kiss, Domonkos Kov{\'a}cs, S{\'a}ndor Kuns{\'a}gi-M{\'a}t{\'e}, P{\'e}ter Lengyel, G{\'a}bor N{\'e}meth, Levente Pet{\H{o}}, Dezs{\H{o}} Ribli, D{\'a}vid Szeghy, Szabolcs Vajna, and Balint~Viktor Varga.
\newblock aimotive dataset: A multimodal dataset for robust autonomous driving with long-range perception.
\newblock In \emph{International Conference on Learning Representations (ICLR) Workshop on Scene Representations for Autonomous Driving}, 2023.

\bibitem[Mei and Rives(2007)]{Mei2007mei}
Christopher Mei and Patrick Rives.
\newblock Single view point omnidirectional camera calibration from planar grids.
\newblock \emph{Proceedings of the IEEE International Conference on Robotics and Automation (ICRA)}, pages 3945--3950, 2007.

\bibitem[Nathan~Silberman and Fergus(2012)]{silberman2012nyu}
Pushmeet~Kohli Nathan~Silberman, Derek~Hoiem and Rob Fergus.
\newblock Indoor segmentation and support inference from rgbd images.
\newblock In \emph{The European Conference on Computer Vision (ECCV)}, 2012.

\bibitem[Neyman and Pearson(1933)]{neyman1933tests}
Jerzy Neyman and Egon~Sharpe Pearson.
\newblock On the problem of the most efficient tests of statistical hypotheses.
\newblock \emph{Philosophical Transactions of the Royal Society of London. Series A, Containing Papers of a Mathematical or Physical Character}, 231\penalty0 (694-706):\penalty0 289--337, 1933.

\bibitem[Nickolls et~al.(2008)Nickolls, Buck, Garland, and Skadron]{nickolls2008cuda}
John Nickolls, Ian Buck, Michael Garland, and Kevin Skadron.
\newblock Scalable parallel programming with cuda: Is cuda the parallel programming model that application developers have been waiting for?
\newblock \emph{Queue}, 6\penalty0 (2):\penalty0 40--53, 2008.

\bibitem[Oquab et~al.(2023)Oquab, Darcet, Moutakanni, Vo, Szafraniec, Khalidov, Fernandez, Haziza, Massa, El-Nouby, et~al.]{oquab2023dinov2}
Maxime Oquab, Timoth{\'e}e Darcet, Th{\'e}o Moutakanni, Huy Vo, Marc Szafraniec, Vasil Khalidov, Pierre Fernandez, Daniel Haziza, Francisco Massa, Alaaeldin El-Nouby, et~al.
\newblock Dinov2: Learning robust visual features without supervision.
\newblock \emph{arXiv preprint arXiv:2304.07193}, 2023.

\bibitem[{\"O}rnek et~al.(2022){\"O}rnek, Mudgal, Wald, Wang, Navab, and Tombari]{ornek20222metrics}
Evin~P{\i}nar {\"O}rnek, Shristi Mudgal, Johanna Wald, Yida Wang, Nassir Navab, and Federico Tombari.
\newblock From 2d to 3d: Re-thinking benchmarking of monocular depth prediction.
\newblock \emph{arXiv preprint arXiv:2203.08122}, 2022.

\bibitem[Pan et~al.(2023)Pan, Charron, Yang, Peters, Whelan, Kong, Parkhi, Newcombe, and Ren]{pan2023adt}
Xiaqing Pan, Nicholas Charron, Yongqian Yang, Scott Peters, Thomas Whelan, Chen Kong, Omkar Parkhi, Richard Newcombe, and Yuheng~Carl Ren.
\newblock Aria digital twin: A new benchmark dataset for egocentric 3d machine perception.
\newblock In \emph{Proceedings of the IEEE/CVF International Conference on Computer Vision (ICCV)}, pages 20133--20143, 2023.

\bibitem[Park et~al.(2021)Park, Ambrus, Guizilini, Li, and Gaidon]{park2021dd3d}
Dennis Park, Rares Ambrus, Vitor Guizilini, Jie Li, and Adrien Gaidon.
\newblock Is pseudo-lidar needed for monocular 3d object detection?
\newblock In \emph{IEEE/CVF International Conference on Computer Vision (ICCV)}, 2021.

\bibitem[Paszke et~al.(2019)Paszke, Gross, Massa, Lerer, Bradbury, Chanan, Killeen, Lin, Gimelshein, Antiga, Desmaison, Kopf, Yang, DeVito, Raison, Tejani, Chilamkurthy, Steiner, Fang, Bai, and Chintala]{pytorch}
Adam Paszke, Sam Gross, Francisco Massa, Adam Lerer, James Bradbury, Gregory Chanan, Trevor Killeen, Zeming Lin, Natalia Gimelshein, Luca Antiga, Alban Desmaison, Andreas Kopf, Edward Yang, Zachary DeVito, Martin Raison, Alykhan Tejani, Sasank Chilamkurthy, Benoit Steiner, Lu Fang, Junjie Bai, and Soumith Chintala.
\newblock Pytorch: An imperative style, high-performance deep learning library.
\newblock In \emph{Advances in Neural Information Processing Systems (NeurIPS)}, pages 8024--8035. Curran Associates, Inc., 2019.

\bibitem[Patil et~al.(2022)Patil, Sakaridis, Liniger, and Gool]{Patil2022p3depth}
Vaishakh Patil, Christos Sakaridis, Alexander Liniger, and Luc~Van Gool.
\newblock {P3Depth}: Monocular depth estimation with a piecewise planarity prior.
\newblock In \emph{Proceedings of the IEEE/CVF Conference on Computer Vision and Pattern Recognition (CVPR)}, 2022.

\bibitem[Piccinelli et~al.(2023)Piccinelli, Sakaridis, and Yu]{piccinelli2023idisc}
Luigi Piccinelli, Christos Sakaridis, and Fisher Yu.
\newblock {iDisc}: Internal discretization for monocular depth estimation.
\newblock In \emph{Proceedings of the IEEE/CVF Conference on Computer Vision and Pattern Recognition (CVPR)}, 2023.

\bibitem[Piccinelli et~al.(2024)Piccinelli, Yang, Sakaridis, Segu, Li, Van~Gool, and Yu]{piccinelli2024unidepth}
Luigi Piccinelli, Yung-Hsu Yang, Christos Sakaridis, Mattia Segu, Siyuan Li, Luc Van~Gool, and Fisher Yu.
\newblock Unidepth: Universal monocular metric depth estimation.
\newblock In \emph{Proceedings of the IEEE/CVF Conference on Computer Vision and Pattern Recognition (CVPR)}, pages 10106--10116, 2024.

\bibitem[Piccinelli et~al.(2025)Piccinelli, Sakaridis, Yang, Segu, Li, Abbeloos, and Gool]{piccinelli2025unidepthv2}
Luigi Piccinelli, Christos Sakaridis, Yung-Hsu Yang, Mattia Segu, Siyuan Li, Wim Abbeloos, and Luc~Van Gool.
\newblock {U}ni{D}epth{V2}: Universal monocular metric depth estimation made simpler.
\newblock \emph{arXiv:2502.20110}, 2025.

\bibitem[Ramakrishnan et~al.(2021)Ramakrishnan, Gokaslan, Wijmans, Maksymets, Clegg, Turner, Undersander, Galuba, Westbury, Chang, Savva, Zhao, and Batra]{ramakrishnan2021habitat}
Santhosh~Kumar Ramakrishnan, Aaron Gokaslan, Erik Wijmans, Oleksandr Maksymets, Alexander Clegg, John~M Turner, Eric Undersander, Wojciech Galuba, Andrew Westbury, Angel~X Chang, Manolis Savva, Yili Zhao, and Dhruv Batra.
\newblock Habitat-matterport 3d dataset ({HM}3d): 1000 large-scale 3d environments for embodied {AI}.
\newblock In \emph{Advances in Neural Information Processing Systems (NIPS)}, 2021.

\bibitem[Ranftl et~al.(2020)Ranftl, Lasinger, Hafner, Schindler, and Koltun]{ranftl2020midas}
Ren{\'e} Ranftl, Katrin Lasinger, David Hafner, Konrad Schindler, and Vladlen Koltun.
\newblock Towards robust monocular depth estimation: Mixing datasets for zero-shot cross-dataset transfer.
\newblock \emph{IEEE Transactions on Pattern Analysis and Machine Intelligence (T-PAMI)}, 44\penalty0 (3):\penalty0 1623--1637, 2020.

\bibitem[Scaramuzza(2014)]{Scaramuzza2014Omnidirectional}
Davide Scaramuzza.
\newblock Omnidirectional camera.
\newblock In \emph{Computer Vision, A Reference Guide}, 2014.

\bibitem[Sch\"ops et~al.(2017)Sch\"ops, Sch\"onberger, Galliani, Sattler, Schindler, Pollefeys, and Geiger]{schoeps2017eth3d}
Thomas Sch\"ops, Johannes~L. Sch\"onberger, Silvano Galliani, Torsten Sattler, Konrad Schindler, Marc Pollefeys, and Andreas Geiger.
\newblock A multi-view stereo benchmark with high-resolution images and multi-camera videos.
\newblock In \emph{Proceedings of the IEEE/CVF Conference on Computer Vision and Pattern Recognition (CVPR)}, 2017.

\bibitem[Simon and Liu(2020)]{niklaus2022softmaxsplatting}
Niklaus Simon and Feng Liu.
\newblock Softmax splatting for video frame interpolation.
\newblock In \emph{Proceedings of the IEEE/CVF Conference on Computer Vision and Pattern Recognition (CVPR)}, 2020.

\bibitem[Sun et~al.(2020)Sun, Kretzschmar, Dotiwalla, Chouard, Patnaik, Tsui, Guo, Zhou, Chai, Caine, et~al.]{sun2020waymo}
Pei Sun, Henrik Kretzschmar, Xerxes Dotiwalla, Aurelien Chouard, Vijaysai Patnaik, Paul Tsui, James Guo, Yin Zhou, Yuning Chai, Benjamin Caine, et~al.
\newblock Scalability in perception for autonomous driving: Waymo open dataset.
\newblock In \emph{Proceedings of the IEEE/CVF Conference on Computer Vision and Pattern Recognition (CVPR)}, pages 2446--2454, 2020.

\bibitem[Touvron et~al.(2021)Touvron, Cord, Sablayrolles, Synnaeve, and J{\'e}gou]{touvron2021layerscale}
Hugo Touvron, Matthieu Cord, Alexandre Sablayrolles, Gabriel Synnaeve, and Herv{\'e} J{\'e}gou.
\newblock Going deeper with image transformers.
\newblock In \emph{Proceedings of the IEEE/CVF International Conference on Computer Vision (ICCV)}, pages 32--42, 2021.

\bibitem[Usenko et~al.(2018)Usenko, Demmel, and Cremers]{Usenko2018DS}
Vladyslav~C. Usenko, Nikolaus Demmel, and Daniel Cremers.
\newblock The double sphere camera model.
\newblock \emph{International Conference on 3D Vision (3DV)}, pages 552--560, 2018.

\bibitem[Vasiljevic et~al.(2019)Vasiljevic, Kolkin, Zhang, Luo, Wang, Dai, Daniele, Mostajabi, Basart, Walter, and Shakhnarovich]{Vasiljevic2019diode}
Igor Vasiljevic, Nicholas~I. Kolkin, Shanyi Zhang, Ruotian Luo, Haochen Wang, Falcon~Z. Dai, Andrea~F. Daniele, Mohammadreza Mostajabi, Steven Basart, Matthew~R. Walter, and Gregory Shakhnarovich.
\newblock {DIODE:} {A} dense indoor and outdoor depth dataset.
\newblock \emph{CoRR}, abs/1908.00463, 2019.

\bibitem[Wang et~al.(2020{\natexlab{a}})Wang, Yeh, Sun, Chiu, and Tsai]{wang2020bifuse}
Fu-En Wang, Yu-Hsuan Yeh, Min Sun, Wei-Chen Chiu, and Yi-Hsuan Tsai.
\newblock Bifuse: Monocular 360 depth estimation via bi-projection fusion.
\newblock In \emph{The IEEE/CVF Conference on Computer Vision and Pattern Recognition (CVPR)}, 2020{\natexlab{a}}.

\bibitem[Wang et~al.(2022)Wang, Yeh, Tsai, Chiu, and Sun]{wang2022bifuse++}
Fu-En Wang, Yu-Hsuan Yeh, Yi-Hsuan Tsai, Wei-Chen Chiu, and Min Sun.
\newblock Bifuse++: Self-supervised and efficient bi-projection fusion for 360 depth estimation.
\newblock \emph{IEEE Transactions on Pattern Analysis and Machine Intelligence (T-PAMI)}, 45\penalty0 (5):\penalty0 5448--5460, 2022.

\bibitem[Wang et~al.(2024{\natexlab{a}})Wang, Xu, Dai, Xiang, Deng, Tong, and Yang]{wang2024moge}
Ruicheng Wang, Sicheng Xu, Cassie Dai, Jianfeng Xiang, Yu Deng, Xin Tong, and Jiaolong Yang.
\newblock Moge: Unlocking accurate monocular geometry estimation for open-domain images with optimal training supervision.
\newblock \emph{arXiv preprint arXiv:2410.19115}, 2024{\natexlab{a}}.

\bibitem[Wang et~al.(2024{\natexlab{b}})Wang, Leroy, Cabon, Chidlovskii, and Revaud]{wang2024duster}
Shuzhe Wang, Vincent Leroy, Yohann Cabon, Boris Chidlovskii, and J{\'e}r{\^o}me Revaud.
\newblock Dust3r: Geometric 3d vision made easy.
\newblock In \emph{Proceedings of the IEEE/CVF Conference on Computer Vision and Pattern Recognition (CVPR)}, pages 20697--20709, 2024{\natexlab{b}}.

\bibitem[Wang et~al.(2020{\natexlab{b}})Wang, Zhu, Wang, Hu, Qiu, Wang, Hu, Kapoor, and Scherer]{wang2020tartanair}
Wenshan Wang, Delong Zhu, Xiangwei Wang, Yaoyu Hu, Yuheng Qiu, Chen Wang, Yafei Hu, Ashish Kapoor, and Sebastian Scherer.
\newblock Tartanair: A dataset to push the limits of visual slam.
\newblock In \emph{2020 IEEE/RSJ International Conference on Intelligent Robots and Systems (IROS)}, pages 4909--4916. IEEE, 2020{\natexlab{b}}.

\bibitem[Wang et~al.(2019)Wang, Chao, Garg, Hariharan, Campbell, and Weinberger]{wang2019depth4vehicles}
Yan Wang, Wei-Lun Chao, Divyansh Garg, Bharath Hariharan, Mark Campbell, and Kilian~Q Weinberger.
\newblock Pseudo-lidar from visual depth estimation: Bridging the gap in 3d object detection for autonomous driving.
\newblock In \emph{Proceedings of the IEEE/CVF Conference on Computer Vision and Pattern Recognition}, pages 8445--8453, 2019.

\bibitem[Wilson et~al.(2021)Wilson, Qi, Agarwal, Lambert, Singh, Khandelwal, Pan, Kumar, Hartnett, Pontes, Ramanan, Carr, and Hays]{2021argoverse2}
Benjamin Wilson, William Qi, Tanmay Agarwal, John Lambert, Jagjeet Singh, Siddhesh Khandelwal, Bowen Pan, Ratnesh Kumar, Andrew Hartnett, Jhony~Kaesemodel Pontes, Deva Ramanan, Peter Carr, and James Hays.
\newblock Argoverse 2: Next generation datasets for self-driving perception and forecasting.
\newblock In \emph{Advances in Neural Information Processing Systems}, 2021.

\bibitem[Xia et~al.(2024)Xia, Fu, Liu, and Wang]{xia2024wildrgbd}
Hongchi Xia, Yang Fu, Sifei Liu, and Xiaolong Wang.
\newblock Rgbd objects in the wild: Scaling real-world 3d object learning from rgb-d videos.
\newblock In \emph{Proceedings of the IEEE/CVF Conference on Computer Vision and Pattern Recognition (CVPR)}, pages 22378--22389, 2024.

\bibitem[Yang et~al.(2019)Yang, Song, Huang, Deng, Shi, and Zhou]{yang2019drivingstereo}
Guorun Yang, Xiao Song, Chaoqin Huang, Zhidong Deng, Jianping Shi, and Bolei Zhou.
\newblock Drivingstereo: A large-scale dataset for stereo matching in autonomous driving scenarios.
\newblock In \emph{Proceedings of the IEEE/CVF Conference on Computer Vision and Pattern Recognition (CVPR)}, 2019.

\bibitem[Yang et~al.(2021)Yang, Tang, Ding, Sebe, and Ricci]{Yang2021}
Guanglei Yang, Hao Tang, Mingli Ding, Nicu Sebe, and Elisa Ricci.
\newblock Transformer-based attention networks for continuous pixel-wise prediction.
\newblock \emph{Proceedings of the IEEE/CVF International Conference on Computer Vision (ICCV)}, pages 16249--16259, 2021.

\bibitem[Yang et~al.(2024{\natexlab{a}})Yang, Kang, Huang, Xu, Feng, and Zhao]{yang2024da1}
Lihe Yang, Bingyi Kang, Zilong Huang, Xiaogang Xu, Jiashi Feng, and Hengshuang Zhao.
\newblock Depth anything: Unleashing the power of large-scale unlabeled data.
\newblock In \emph{Proceedings of the IEEE/CVF Conference on Computer Vision and Pattern Recognition (CVPR)}, pages 10371--10381, 2024{\natexlab{a}}.

\bibitem[Yang et~al.(2024{\natexlab{b}})Yang, Kang, Huang, Zhao, Xu, Feng, and Zhao]{yang2024da2}
Lihe Yang, Bingyi Kang, Zilong Huang, Zhen Zhao, Xiaogang Xu, Jiashi Feng, and Hengshuang Zhao.
\newblock Depth anything v2.
\newblock \emph{arXiv preprint arXiv:2406.09414}, 2024{\natexlab{b}}.

\bibitem[Yao et~al.(2020)Yao, Luo, Li, Zhang, Ren, Zhou, Fang, and Quan]{yao2020blendedmvs}
Yao Yao, Zixin Luo, Shiwei Li, Jingyang Zhang, Yufan Ren, Lei Zhou, Tian Fang, and Long Quan.
\newblock Blendedmvs: A large-scale dataset for generalized multi-view stereo networks.
\newblock In \emph{Proceedings of the IEEE/CVF Conference on Computer Vision and Pattern Recognition (CVPR)}, pages 1790--1799, 2020.

\bibitem[Yeshwanth et~al.(2023)Yeshwanth, Liu, Nie{\ss}ner, and Dai]{yeshwanthliu2023scannetpp}
Chandan Yeshwanth, Yueh-Cheng Liu, Matthias Nie{\ss}ner, and Angela Dai.
\newblock Scannet++: A high-fidelity dataset of 3d indoor scenes.
\newblock In \emph{Proceedings of the IEEE/CVF International Conference on Computer Vision (ICCV)}, 2023.

\bibitem[Yin et~al.(2023)Yin, Zhang, Chen, Cai, Yu, Wang, Chen, and Shen]{yin2023metric3d}
Wei Yin, Chi Zhang, Hao Chen, Zhipeng Cai, Gang Yu, Kaixuan Wang, Xiaozhi Chen, and Chunhua Shen.
\newblock Metric3d: Towards zero-shot metric 3d prediction from a single image.
\newblock In \emph{Proceedings of the IEEE/CVF International Conference on Computer Vision (ICCV)}, pages 9043--9053, 2023.

\bibitem[Yuan et~al.(2022)Yuan, Gu, Dai, Zhu, and Tan]{Yuan2022newcrf}
Weihao Yuan, Xiaodong Gu, Zuozhuo Dai, Siyu Zhu, and Ping Tan.
\newblock Neural window fully-connected crfs for monocular depth estimation.
\newblock In \emph{Proceedings of the IEEE/CVF Conference on Computer Vision and Pattern Recognition (CVPR)}, pages 3906--3915. {IEEE}, 2022.

\bibitem[Zamir et~al.(2018)Zamir, Sax, Shen, Guibas, Malik, and Savarese]{zamir2018taskonomy}
Amir~R Zamir, Alexander Sax, William~B Shen, Leonidas Guibas, Jitendra Malik, and Silvio Savarese.
\newblock Taskonomy: Disentangling task transfer learning.
\newblock In \emph{Proceedings of the IEEE/CVF Conference on Computer Vision and Pattern Recognition (CVPR)}. IEEE, 2018.

\bibitem[Zheng et~al.(2023)Zheng, Harley, Shen, Wetzstein, and Guibas]{zheng2023pointodyssey}
Yang Zheng, Adam~W Harley, Bokui Shen, Gordon Wetzstein, and Leonidas~J Guibas.
\newblock Pointodyssey: A large-scale synthetic dataset for long-term point tracking.
\newblock In \emph{Proceedings of the IEEE/CVF International Conference on Computer Vision (ICCV)}, pages 19855--19865, 2023.

\bibitem[Zhou et~al.(2019)Zhou, Krähenbühl, and Koltun]{Zhou2019}
Brady Zhou, Philipp Krähenbühl, and Vladlen Koltun.
\newblock Does computer vision matter for action?
\newblock \emph{Science Robotics}, 4, 2019.

\end{thebibliography}
}

\clearpage
\renewcommand\thesection{\Alph{section}}

\appendix
\setcounter{section}{0}
\twocolumn[{%
 \centering
 \textbf{\Large  Supplementary Material }
 \vspace{20pt}
}]

This supplementary material offers further insights into our work.
In \cref{sec:supp:arch} we describe the network architecture in more detail, necessarily \cref{sec:supp:arch} overlaps with Sec. 3.
Moreover, we analyze the complexity of \ourmodel and compare it with other methods in \cref{ssec:supp:arch:complexity}.
Also, we provide further alternatives to our design choices and ablate them in \cref{ssec:supp:arch:alternative}.
\cref{sec:supp:train} outlines the training pipeline and hyperparameters chosen in \cref{sec:supp:train:hp}, altogether with training and validation data in \cref{sec:supp:train:data}, and the camera augmentations in \cref{sec:supp:train:aug} for completeness and reproducibility.
Furthermore, \cref{sec:supp:quant} provides a more detailed quantitative evaluation with per-dataset evaluation in \cref{sec:supp:quant:atomic}
The results corresponding to \ourmodel finetuned on KITTI and NYUv2 are reported in \cref{sec:supp:quant:ft}.
In \cref{sec:supp:faq}, we provide answers to possible questions that may arise.
Eventually, additional visualizations are provided in \cref{sec:supp:qual}.

\section{Architecture}
\label{sec:supp:arch}

\noindent{}\textbf{Encoder.} Our model architecture employs a Vision Transformer (ViT)~\cite{Dosovitskiy2020VIT} as the encoder, demonstrating its effectiveness across different scales, from Small to Large.
The ViT backbones were originally developed for classification tasks, and as such, we modify them by removing the final three layers: the pooling layer, the fully connected layer, and the $\mathrm{softmax}$ layer.
We extract feature maps and class tokens from the last four layers of the modified ViT backbone. These outputs are flattened and processed using LayerNorm~\cite{Ba2016layer} followed by a linear projection layer.
The linear layer maps the features and class tokens to a common channel dimension, which is set to 512, 384, and 256 for Large, Base, and Small ViT variants, respectively.
Importantly, the normalization and linear layer weights are distinct and are not shared between the different feature resolutions and the class tokens.
The dense feature maps are subsequently passed to the Radial Module, while the class tokens are directed to the Angular Module.

\noindent{}\textbf{Angular Module.} The four class tokens extracted from the encoder are first projected to dimensions of $3D$, $3D$, $5D$, and $7D$, respectively.
These are then divided into chunks based on the channel dimension $d$, yielding token groups of size $3$, $3$, $5$, and $7$.
These token groups serve as the initialization for domain tokens, representing the spherical harmonics (SH) coefficients: 1st-degree, 2nd-degree, and 3rd-degree, respectively.
In total, there are 18 tokens ($\mathbf{T}$), which are processed through two layers of a Transformer Encoder.
Each Transformer Encoder layer consists of self-attention with eight heads and a Multi-Layer Perceptron (MLP) that has a single hidden layer of dimension $4C$ and uses the Gaussian Error Linear Unit (GELU) activation function~\cite{Hendrycks2016gelu}.
Both self-attention and MLP layers include residual connections to improve learning stability.
Each of the 18 tokens is then projected to a scalar dimension.
The first three tokens specifically define the domain for the spherical harmonics.
The first token determines the horizontal field of view (HFov), calculated as $2\pi \cdot \sigma(\mathbf{T}_0)$, where $\sigma$ denotes the sigmoid function.
The second and third tokens represent the poles of the spherical harmonics, \ie the center of projection relative to the image shape, computed as $c_x = \frac{\sigma(\mathbf{T}_1) W}{2}$ and $c_y = \frac{\sigma(\mathbf{T}_2) H}{2}$, respectively, where $H$ and $W$ are the image height and width.
The vertical FoV is derived under the assumption of square pixels: $\mathrm{HFov} \times \frac{H}{W}$.
Using this domain definition, we compute the spherical harmonics up to the 3rd degree, excluding the constant component, yielding 15 harmonic tensors of size $\mathbb{R}^{H \times W \times 3}$.
The pencil of rays $\mathbf{C}$ is then constructed as a linear combination of these harmonics and the corresponding 15 processed tokens ($\mathbf{T_{3:18}}$).

\noindent{}\textbf{Radial Module.} The sine-encoded camera rays $\mathbf{C}$ are used to condition each resolution level of the dense feature maps $\mathbf{F}$ via a Transformer Decoder layer.
In this setup, the dense features $\mathbf{F}$ serve as the \textit{query}, while the sine-encoded camera rays provide the \textit{keys} and \textit{values}.
The cross-attention mechanism includes a residual connection without any learnable gain factors, such as LayerScale.
The conditioned features are then refined in a Feature Pyramid Network (FPN) manner: the deepest features are processed through two Residual Convolution blocks~\cite{He2015}, followed by bilinear upsampling and a projection step that halves the channel dimension.
These upsampled features are then combined with the features from the next layer, which are similarly projected to match channel dimension and upsampled using a single 2x2 transposed convolution.
This process continues until all remaining three feature maps are consumed.
The final output features are upsampled to the input image resolution and projected to a single-channel dimension, yielding the log-radius $\mathbf{R}_{\log}$.
The same projection, architectural-wise but with separate weights, is used to generate the log-confidence $\mathbf{\Sigma}_{\log}$.
The final radius and confidence values are obtained by exponentiating these tensors element-wise, transforming them from log-space to the original space.

\begin{table}[t]
    \centering
    \caption{\textbf{Parameters and efficiency comparison.} Comparison of performance of methods based on input size, latency, and number of trainable parameters.
    Tested on RTX3090 GPU, 16-bit precision float, and synchronized timers.
    The last two rows correspond to the Angular and Radial Modules evaluated independently. 
    All models are based on ViT-L backbone.}
    \vspace{-10pt}
    \resizebox{\linewidth}{!}{
    \begin{tabular}{l|cccc}
    \toprule
    \textbf{Method} & Input Size & Latency (ms) & Parameters (M)\\
    \toprule
    ZoeDepth~\cite{bhat2023zoedepth} & $512 \times 512$ &  144.8 & 345.9\\
    DepthAnything v2~\cite{yang2024da2} & $518 \times 518$ & 78.1 & 334.7\\
    UniDepth~\cite{piccinelli2024unidepth} & $518 \times 518$ & 146.4 & 347.0\\
    Metric3Dv2~\cite{hu2024metric3dv2} & $518 \times 518$ & 135.6 & 441.9\\
    MASt3R~\cite{hu2024metric3dv2} & $512 \times 512$ & 154.7 & 668.6\\
    DepthPro~\cite{bochkovskii2024depthpro} & $1536 \times 1536$ & 808.1 & 952.0\\
    \midrule
    \ourmodel & $518 \times 518$ & 88.4 & 358.8\\
    \hspace{1em} Radial Module & \textcolor{gray}{-} & 21.9 & 38.2\\
    \hspace{1em} Angular Module & \textcolor{gray}{-} & 3.1 & 12.1\\
    \bottomrule
    \end{tabular}}
    \label{tab:supp:inference}
\end{table}

\subsection{Complexity}
\label{ssec:supp:arch:complexity}

We perform a detailed analysis of the computational cost of \ourmodel, presented in \Cref{tab:supp:inference}, and compare it to other state-of-the-art methods.
To ensure a fair and consistent comparison, we use input sizes that are as similar as possible across all models.
However, this approach introduces certain challenges.
DepthPro, for instance, has an entangled and multi-resolution architecture, which complicates tuning the input size consistently across methods.
Its architectural design does not easily allow for adjustments, making it difficult to align with a standardized input size.
Additionally, the performance of models like DepthPro and Metric3D, as evaluated in our main experiments in Sec. 4, shows a significant drop when tested with image shapes that differ from those used during training.
This sensitivity highlights a fundamental limitation: these methods are heavily optimized for specific image resolutions, and deviations from these resolutions can lead to substantial performance degradation.
Consequently, while we strive to measure computation under the most equitable conditions, it is essential to acknowledge that these models are not well-suited for resolutions that differ from their training setup.
In contrast, \ourmodel is designed to be flexible \wrt image shape, maintaining robust performance across different resolutions.
For our experiments, we chose the same input shape as DepthAnything v2, as it provides a balanced trade-off between computational efficiency and performance.
Furthermore, to account for the asynchronous nature of CUDA kernel threading, we ensure precise inference time measurements by enabling proper synchronization and utilizing CUDA event recording.
This approach guarantees an accurate reflection of computational cost, avoiding any misrepresentation caused by asynchronous operations.
As shown in \Cref{tab:supp:inference}, \ourmodel is among the most efficient models.
The primary differences in computational cost, especially when compared to DepthAnything v2, stem from the inclusion of our Angular Module and Scale components.
These components are essential for our model to handle absolute metric depth and camera-specific adjustments, features that relative depth estimation networks do not require.
Despite this additional complexity, our model's efficiency remains competitive, underscoring its design's effectiveness in addressing diverse camera geometries while maintaining high performance.

\begin{table}[]
\centering
\caption{\textbf{Ablation on camera conditioning design.} \textit{Camera Cond.} corresponds to the type of camera conditioning employed to condition the depth features with camera ones. \textit{Add} refers to a simple addition in the feature space. \textit{Cat} represents a simple concatenation and projection from $2C$ to $C$ channel dimension. \textit{Prompt} is our attention-based conditioning.}
\vspace{-10pt}
\label{tab:supp:ablations_addcat}
\resizebox{\linewidth}{!}{%
\begin{tabular}{cc|cc|cc|cc|cc}
\toprule
 & \multirow{2}{*}{\textbf{Camera Cond.}} & \multicolumn{2}{c|}{S.FoV} & \multicolumn{2}{c|}{S.FoV\textsubscript{Dist}} & \multicolumn{2}{c|}{L.FoV} & \multicolumn{2}{c}{Pano} \\
 & & $\mathrm{F_A}\uparrow$ & $\mathrm{\rho_A}\uparrow$ & $\mathrm{F_A}\uparrow$ & $\mathrm{\rho_A}\uparrow$ & $\mathrm{F_A}\uparrow$ & $\mathrm{\rho_A}\uparrow$ & $\mathrm{F_A}\uparrow$ & $\mathrm{\rho_A}\uparrow$ \\
\midrule
1 & Add & $53.0$ & $78.9$ & $26.3$ & $41.6$ & $45.0$ & $58.5$ & $42.5$ & $18.2$ \\
2 & Cat & $54.7$ & $79.0$ & $28.7$ & $44.6$ & $46.6$ & $58.1$ & $42.3$ & $18.1$\\
3 & Prompt & $57.3$ & $79.8$ & $44.6$ & $59.3$ & $53.5$ & $64.8$ & $58.6$ & $26.3$ \\
\bottomrule
\end{tabular}%
}
\end{table}
\begin{table}[]
\centering
\caption{\textbf{Ablation on camera tokens processing.} \textit{T-Enc.} indicates if the camera tokens are processed in the Angular Module either via the transformer encoder layer or not, in the latter case the tokens are fed directly to the final projections.}
\vspace{-10pt}
\label{tab:supp:ablations_tenc}
\resizebox{\linewidth}{!}{%
\begin{tabular}{cc|cc|cc|cc|cc}
\toprule
 & \multirow{2}{*}{\textbf{T-Enc}} & \multicolumn{2}{c|}{S.FoV} & \multicolumn{2}{c|}{S.FoV\textsubscript{Dist}} & \multicolumn{2}{c|}{L.FoV} & \multicolumn{2}{c}{Pano} \\
 & & $\mathrm{F_A}\uparrow$ & $\mathrm{\rho_A}\uparrow$ & $\mathrm{F_A}\uparrow$ & $\mathrm{\rho_A}\uparrow$ & $\mathrm{F_A}\uparrow$ & $\mathrm{\rho_A}\uparrow$ & $\mathrm{F_A}\uparrow$ & $\mathrm{\rho_A}\uparrow$ \\
\midrule
1 & \xmark & $55.7$ & $77.3$ & $43.2$ & $56.6$ & $50.9$ & $63.2$ & $54.9$ & $20.7$ \\
2 & \cmark & $57.3$ & $79.8$ & $44.6$ & $59.3$ & $53.5$ & $64.8$ & $58.6$ & $26.3$ \\
\bottomrule
\end{tabular}%
}
\end{table}

\subsection{Architectural Alternatives}
\label{ssec:supp:arch:alternative}

Despite the camera conditioning has been proven superior in UniDepth~\cite{piccinelli2024unidepth}, we ablate alternative architectural choices for both the Transformer Encoder and Decoder components.
In particular, we have chosen the most typical alternatives for conditioning: a simple addition or concatenation in place. 
While the camera tokens processing ``alternative'' involves an identity that shortcuts the camera tokens to the final projection layers.
\Cref{tab:supp:ablations_tenc} shows how the camera tokens processing, via the encoder layer, does not present large changes, showing how the class tokens from different layers are already informative.
However, \Cref{tab:supp:ablations_addcat} clearly shows how the simpler conditioning alternatives, such as addition or concatenation, underperform our attention-based conditioning.
This highlights how conditioning plays an important role in final performance and how strongly designed conditioning is paramount to achieving proper generalization.

\section{Training Details}
\label{sec:supp:train}

\begin{table}[t]
    \centering
    \caption{\textbf{Training Hyperparamters.} All training hyperparameters with corresponding values are presented.}
    \vspace{-10pt}
    \resizebox{\linewidth}{!}{
    \begin{tabular}{l|c}
    \toprule
    Hyperparameter & Value\\
    \midrule
    Steps & 250k\\
    Batch Size & 128\\
    LR & $5\cdot 10^{-5}$\\
    LR Encoder & $5\cdot 10^{-6}$\\
    Optimizer & AdamW~\cite{Loshchilov2017adamw}\\
    $(\beta_1, \beta_2)$ & $(0.9, 0.999)$\\
    Weight Decay & 0.1\\
    Gradient Clip Norm & 1.0\\
    Precision & 16-bit Float\\
    \multirow{2}{*}{LR Scheduler} & Cosine to 0.1\\
    & start after 75k iters\\
    \multirow{2}{*}{EMA} & 0.9995\\
    & start after 75k iters\\
    \midrule
    Color jitter prob & 80\%\\
    Color jitter intensity & $[0.0, 0.5]$\\
    Gamma prob & 80\%\\
    Gamma intensity & $[0.5, 1.5]$\\
    Horizontal flip prob & 50\%\\
    Greyscale prob & 20\%\\
    Gaussian blur prob & 20\%\\
    Gaussian blur sigma & $[0.1,2.0]$\\
    Random zoom & $[0.5, 2.0]$\\
    Random translation & $[-0.05, 0.05]$\\
    \midrule
    Image ratio & $[1:2, 2:1]$\\
    \multirow{2}{*}{Resolution} & $0.28\text{MP}$ \\
    & $[0.2\text{MP},0.6\text{MP}]$ last 50k iters\\
    \bottomrule
    \end{tabular}}
    \label{tab:supp:hp}
\end{table}

\begin{table}[t]
    \centering
    \caption{\textbf{Training Losses.} Training losses with corresponding weight and input.}
    \vspace{-10pt}
    \resizebox{\linewidth}{!}{
    \begin{tabular}{l|ccc}
    \toprule
    Loss & Inputs & Weight & Parameters\\
    \midrule
    L1 & Radius (log) & 2.0 ($\eta$) & -\\
    L1-asymmetric & Polar & 0.75 & $\alpha = 0.7$\\
    L1 & Azimuth & 0.25 & -\\
    \multirow{2}{*}{L1}  & Confidence (log), & \multirow{2}{*}{0.1 ($\gamma$)} & \multirow{2}{*}{-}\\
     & Radius error (detached) & &\\    
    \bottomrule
    \end{tabular}}
    \label{tab:supp:loss}
\end{table}
\subsection{Hyperparameters.}
\label{sec:supp:train:hp}

The training parameters, \ie those for optimization, scheduling, and augmentations, are described in \Cref{tab:supp:hp}.
The losses utilized, with the input and corresponding weights, are outlined in \Cref{tab:supp:loss}.

\begin{table}[t]
    \centering
    \small
    \caption{\textbf{Training Datasets.} List of the validation datasets: number of images, scene type, acquisition method, and sampling frequency are reported. SfM: Structure-from-Motion. MVS: Multi-View Stereo. Syn: Synthetic. Rec: Mesh reconstruction. KB: Kannala-Brandt~\cite{Kannala2006KB}. Equi: Equirectangular}
    \vspace{-10pt}
    \resizebox{\columnwidth}{!}{
    \begin{tabular}{l|ccccc}
        \toprule
        \textbf{Dataset} & \textbf{Images} & \textbf{Scene} & \textbf{Acquisition} & \textbf{Camera} & \textbf{Sampling}\\
        \midrule
        A2D2~\cite{geyer2020a2d2} & 78k & Outdoor & LiDAR & Pinhole & 2.5\%\\
        aiMotive~\cite{matuszka2023aimotive} & 178k & Outdoor & LiDAR & Mei~\cite{Mei2007mei} & 0.3\%\\
        Argoverse2~\cite{2021argoverse2} & 403k & Outdoor & LiDAR & Pinhole & 7.6\%\\
        ARKit-Scenes~\cite{baruch2021arkitscenes} & 1.75M & Indoor & LiDAR & Pinhole & 1.3\%\\
        ASE~\cite{engel2023ase} & 2.72M & Indoor & Syn & Fisheye624 & 10.1\%\\
        BEDLAM~\cite{black2023bedlam} & 24k & Various & Syn & Pinhole & 2.0\%\\
        BlendedMVS~\cite{yao2020blendedmvs} & 114k & Outdoor & MVS & Pinhole & 2.5\%\\
        DL3DV~\cite{ling2024dl3dv} & 306k & Outdoor & SfM & KB~\cite{Kannala2006KB} & 4.7\%\\
        DrivingStereo~\cite{yang2019drivingstereo} & 63k & Outdoor & MVS & Pinhole & 2.5\%\\
        DynamicReplica~\cite{karaev2023dynamicreplica} & 120k & Indoor & Syn & Pinhole & 1.3\%\\
        EDEN~\cite{le2021eden} & 368k & Outdoor & Syn & Pinhole & 2.5\%\\
        FutureHouse~\cite{li2022fuutrehouse} & 28.3 & Indoor & Syn & Equi & 2.5\%\\
        HOI4D~\cite{liu2022hoi4d} & 59k & Egocentric & RGB-D & KB~\cite{Kannala2006KB} & 1.7\%\\
        HM3D~\cite{ramakrishnan2021habitat} & 540k & Indoor & Rec & Pinhole & 5.2\%\\
        Matterport3D~\cite{chang2017matterport3d} & 10.8k & Indoor & Rec & Equi & 2.0\%\\
        Mapillary PSD~\cite{Lopez2020mapillary} & 742k & Outdoor & SfM & Pinhole & 2.0\%\\
        MatrixCity~\cite{li2023matrixcity} & 190k & Outdoor & Syn & Pinhole & 5.0\%\\
        MegaDepth~\cite{li2018megadepth} & 273k & Outdoor & SfM & Pinhole & 8.0\%\\
        NianticMapFree~\cite{arnold2022mapfree} & 25k & Outdoor & SfM & Pinhole & 2.0\%\\
        PointOdyssey~\cite{zheng2023pointodyssey} & 33k & Various & Syn & Pinhole & 1.7\%\\
        ScanNet~\cite{dai2017scannet} & 83k & Indoor & RGB-D & Pinhole & 5.0\%\\
        ScanNet++~\cite{yeshwanthliu2023scannetpp} & 39k & Indoor & Rec & Pinhole & 3.0\%\\
        TartanAir~\cite{wang2020tartanair} & 306k & Various & Syn & Pinhole & 5.5\%\\
        Taskonomy~\cite{zamir2018taskonomy} & 1.94M & Indoor & RGB-D & Pinhole & 6.0\%\\
        Waymo~\cite{sun2020waymo} & 223k & Outdoor & LiDAR & Pinhole & 7.5\%\\
        WildRGBD~\cite{xia2024wildrgbd} & 1.35M & Indoor & RGB-D & Pinhole & 7.5\%\\
        \bottomrule
    \end{tabular}}
    \label{tab:supp:train_ds}
\end{table}
\begin{table}[t]
    \centering
    \small
    \caption{\textbf{Validation Datasets.} List of the validation datasets: number of images, scene type, acquisition method, and max evaluation distance are reported. 1\textsuperscript{st} group: small FoV, 2\textsuperscript{nd} group: large FoV, 3\textsuperscript{rd}: Panoramic. Rec: Mesh reconstruction.}
    \vspace{-10pt}
    \resizebox{\columnwidth}{!}{
    \begin{tabular}{l|cccc}
        \toprule
        \textbf{Dataset} & \textbf{Images} & \textbf{Scene} & \textbf{Acquisition} & \textbf{Max Distance}\\
        \midrule
        KITTI~\cite{Geiger2012kitti} & 652 & Outdoor & LiDAR & 80.0\\
        NYU~\cite{silberman2012nyu} & 654 & Indoor & RGB-D & 10.0\\
        IBims-1~\cite{koch2022ibims} & 100 & Indoor & RGB-D & 25.0\\
        Diode~\cite{Vasiljevic2019diode} & 325 & Indoor & LiDAR & 25.0\\
        ETH3D~\cite{schoeps2017eth3d} & 454 & Outdoor & RGB-D & 50.0\\
        NuScenes~\cite{nuscenes} & 3.6k & Outdoor & LiDAR & 80.0\\
        \midrule
        ScanNet++~\cite{yeshwanthliu2023scannetpp} & 779 & Indoor & Rec & 10.0\\
        ADT~\cite{pan2023adt} & 469 & Indoor & Rec & 20.0\\
        KITTI360~\cite{Liao2022KITTI360} & 527 & Outdoor & LiDAR & 80.0\\
        \midrule
        Stanford-2D3D~\cite{armeni20172d3ds} & 1413 & Indoor & Rec & 10.0\\
        \bottomrule
    \end{tabular}}
    \label{tab:supp:val_ds}
\end{table}
\subsection{Data}
\label{sec:supp:train:data}

Details of training and validation datasets are presented in \Cref{tab:supp:train_ds} and \Cref{tab:supp:val_ds}.

\noindent{}\textbf{Training Datasets.} The datasets utilized for training are a mixture of different cameras and domains as shown in \Cref{tab:supp:train_ds}.
The sequence-based datasets are sub-sampled during collection in a way that the interval between two consecutive frames is not smaller than half a second.
No post-processing is applied.
The total amount of training samples accounts for more than 8M samples.
The datasets are sampled in each batch with a probability corresponding to the values in \textit{Sampling} column in \Cref{tab:supp:train_ds}.
This probability is related to the number of scenes present in each dataset.
However, probabilities are changed based on a simple qualitative data inspection, such that the most diverse datasets are sampled more.
Most of the datasets involve pinhole images or rectified cameras, \eg MegaDepth~\cite{li2018megadepth} or NianticMapFree~\cite{arnold2022mapfree}, other datasets provide only the pinhole calibration despite being clearly distorted, \ie Mapillary~\cite{Lopez2020mapillary}, there the entire samples are masked out in the camera loss computation. 

\noindent{}\textbf{Validation Datasets.} \Cref{tab:supp:val_ds} presents all the validation datasets and splits them into 3 groups: small FoV, large FoV, and Panoramic.
As per standard practice, KITTI Eigen-split corresponds to the corrected and accumulated GT depth maps with 45 images with inaccurate GT discarded from the original 697 images.
The small FoV with distortion presented in Sec. 3 and used for evaluation is obtained based on synthesized cameras from ETH3D, Diode (Indoor), and IBims-1, all distorted images and cameras are manually checked for realism, after being generated with the pipeline presented in \cref{sec:supp:train:aug}.

\begin{table}[t]
    \centering
    \caption{\textbf{Camera Sampling for S.FoV\textsubscript{Dist} generation.} The parameters to generate S.FoV\textsubscript{Dist} images are listed. We employed different camera models with different parameter ranges. The sampling is uniform sampling within the ranges. The seed is 13.
    }
    \vspace{-10pt}
    \resizebox{\linewidth}{!}{
    \begin{tabular}{l|ccc}
    \toprule
    Model & Probability & Parameter & Range\\
    \midrule
    \multirow{2}{*}{EUCM} & \multirow{2}{*}{0.1} & $\alpha$ & $[0,1]$\\
     & & $\beta$ & $[0.25, 4]$\\
    \midrule
    \multirow{3}{*}{Fisheye624} & \multirow{3}{*}{0.35} & $\{k_i\}_{i=1}^{6}$ & $[0.6,0.8]$\\
     & & $\{t_i\}_{i=1}^{2}$ & $[-0.01,0.01]$\\
     & & $\{s_i\}_{i=1}^{4}$ & $[-0.01,0.01]$\\
    \midrule
    \multirow{3}{*}{Fisheye624} & \multirow{3}{*}{0.35} & $\{k_i\}_{i=1}^{6}$ & $[-0.6,-0.4]$\\
     & & $\{t_i\}_{i=1}^{2}$& $[-0.01,0.01]$\\
     & & $\{s_i\}_{i=1}^{4}$ & $[-0.01,0.01]$\\
    \midrule
    \multirow{3}{*}{Fisheye624} & \multirow{3}{*}{0.2} & $\{k_i\}_{i=1}^{6}$ & $[-0.2,0.2]$\\
     & & $\{t_i\}_{i=1}^{2}$ & $[-0.05,0.05]$\\
     & & $\{s_i\}_{i=1}^{4}$ & $[-0.05,0.05]$\\
    \bottomrule
    \end{tabular}}
    \label{tab:supp:camera_gen}
\end{table}

\begin{table}[t]
    \centering
    \caption{\textbf{Camera Sampling for Camera Augmentation.}
    The parameters to generate an augmented camera during training images are listed. We employed different camera models with different parameter ranges. The sampling is uniform sampling within the ranges. When some parameters are not listed, \eg $\{k_i\}_{i=4}^{6}$ for Kannala-Brandt model, they are set to 0.
    }
    \vspace{-10pt}
    \resizebox{\linewidth}{!}{
    \begin{tabular}{l|ccc}
    \toprule
    Model & Probability & Parameter & Range\\
    \midrule
    \multirow{2}{*}{EUCM} & \multirow{2}{*}{0.1} & $\alpha$ & $[0,1]$\\
     & & $\beta$ & $[0.25, 4]$\\
    \midrule
    \multirow{3}{*}{Fisheye624} & \multirow{3}{*}{0.15} & $\{k_i\}_{i=1}^{6}$ & $[0.1,0.5]$\\
     & & $\{t_i\}_{i=1}^{2}$ & $[-0.005,0.005]$\\
     & & $\{s_i\}_{i=1}^{4}$ & $[-0.01,0.01]$\\
    \midrule
    \multirow{3}{*}{Fisheye624} & \multirow{3}{*}{0.15} & $\{k_i\}_{i=1}^{6}$ & $[-0.5,-0.1]$\\
     & & $\{t_i\}_{i=1}^{2}$& $[-0.005,0.005]$\\
     & & $\{s_i\}_{i=1}^{4}$ & $[-0.01,0.01]$\\
    \midrule
    \multirow{2}{*}{Kannala-Brandt} & \multirow{2}{*}{0.2} & $\{k_i\}_{i=1}^{3}$ & $[-0.05,0.05]$\\
     & & $\{t_i\}_{i=1}^{2}$ & $[-0.02,0.02]$\\
    \midrule
    \multirow{2}{*}{Kannala-Brandt} & \multirow{2}{*}{0.4} & $\{k_i\}_{i=1}^{3}$ & $[-0.5,0.5]$\\
     & & $\{t_i\}_{i=1}^{2}$ & $[-0.001,0.001]$\\
    \bottomrule
    \end{tabular}}
    \label{tab:supp:camera_aug}
\end{table}
\subsection{Camera Augmentations}
\label{sec:supp:train:aug}

To address the limited diversity of distorted camera data, we augment images captured with pinhole cameras by artificially deforming them, thereby simulating images from distorted camera models, \eg Fisheye624 or radial Kannala-Brandt~\cite{Kannala2006KB}.
The augmentation process involves two main steps.
First, we compute a deformation field. This starts with unprojecting the 2D depth map obtained from a pinhole camera into a 3D point cloud.
We then project these 3D points onto the image plane of a randomly sampled distorted camera model to obtain the new 2D coordinates.
The deformation field is defined as the distance between the original 2D image coordinates and the newly projected 2D coordinates.
This flow indicates how the original image should be warped to mimic the appearance of a distorted camera view.
Next, we warp the image using softmax-based splatting~\cite{niklaus2022softmaxsplatting}, a technique that projects pixels based on the computed deformation field while preserving image details.
To ensure the warping process does not create artifacts like holes, we use an ``importance'' metric, which is the inverse of the depth value for each pixel.
This metric prioritizes closer points, ensuring that details and correct parallax are maintained during the warping.
For non-synthetic images, where ground-truth depth maps are unavailable, we generate depth predictions in an inference-only mode to compute the deformation.
To ensure these predictions are accurate enough to create realistic deformations, we apply this augmentation only after the model has been trained for 10,000 steps.
By this point, the model has learned a decently reliable (scale-invariant) depth representation.
The specific camera parameters used to sample the new random camera are listed in \Cref{tab:supp:camera_aug}.

\noindent{}\textbf{Validation datasets generation.} Generating validation datasets for testing models on distorted images with reduced fields of view presents an additional challenge, as most distortions are typically associated with large fields of view.
To simulate this, we use synthetic camera parameters to deform RGB images from datasets such as ETH3D~\cite{schoeps2017eth3d}, IBims-1~\cite{koch2022ibims}, and Diode (Indoor)~\cite{Vasiljevic2019diode}.
These datasets are chosen because they provide nearly complete ground-truth depth maps, making the deformation process well-posed and realistic.
Any small gaps or holes in the depth maps are filled using inpainting.
Importantly, the 3D ground-truth data remains unchanged, as it is invariant to the camera model used.
To ensure realism, we manually validate each deformed image and will release both the code for data generation and the resulting validation data.

\section{Additional Quantitative Results}
\label{sec:supp:quant}


\begin{table}[]
    \centering
    \caption{\textbf{Comparison on NYU validation set.} All models are trained on NYU. The first four are trained only on NYU. The last four are fine-tuned on NYU.}
    \vspace{-10pt}
    \resizebox{\columnwidth}{!}{
    \begin{tabular}{l|ccc|ccc}
        \toprule
        \multirow{2}{*}{\textbf{Method}} & $\mathrm{\delta}_{1}$ & $\mathrm{\delta}_{2}$ & $\mathrm{\delta}_{3}$ & $\mathrm{A.Rel}$ & $\mathrm{RMS}$ & $\mathrm{Log}_{10}$\\
         & \multicolumn{3}{c|}{\textit{Higher is better}} & \multicolumn{3}{c}{\textit{Lower is better}}\\
        \toprule
        BTS~\cite{Lee2019bts} & $88.5$ & $97.8$ & $99.4$ & $10.9$ & $0.391$ & $0.046$\\
        AdaBins~\cite{Bhat2020adabins} & $90.1$ & $98.3$ & $99.6$ & $10.3$ & $0.365$ & $0.044$\\
        NeWCRF~\cite{Yuan2022newcrf} & $92.1$ & $99.1$ & $\scnd{99.8}$ & $9.56$ & $0.333$ & $0.040$\\
        iDisc~\cite{piccinelli2023idisc} & $93.8$ & $99.2$ & $\scnd{99.8}$ & $8.61$ & $0.313$ & $0.037$\\
        ZoeDepth~\cite{bhat2023zoedepth} & $95.2$ & $99.5$ & $\scnd{99.8}$ & $7.70$ & $0.278$ & $0.033$\\
        Metric3Dv2~\cite{hu2024metric3dv2} & $\mathbf{98.9}$ & $\mathbf{99.8}$ & $\mathbf{100}$ & $\scnd{4.70}$ & $\scnd{0.183}$ & $\scnd{0.020}$\\
        DepthAnythingv2~\cite{yang2024da2} & $\scnd{98.4}$ & $\mathbf{99.8}$ & $\mathbf{100}$ & $5.60$ & $0.206$ & $0.024$\\
        \midrule 
        \ourmodel & $\best{98.9}$ & $\best{99.8}$ & $\best{100}$ & $\best{4.43}$ & $\best{0.173}$ & $\best{0.019}$\\
        \bottomrule
    \end{tabular}}
    \label{tab:supp:ft_nyu}
\end{table}

\begin{table}[]
    \centering
    \caption{\textbf{Comparison on KITTI Eigen-split validation set.} All models are trained on KITTI E-ign-split training and tested on the corresponding validator split. The first are trained only on KITTI. The last 4 are fine-tuned on KITTI.}
    \vspace{-10pt}
    \resizebox{\columnwidth}{!}{
    \begin{tabular}{l|ccc|ccc}
        \toprule
        \multirow{2}{*}{\textbf{Method}} & $\mathrm{\delta}_{1}$ & $\mathrm{\delta}_{2}$ & $\mathrm{\delta}_{3}$ & $\mathrm{A.Rel}$ & $\mathrm{RMS}$ & $\mathrm{RMS}_{\log}$\\
         & \multicolumn{3}{c|}{\textit{Higher is better}} & \multicolumn{3}{c}{\textit{Lower is better}}\\
        \toprule
        BTS~\cite{Lee2019bts} & $96.2$ & $99.4$ & $99.8$ & $5.63$ & $2.43$ & $0.089$\\
        AdaBins~\cite{Bhat2020adabins} & $96.3$ & $99.5$ & $99.8$ & $5.85$ & $2.38$ & $0.089$\\
        NeWCRF~\cite{Yuan2022newcrf} & $97.5$ & $\underline{99.7}$ & $\underline{99.9}$ & $5.20$ & $2.07$ & $0.078$\\
        iDisc~\cite{piccinelli2023idisc} & $97.5$ & $\underline{99.7}$ & $\underline{99.9}$ &$5.09$ & $2.07$ & $0.077$\\
        ZoeDepth~\cite{bhat2023zoedepth} & $96.5$ & $99.1$ & $99.4$ & $5.76$ & $2.39$ & $0.089$ \\
        Metric3Dv2~\cite{yin2023metric3d} & $\underline{98.5}$ & $\mathbf{99.8}$ & $\mathbf{100}$ & $\underline{4.40}$ & $1.99$ & $\mathbf{0.064}$\\
        DepthAnythingv2~\cite{yang2024da2} & $98.3$ & $\mathbf{99.8}$ & $\mathbf{100}$ & $4.50$ & $\underline{1.86}$ & $\underline{0.067}$ \\
        \midrule
        \ourmodel & $\mathbf{99.0}$ & $\mathbf{99.8}$ & $\underline{99.9}$ & $\mathbf{3.69}$ & $\mathbf{1.68}$ & $\mathbf{0.060}$ \\
        \bottomrule
    \end{tabular}}
    \label{tab:supp:ft_kitti}
\end{table}

\subsection{Fine-tuning}
\label{sec:supp:quant:ft}

We evaluate the fine-tuning capability of \ourmodel by resuming training with either KITTI or NYU as the sole training dataset. The fine-tuning process starts from the weights and optimizer states obtained after the large-scale pretraining phase, ensuring a fair and consistent initialization.
The standard SILog loss is used as the training objective, with a batch size of 16, and the model is trained for an additional 40,000 steps.
To focus the evaluation on the impact of in-domain data, we disable all augmentations except for horizontal flipping and omit the asymmetric component of the angular loss during fine-tuning.
For evaluation, we adhere to the standard practices for both datasets to ensure comparability with prior work. KITTI results are reported using the Garg~\cite{Garg2016} evaluation crop, and the maximum evaluation depths for KITTI and NYU are set to 80 and 10 meters, respectively
mportantly, we do not apply any test-time augmentations or tuning, such as varying the input size, to maintain consistency and avoid introducing additional confounding factors.
Our results demonstrate that \ourmodel benefits significantly from in-domain fine-tuning. \Cref{tab:supp:ft_kitti} highlights the model's ability to perform exceptionally well on highly structured and calibrated datasets like KITTI, even though \ourmodel is inherently designed for flexibility and cross-domain generalization.
This suggests that the model can effectively adapt to well-structured data when fine-tuned.
This fine-tuning analysis highlights the adaptability of \ourmodel to diverse settings while maintaining its primary design focus on flexibility.
Similarly, \Cref{tab:supp:ft_nyu} shows that \ourmodel remains competitive when fine-tuned on less structured domains like NYU, which represent typical indoor environments.
These results reinforce the importance of in-domain data for achieving optimal performance, particularly on datasets with distinct properties or domain-specific challenges. 
In addition, the results underline the robustness of our model, as it achieves strong performance across significantly different dataset characteristics.

\begin{table}[ht]
\centering
\caption{\textbf{Comparison on zero-shot evaluation for NYUv2.} Missing values (-) indicate the model's inability to produce the respective output. \dag: ground-truth camera for 3D reconstruction. \ddag: ground-truth camera for 2D depth map inference.}
\label{tab:results:atomic_nyuv2depth}
\resizebox{\linewidth}{!}{%
\begin{tabular}{l|cccccc}
\toprule
\textbf{Method}  & $\mathrm{\delta_1}\uparrow$ & $\mathrm{A.Rel}\downarrow$ & $\mathrm{RMSE}\downarrow$ & $\mathrm{\delta_1^{SSI}}\uparrow$ & $\mathrm{F_A}\uparrow$ & $\mathrm{\rho_A}\uparrow$ \\
\midrule
DepthAnything~\cite{yang2024da1} & \textcolor{gray}{-} & \textcolor{gray}{-} & \textcolor{gray}{-} & $97.8$ & \textcolor{gray}{-} & \textcolor{gray}{-} \\
DepthAnythingv2~\cite{yang2024da2} & \textcolor{gray}{-} & \textcolor{gray}{-} & \textcolor{gray}{-} & $97.7$ & \textcolor{gray}{-} & \textcolor{gray}{-} \\
Metric3D\textsuperscript{\dag \ddag}~\cite{yin2023metric3d} & $68.1$ & $44.2$ & $1.23$ & $89.0$ & \textcolor{gray}{-} & \textcolor{gray}{-} \\
Metric3Dv2\textsuperscript{\dag \ddag}~\cite{hu2024metric3dv2} & $93.4$ & $9.1$ & $0.399$ & $98.1$ & \textcolor{gray}{-} & \textcolor{gray}{-} \\
ZoeDepth\textsuperscript{\dag}~\cite{bhat2023zoedepth} & $94.2$ & $8.2$ & $0.305$ & $98.0$ & \textcolor{gray}{-} & \textcolor{gray}{-} \\
UniDepth~\cite{piccinelli2024unidepth} & $\best{98.0}$ & $\best{7.3}$ & $\best{0.230}$ & $\best{99.0}$ & $\best{83.1}$ & $\best{99.2}$ \\
MASt3R~\cite{leroy2024master} & $83.9$ & $13.0$ & $0.435$ & $94.8$ & $69.6$ & $90.7$ \\
DepthPro~\cite{bochkovskii2024depthpro} & $92.2$ & $10.1$ & $0.357$ & $97.2$ & $73.0$ & $\scnd{93.1}$ \\
\midrule
\ourmodel-Small & $90.4$ & $11.2$ & $0.351$ & $97.4$ & $69.1$ & $83.0$ \\
\ourmodel-Base & $93.1$ & $10.3$ & $0.325$ & $97.9$ & $75.4$ & $89.1$ \\
\ourmodel-Large & $\scnd{96.5}$ & $\scnd{7.4}$ & $\scnd{0.259}$ & $\scnd{98.2}$ & $\scnd{82.5}$ & $91.2$ \\
\bottomrule
\end{tabular}%
}
\end{table}
\begin{table}[ht]
\centering
\caption{\textbf{Comparison on zero-shot evaluation for KITTI.} Missing values (-) indicate the model's inability to produce the respective output. \dag: ground-truth camera for 3D reconstruction. \ddag: ground-truth camera for 2D depth map inference.}
\label{tab:results:atomic_kitti}
\resizebox{\linewidth}{!}{%
\begin{tabular}{l|cccccc}
\toprule
\textbf{Method}  & $\mathrm{\delta_1}\uparrow$ & $\mathrm{A.Rel}\downarrow$ & $\mathrm{RMSE}\downarrow$ & $\mathrm{\delta_1^{SSI}}\uparrow$ & $\mathrm{F_A}\uparrow$ & $\mathrm{\rho_A}\uparrow$ \\
\midrule
DepthAnything~\cite{yang2024da1} & \textcolor{gray}{-} & \textcolor{gray}{-} & \textcolor{gray}{-} & $88.5$ & \textcolor{gray}{-} & \textcolor{gray}{-} \\
DepthAnythingv2~\cite{yang2024da2} & \textcolor{gray}{-} & \textcolor{gray}{-} & \textcolor{gray}{-} & $88.4$ & \textcolor{gray}{-} & \textcolor{gray}{-} \\
Metric3D\textsuperscript{\dag \ddag}~\cite{yin2023metric3d} & $3.3$ & $49.8$ & $10.35$ & $97.0$ & \textcolor{gray}{-} & \textcolor{gray}{-} \\
Metric3Dv2\textsuperscript{\dag \ddag}~\cite{hu2024metric3dv2} & $2.3$ & $56.3$ & $12.81$ & $96.7$ & \textcolor{gray}{-} & \textcolor{gray}{-} \\
ZoeDepth\textsuperscript{\dag}~\cite{bhat2023zoedepth} & $\scnd{93.6}$ & $\scnd{8.2}$ & $\scnd{3.24}$ & $96.7$ & \textcolor{gray}{-} & \textcolor{gray}{-} \\
UniDepth~\cite{piccinelli2024unidepth} & $\best{98.0}$ & $\best{4.8}$ & $\best{2.14}$ & $\best{98.3}$ & $\best{85.8}$ & $\best{97.5}$ \\
MASt3R~\cite{leroy2024master} & $2.8$ & $58.2$ & $11.88$ & $90.9$ & $10.9$ & $77.7$ \\
DepthPro~\cite{bochkovskii2024depthpro} & $78.2$ & $17.2$ & $5.27$ & $94.8$ & $62.4$ & $80.9$ \\
\midrule
\ourmodel-Small & $92.1$ & $11.6$ & $3.76$ & $96.4$ & $\scnd{77.7}$ & $\scnd{85.6}$ \\
\ourmodel-Base & $93.1$ & $12.6$ & $3.84$ & $\scnd{97.3}$ & $76.6$ & $82.7$ \\
\ourmodel-Large & $81.2$ & $17.4$ & $4.77$ & $96.8$ & $71.4$ & $79.3$ \\
\bottomrule
\end{tabular}%
}
\end{table}
\begin{table}[ht]
\centering
\caption{\textbf{Comparison on zero-shot evaluation for IBims-1.} Missing values (-) indicate the model's inability to produce the respective output. \dag: ground-truth camera for 3D reconstruction. \ddag: ground-truth camera for 2D depth map inference.}
\label{tab:results:atomic_ibims}
\resizebox{\linewidth}{!}{%
\begin{tabular}{l|cccccc}
\toprule
\textbf{Method}  & $\mathrm{\delta_1}\uparrow$ & $\mathrm{A.Rel}\downarrow$ & $\mathrm{RMSE}\downarrow$ & $\mathrm{\delta_1^{SSI}}\uparrow$ & $\mathrm{F_A}\uparrow$ & $\mathrm{\rho_A}\uparrow$ \\
\midrule
DepthAnything~\cite{yang2024da1} & \textcolor{gray}{-} & \textcolor{gray}{-} & \textcolor{gray}{-} & $97.0$ & \textcolor{gray}{-} & \textcolor{gray}{-} \\
DepthAnythingv2~\cite{yang2024da2} & \textcolor{gray}{-} & \textcolor{gray}{-} & \textcolor{gray}{-} & $98.0$ & \textcolor{gray}{-} & \textcolor{gray}{-} \\
Metric3D\textsuperscript{\dag \ddag}~\cite{yin2023metric3d} & $75.1$ & $19.3$ & $0.633$ & $96.2$ & \textcolor{gray}{-} & \textcolor{gray}{-} \\
Metric3Dv2\textsuperscript{\dag \ddag}~\cite{hu2024metric3dv2} & $68.4$ & $20.7$ & $0.700$ & $\best{98.8}$ & \textcolor{gray}{-} & \textcolor{gray}{-} \\
ZoeDepth\textsuperscript{\dag}~\cite{bhat2023zoedepth} & $49.8$ & $21.5$ & $0.989$ & $95.8$ & \textcolor{gray}{-} & \textcolor{gray}{-} \\
UniDepth~\cite{piccinelli2024unidepth} & $15.7$ & $41.0$ & $1.25$ & $98.1$ & $30.3$ & $\best{76.6}$ \\
MASt3R~\cite{leroy2024master} & $61.0$ & $19.7$ & $0.883$ & $95.1$ & $55.7$ & $\scnd{76.0}$ \\
DepthPro~\cite{bochkovskii2024depthpro} & $82.3$ & $17.0$ & $0.573$ & $98.0$ & $62.8$ & $75.9$ \\
\midrule
\ourmodel-Small & $\scnd{87.7}$ & $13.0$ & $0.484$ & $97.7$ & $67.3$ & $74.6$ \\
\ourmodel-Base & $87.6$ & $\scnd{12.5}$ & $\scnd{0.452}$ & $98.0$ & $\scnd{67.5}$ & $73.4$ \\
\ourmodel-Large & $\best{91.9}$ & $\best{10.4}$ & $\best{0.406}$ & $\scnd{98.5}$ & $\best{69.8}$ & $75.4$ \\
\bottomrule
\end{tabular}%
}
\end{table}
\begin{table}[ht]
\centering
\caption{\textbf{Comparison on zero-shot evaluation for ETH3D.} Missing values (-) indicate the model's inability to produce the respective output. \dag: ground-truth camera for 3D reconstruction. \ddag: ground-truth camera for 2D depth map inference.}
\label{tab:results:atomic_eth3d}
\resizebox{\linewidth}{!}{%
\begin{tabular}{l|cccccc}
\toprule
\textbf{Method}  & $\mathrm{\delta_1}\uparrow$ & $\mathrm{A.Rel}\downarrow$ & $\mathrm{RMSE}\downarrow$ & $\mathrm{\delta_1^{SSI}}\uparrow$ & $\mathrm{F_A}\uparrow$ & $\mathrm{\rho_A}\uparrow$ \\
\midrule
DepthAnything~\cite{yang2024da1} & \textcolor{gray}{-} & \textcolor{gray}{-} & \textcolor{gray}{-} & $93.2$ & \textcolor{gray}{-} & \textcolor{gray}{-} \\
DepthAnythingv2~\cite{yang2024da2} & \textcolor{gray}{-} & \textcolor{gray}{-} & \textcolor{gray}{-} & $93.3$ & \textcolor{gray}{-} & \textcolor{gray}{-} \\
Metric3D\textsuperscript{\dag \ddag}~\cite{yin2023metric3d} & $19.7$ & $136.8$ & $10.45$ & $81.1$ & \textcolor{gray}{-} & \textcolor{gray}{-} \\
Metric3Dv2\textsuperscript{\dag \ddag}~\cite{hu2024metric3dv2} & $\best{90.0}$ & $\best{12.7}$ & $\best{1.85}$ & $89.7$ & \textcolor{gray}{-} & \textcolor{gray}{-} \\
ZoeDepth\textsuperscript{\dag}~\cite{bhat2023zoedepth} & $33.8$ & $54.7$ & $3.45$ & $86.1$ & \textcolor{gray}{-} & \textcolor{gray}{-} \\
UniDepth~\cite{piccinelli2024unidepth} & $18.5$ & $53.3$ & $3.50$ & $93.9$ & $27.6$ & $42.6$ \\
MASt3R~\cite{leroy2024master} & $21.4$ & $45.3$ & $4.43$ & $91.3$ & $28.4$ & $\best{92.2}$ \\
DepthPro~\cite{bochkovskii2024depthpro} & $39.7$ & $65.2$ & $36.31$ & $81.1$ & $41.2$ & $77.4$ \\
\midrule
\ourmodel-Small & $53.6$ & $60.0$ & $4.89$ & $94.2$ & $44.3$ & $80.7$ \\
\ourmodel-Base & $68.4$ & $28.5$ & $3.77$ & $\scnd{95.8}$ & $\best{53.8}$ & $\scnd{82.0}$ \\
\ourmodel-Large & $\scnd{68.7}$ & $\scnd{23.6}$ & $\scnd{2.63}$ & $\best{95.9}$ & $\scnd{53.6}$ & $81.3$ \\
\bottomrule
\end{tabular}%
}
\end{table}
\begin{table}[ht]
\centering
\caption{\textbf{Comparison on zero-shot evaluation for Diode (Indoor).} Missing values (-) indicate the model's inability to produce the respective output. \dag: ground-truth camera for 3D reconstruction. \ddag: ground-truth camera for 2D depth map inference.}
\label{tab:results:atomic_diodeindoor}
\resizebox{\linewidth}{!}{%
\begin{tabular}{l|cccccc}
\toprule
\textbf{Method}  & $\mathrm{\delta_1}\uparrow$ & $\mathrm{A.Rel}\downarrow$ & $\mathrm{RMSE}\downarrow$ & $\mathrm{\delta_1^{SSI}}\uparrow$ & $\mathrm{F_A}\uparrow$ & $\mathrm{\rho_A}\uparrow$ \\
\midrule
DepthAnything~\cite{yang2024da1} & \textcolor{gray}{-} & \textcolor{gray}{-} & \textcolor{gray}{-} & $97.5$ & \textcolor{gray}{-} & \textcolor{gray}{-} \\
DepthAnythingv2~\cite{yang2024da2} & \textcolor{gray}{-} & \textcolor{gray}{-} & \textcolor{gray}{-} & $97.6$ & \textcolor{gray}{-} & \textcolor{gray}{-} \\
Metric3D\textsuperscript{\dag \ddag}~\cite{yin2023metric3d} & $40.4$ & $61.1$ & $2.34$ & $91.3$ & \textcolor{gray}{-} & \textcolor{gray}{-} \\
Metric3Dv2\textsuperscript{\dag \ddag}~\cite{hu2024metric3dv2} & $\best{94.0}$ & $\best{9.3}$ & $\best{0.399}$ & $\best{98.5}$ & \textcolor{gray}{-} & \textcolor{gray}{-} \\
ZoeDepth\textsuperscript{\dag}~\cite{bhat2023zoedepth} & $34.9$ & $33.6$ & $2.07$ & $91.8$ & \textcolor{gray}{-} & \textcolor{gray}{-} \\
UniDepth~\cite{piccinelli2024unidepth} & $\scnd{76.2}$ & $17.2$ & $0.954$ & $97.2$ & $\best{63.0}$ & $\best{96.1}$ \\
MASt3R~\cite{leroy2024master} & $52.6$ & $27.9$ & $1.68$ & $92.3$ & $48.8$ & $70.2$ \\
DepthPro~\cite{bochkovskii2024depthpro} & $67.1$ & $19.9$ & $0.900$ & $93.9$ & $50.3$ & $71.5$ \\
\midrule
\ourmodel-Small & $57.2$ & $21.4$ & $0.968$ & $96.1$ & $49.3$ & $\scnd{92.5}$ \\
\ourmodel-Base & $55.1$ & $19.6$ & $0.859$ & $97.4$ & $50.1$ & $91.2$ \\
\ourmodel-Large & $71.3$ & $\scnd{16.1}$ & $\scnd{0.767}$ & $\scnd{97.9}$ & $\scnd{53.8}$ & $79.5$ \\
\bottomrule
\end{tabular}%
}
\end{table}
\begin{table}[ht]
\centering
\caption{\textbf{Comparison on zero-shot evaluation for nuScenes.} Missing values (-) indicate the model's inability to produce the respective output. \dag: ground-truth camera for 3D reconstruction. \ddag: ground-truth camera for 2D depth map inference.}
\label{tab:results:atomic_nuscenes}
\resizebox{\linewidth}{!}{%
\begin{tabular}{l|cccccc}
\toprule
\textbf{Method}  & $\mathrm{\delta_1}\uparrow$ & $\mathrm{A.Rel}\downarrow$ & $\mathrm{RMSE}\downarrow$ & $\mathrm{\delta_1^{SSI}}\uparrow$ & $\mathrm{F_A}\uparrow$ & $\mathrm{\rho_A}\uparrow$ \\
\midrule
DepthAnything~\cite{yang2024da1} & \textcolor{gray}{-} & \textcolor{gray}{-} & \textcolor{gray}{-} & $79.0$ & \textcolor{gray}{-} & \textcolor{gray}{-} \\
DepthAnythingv2~\cite{yang2024da2} & \textcolor{gray}{-} & \textcolor{gray}{-} & \textcolor{gray}{-} & $79.4$ & \textcolor{gray}{-} & \textcolor{gray}{-} \\
Metric3D\textsuperscript{\dag \ddag}~\cite{yin2023metric3d} & $75.4$ & $23.7$ & $8.94$ & $64.0$ & \textcolor{gray}{-} & \textcolor{gray}{-} \\
Metric3Dv2\textsuperscript{\dag \ddag}~\cite{hu2024metric3dv2} & $84.1$ & $23.6$ & $9.40$ & $64.8$ & \textcolor{gray}{-} & \textcolor{gray}{-} \\
ZoeDepth\textsuperscript{\dag}~\cite{bhat2023zoedepth} & $33.8$ & $42.0$ & $\scnd{7.41}$ & $64.8$ & \textcolor{gray}{-} & \textcolor{gray}{-} \\
UniDepth~\cite{piccinelli2024unidepth} & $\scnd{84.6}$ & $\best{12.7}$ & $\best{4.56}$ & $83.1$ & $\scnd{64.4}$ & $\scnd{97.7}$ \\
MASt3R~\cite{leroy2024master} & $2.7$ & $65.6$ & $13.76$ & $63.5$ & $13.6$ & $78.3$ \\
DepthPro~\cite{bochkovskii2024depthpro} & $56.6$ & $28.7$ & $11.29$ & $59.1$ & $46.5$ & $79.1$ \\
\midrule
\ourmodel-Small & $80.9$ & $18.9$ & $8.43$ & $83.8$ & $59.4$ & $95.8$ \\
\ourmodel-Base & $\best{84.9}$ & $\scnd{16.7}$ & $9.15$ & $\scnd{86.7}$ & $\best{65.5}$ & $\best{97.8}$ \\
\ourmodel-Large & $84.0$ & $18.9$ & $10.83$ & $\best{87.0}$ & $60.3$ & $86.9$ \\
\bottomrule
\end{tabular}%
}
\end{table}
\begin{table}[ht]
\centering
\caption{\textbf{Comparison on zero-shot evaluation for IBims-1\textsubscript{Dist}.} Missing values (-) indicate the model's inability to produce the respective output. \dag: ground-truth camera for 3D reconstruction. \ddag: ground-truth camera for 2D depth map inference.}
\label{tab:results:atomic_ibims_f}
\resizebox{\linewidth}{!}{%
\begin{tabular}{l|cccccc}
\toprule
\textbf{Method}  & $\mathrm{\delta_1}\uparrow$ & $\mathrm{A.Rel}\downarrow$ & $\mathrm{RMSE}\downarrow$ & $\mathrm{\delta_1^{SSI}}\uparrow$ & $\mathrm{F_A}\uparrow$ & $\mathrm{\rho_A}\uparrow$ \\
\midrule
DepthAnything~\cite{yang2024da1} & \textcolor{gray}{-} & \textcolor{gray}{-} & \textcolor{gray}{-} & $97.1$ & \textcolor{gray}{-} & \textcolor{gray}{-} \\
DepthAnythingv2~\cite{yang2024da2} & \textcolor{gray}{-} & \textcolor{gray}{-} & \textcolor{gray}{-} & $93.4$ & \textcolor{gray}{-} & \textcolor{gray}{-} \\
Metric3D\textsuperscript{\dag \ddag}~\cite{yin2023metric3d} & $56.8$ & $26.5$ & $0.947$ & $93.3$ & \textcolor{gray}{-} & \textcolor{gray}{-} \\
Metric3Dv2\textsuperscript{\dag \ddag}~\cite{hu2024metric3dv2} & $61.3$ & $22.1$ & $0.940$ & $93.3$ & \textcolor{gray}{-} & \textcolor{gray}{-} \\
ZoeDepth\textsuperscript{\dag}~\cite{bhat2023zoedepth} & $30.0$ & $28.0$ & $1.28$ & $94.5$ & \textcolor{gray}{-} & \textcolor{gray}{-} \\
UniDepth~\cite{piccinelli2024unidepth} & $48.7$ & $23.0$ & $0.966$ & $97.2$ & $53.3$ & $69.3$ \\
MASt3R~\cite{leroy2024master} & $31.8$ & $31.9$ & $1.30$ & $92.8$ & $44.1$ & $69.7$ \\
DepthPro~\cite{bochkovskii2024depthpro} & $27.2$ & $47.6$ & $1.86$ & $83.0$ & $32.4$ & $69.5$ \\
\midrule
\ourmodel-Small & $\scnd{67.2}$ & $\scnd{17.1}$ & $0.726$ & $97.6$ & $\scnd{62.6}$ & $71.5$ \\
\ourmodel-Base & $66.0$ & $17.9$ & $\scnd{0.695}$ & $\scnd{98.3}$ & $59.8$ & $\scnd{72.7}$ \\
\ourmodel-Large & $\best{70.9}$ & $\best{15.0}$ & $\best{0.615}$ & $\best{98.6}$ & $\best{67.9}$ & $\best{77.3}$ \\
\bottomrule
\end{tabular}%
}
\end{table}
\begin{table}[ht]
\centering
\caption{\textbf{Comparison on zero-shot evaluation for ETH3D\textsubscript{Dist}.} Missing values (-) indicate the model's inability to produce the respective output. \dag: ground-truth camera for 3D reconstruction. \ddag: ground-truth camera for 2D depth map inference.}
\label{tab:results:atomic_eth3d_f}
\resizebox{\linewidth}{!}{%
\begin{tabular}{l|cccccc}
\toprule
\textbf{Method}  & $\mathrm{\delta_1}\uparrow$ & $\mathrm{A.Rel}\downarrow$ & $\mathrm{RMSE}\downarrow$ & $\mathrm{\delta_1^{SSI}}\uparrow$ & $\mathrm{F_A}\uparrow$ & $\mathrm{\rho_A}\uparrow$ \\
\midrule
DepthAnything~\cite{yang2024da1} & \textcolor{gray}{-} & \textcolor{gray}{-} & \textcolor{gray}{-} & $91.8$ & \textcolor{gray}{-} & \textcolor{gray}{-} \\
DepthAnythingv2~\cite{yang2024da2} & \textcolor{gray}{-} & \textcolor{gray}{-} & \textcolor{gray}{-} & $83.9$ & \textcolor{gray}{-} & \textcolor{gray}{-} \\
Metric3D\textsuperscript{\dag \ddag}~\cite{yin2023metric3d} & $19.6$ & $123.6$ & $11.05$ & $80.9$ & \textcolor{gray}{-} & \textcolor{gray}{-} \\
Metric3Dv2\textsuperscript{\dag \ddag}~\cite{hu2024metric3dv2} & $42.8$ & $104.3$ & $9.87$ & $83.5$ & \textcolor{gray}{-} & \textcolor{gray}{-} \\
ZoeDepth\textsuperscript{\dag}~\cite{bhat2023zoedepth} & $25.4$ & $45.9$ & $4.12$ & $86.1$ & \textcolor{gray}{-} & \textcolor{gray}{-} \\
UniDepth~\cite{piccinelli2024unidepth} & $27.6$ & $43.8$ & $4.69$ & $90.1$ & $38.5$ & $67.5$ \\
MASt3R~\cite{leroy2024master} & $14.6$ & $51.8$ & $5.37$ & $87.7$ & $32.0$ & $\scnd{78.5}$ \\
DepthPro~\cite{bochkovskii2024depthpro} & $16.1$ & $72.8$ & $18.77$ & $72.7$ & $29.1$ & $69.9$ \\
\midrule
\ourmodel-Small & $42.1$ & $125.3$ & $12.14$ & $92.9$ & $49.9$ & $68.4$ \\
\ourmodel-Base & $\scnd{47.9}$ & $\scnd{36.5}$ & $\scnd{3.54}$ & $\scnd{95.1}$ & $\scnd{53.5}$ & $67.1$ \\
\ourmodel-Large & $\best{67.0}$ & $\best{22.1}$ & $\best{2.75}$ & $\best{95.5}$ & $\best{63.6}$ & $\best{83.1}$ \\
\bottomrule
\end{tabular}%
}
\end{table}
\begin{table}[ht]
\centering
\caption{\textbf{Comparison on zero-shot evaluation for Diode\textsubscript{Dist} (Indoor).} Missing values (-) indicate the model's inability to produce the respective output. \dag: ground-truth camera for 3D reconstruction. \ddag: ground-truth camera for 2D depth map inference.}
\label{tab:results:atomic_diodeindoor_f}
\resizebox{\linewidth}{!}{%
\begin{tabular}{l|cccccc}
\toprule
\textbf{Method}  & $\mathrm{\delta_1}\uparrow$ & $\mathrm{A.Rel}\downarrow$ & $\mathrm{RMSE}\downarrow$ & $\mathrm{\delta_1^{SSI}}\uparrow$ & $\mathrm{F_A}\uparrow$ & $\mathrm{\rho_A}\uparrow$ \\
\midrule
DepthAnything~\cite{yang2024da1} & \textcolor{gray}{-} & \textcolor{gray}{-} & \textcolor{gray}{-} & $94.2$ & \textcolor{gray}{-} & \textcolor{gray}{-} \\
DepthAnythingv2~\cite{yang2024da2} & \textcolor{gray}{-} & \textcolor{gray}{-} & \textcolor{gray}{-} & $89.3$ & \textcolor{gray}{-} & \textcolor{gray}{-} \\
Metric3D\textsuperscript{\dag \ddag}~\cite{yin2023metric3d} & $26.4$ & $124.0$ & $4.08$ & $89.7$ & \textcolor{gray}{-} & \textcolor{gray}{-} \\
Metric3Dv2\textsuperscript{\dag \ddag}~\cite{hu2024metric3dv2} & $\best{34.1}$ & $35.2$ & $1.61$ & $91.6$ & \textcolor{gray}{-} & \textcolor{gray}{-} \\
ZoeDepth\textsuperscript{\dag}~\cite{bhat2023zoedepth} & $24.0$ & $39.8$ & $2.32$ & $90.1$ & \textcolor{gray}{-} & \textcolor{gray}{-} \\
UniDepth~\cite{piccinelli2024unidepth} & $30.2$ & $34.8$ & $1.85$ & $94.7$ & $\best{37.2}$ & $74.8$ \\
MASt3R~\cite{leroy2024master} & $20.6$ & $46.0$ & $2.41$ & $89.3$ & $29.5$ & $83.0$ \\
DepthPro~\cite{bochkovskii2024depthpro} & $24.7$ & $56.5$ & $2.31$ & $86.0$ & $26.5$ & $75.7$ \\
\midrule
\ourmodel-Small & $27.6$ & $33.4$ & $1.48$ & $95.0$ & $33.0$ & $82.6$ \\
\ourmodel-Base & $\scnd{31.6}$ & $\scnd{30.0}$ & $\scnd{1.35}$ & $\scnd{96.1}$ & $\scnd{37.0}$ & $\scnd{85.1}$ \\
\ourmodel-Large & $26.9$ & $\best{30.0}$ & $\best{1.33}$ & $\best{97.5}$ & $36.1$ & $\best{85.4}$ \\
\bottomrule
\end{tabular}%
}
\end{table}
\begin{table}[ht]
\centering
\caption{\textbf{Comparison on zero-shot evaluation for ScanNet++ (DSLR).} Missing values (-) indicate the model's inability to produce the respective output. \dag: ground-truth camera for 3D reconstruction. \ddag: ground-truth camera for 2D depth map inference.}
\label{tab:results:atomic_scannetpp_f}
\resizebox{\linewidth}{!}{%
\begin{tabular}{l|cccccc}
\toprule
\textbf{Method}  & $\mathrm{\delta_1}\uparrow$ & $\mathrm{A.Rel}\downarrow$ & $\mathrm{RMSE}\downarrow$ & $\mathrm{\delta_1^{SSI}}\uparrow$ & $\mathrm{F_A}\uparrow$ & $\mathrm{\rho_A}\uparrow$ \\
\midrule
DepthAnything~\cite{yang2024da1} & \textcolor{gray}{-} & \textcolor{gray}{-} & \textcolor{gray}{-} & $51.4$ & \textcolor{gray}{-} & \textcolor{gray}{-} \\
DepthAnythingv2~\cite{yang2024da2} & \textcolor{gray}{-} & \textcolor{gray}{-} & \textcolor{gray}{-} & $52.3$ & \textcolor{gray}{-} & \textcolor{gray}{-} \\
Metric3D\textsuperscript{\dag \ddag}~\cite{yin2023metric3d} & $16.5$ & $180.5$ & $1.83$ & $51.2$ & \textcolor{gray}{-} & \textcolor{gray}{-} \\
Metric3Dv2\textsuperscript{\dag \ddag}~\cite{hu2024metric3dv2} & $5.2$ & $237.0$ & $2.51$ & $71.3$ & \textcolor{gray}{-} & \textcolor{gray}{-} \\
ZoeDepth\textsuperscript{\dag}~\cite{bhat2023zoedepth} & $2.0$ & $158.5$ & $1.45$ & $71.2$ & \textcolor{gray}{-} & \textcolor{gray}{-} \\
UniDepth~\cite{piccinelli2024unidepth} & $0.6$ & $162.9$ & $1.59$ & $71.0$ & $9.1$ & $20.2$ \\
MASt3R~\cite{leroy2024master} & $5.8$ & $114.8$ & $1.07$ & $73.0$ & $21.0$ & $16.6$ \\
DepthPro~\cite{bochkovskii2024depthpro} & $9.6$ & $95.8$ & $0.928$ & $74.1$ & $24.4$ & $30.9$ \\
\midrule
\ourmodel-Small & $6.2$ & $92.8$ & $0.931$ & $78.1$ & $23.5$ & $35.1$ \\
\ourmodel-Base & $\scnd{55.4}$ & $\scnd{33.1}$ & $\scnd{0.340}$ & $\scnd{86.6}$ & $\scnd{53.9}$ & $\scnd{65.1}$ \\
\ourmodel-Large & $\best{65.1}$ & $\best{25.3}$ & $\best{0.285}$ & $\best{90.8}$ & $\best{59.1}$ & $\best{70.0}$ \\
\bottomrule
\end{tabular}%
}
\end{table}
\begin{table}[ht]
\centering
\caption{\textbf{Comparison on zero-shot evaluation for ADT.} Missing values (-) indicate the model's inability to produce the respective output. \dag: ground-truth camera for 3D reconstruction. \ddag: ground-truth camera for 2D depth map inference.}
\label{tab:results:atomic_adt}
\resizebox{\linewidth}{!}{%
\begin{tabular}{l|cccccc}
\toprule
\textbf{Method}  & $\mathrm{\delta_1}\uparrow$ & $\mathrm{A.Rel}\downarrow$ & $\mathrm{RMSE}\downarrow$ & $\mathrm{\delta_1^{SSI}}\uparrow$ & $\mathrm{F_A}\uparrow$ & $\mathrm{\rho_A}\uparrow$ \\
\midrule
DepthAnything~\cite{yang2024da1} & \textcolor{gray}{-} & \textcolor{gray}{-} & \textcolor{gray}{-} & $81.7$ & \textcolor{gray}{-} & \textcolor{gray}{-} \\
DepthAnythingv2~\cite{yang2024da2} & \textcolor{gray}{-} & \textcolor{gray}{-} & \textcolor{gray}{-} & $82.6$ & \textcolor{gray}{-} & \textcolor{gray}{-} \\
Metric3D\textsuperscript{\dag \ddag}~\cite{yin2023metric3d} & $72.5$ & $26.2$ & $0.560$ & $85.3$ & \textcolor{gray}{-} & \textcolor{gray}{-} \\
Metric3Dv2\textsuperscript{\dag \ddag}~\cite{hu2024metric3dv2} & $75.6$ & $21.9$ & $0.433$ & $92.4$ & \textcolor{gray}{-} & \textcolor{gray}{-} \\
ZoeDepth\textsuperscript{\dag}~\cite{bhat2023zoedepth} & $11.0$ & $81.4$ & $1.36$ & $83.5$ & \textcolor{gray}{-} & \textcolor{gray}{-} \\
UniDepth~\cite{piccinelli2024unidepth} & $13.3$ & $76.0$ & $1.37$ & $90.8$ & $27.1$ & $32.1$ \\
MASt3R~\cite{leroy2024master} & $44.8$ & $40.1$ & $0.717$ & $86.7$ & $52.5$ & $51.4$ \\
DepthPro~\cite{bochkovskii2024depthpro} & $33.6$ & $45.1$ & $0.902$ & $81.3$ & $47.9$ & $48.0$ \\
\midrule
\ourmodel-Small & $89.8$ & $13.4$ & $0.323$ & $93.8$ & $82.9$ & $92.2$ \\
\ourmodel-Base & $\scnd{93.5}$ & $\scnd{10.3}$ & $\scnd{0.288}$ & $\scnd{95.0}$ & $\scnd{88.1}$ & $\best{93.8}$ \\
\ourmodel-Large & $\best{94.6}$ & $\best{9.3}$ & $\best{0.275}$ & $\best{95.6}$ & $\best{89.5}$ & $\scnd{93.7}$ \\
\bottomrule
\end{tabular}%
}
\end{table}
\begin{table}[ht]
\centering
\caption{\textbf{Comparison on zero-shot evaluation for KITTI360.} Missing values (-) indicate the model's inability to produce the respective output. \dag: ground-truth camera for 3D reconstruction. \ddag: ground-truth camera for 2D depth map inference.}
\label{tab:results:atomic_kitti360}
\resizebox{\linewidth}{!}{%
\begin{tabular}{l|cccccc}
\toprule
\textbf{Method}  & $\mathrm{\delta_1}\uparrow$ & $\mathrm{A.Rel}\downarrow$ & $\mathrm{RMSE}\downarrow$ & $\mathrm{\delta_1^{SSI}}\uparrow$ & $\mathrm{F_A}\uparrow$ & $\mathrm{\rho_A}\uparrow$ \\
\midrule
DepthAnything~\cite{yang2024da1} & \textcolor{gray}{-} & \textcolor{gray}{-} & \textcolor{gray}{-} & $9.5$ & \textcolor{gray}{-} & \textcolor{gray}{-} \\
DepthAnythingv2~\cite{yang2024da2} & \textcolor{gray}{-} & \textcolor{gray}{-} & \textcolor{gray}{-} & $11.3$ & \textcolor{gray}{-} & \textcolor{gray}{-} \\
Metric3D\textsuperscript{\dag \ddag}~\cite{yin2023metric3d} & $0.2$ & $1366.2$ & $34.78$ & $39.7$ & \textcolor{gray}{-} & \textcolor{gray}{-} \\
Metric3Dv2\textsuperscript{\dag \ddag}~\cite{hu2024metric3dv2} & $0.1$ & $1655.3$ & $40.32$ & $43.9$ & \textcolor{gray}{-} & \textcolor{gray}{-} \\
ZoeDepth\textsuperscript{\dag}~\cite{bhat2023zoedepth} & $0.7$ & $1200.2$ & $24.71$ & $41.2$ & \textcolor{gray}{-} & \textcolor{gray}{-} \\
UniDepth~\cite{piccinelli2024unidepth} & $29.4$ & $152.2$ & $4.23$ & $44.0$ & $14.6$ & $7.1$ \\
MASt3R~\cite{leroy2024master} & $16.5$ & $312.8$ & $7.17$ & $41.7$ & $15.7$ & $7.4$ \\
DepthPro~\cite{bochkovskii2024depthpro} & $5.5$ & $103.8$ & $7.35$ & $38.0$ & $5.9$ & $17.5$ \\
\midrule
\ourmodel-Small & $\scnd{74.9}$ & $39.8$ & $\scnd{2.58}$ & $\scnd{81.6}$ & $59.5$ & $\best{82.8}$ \\
\ourmodel-Base & $73.3$ & $\scnd{33.8}$ & $2.62$ & $80.8$ & $\scnd{61.2}$ & $80.9$ \\
\ourmodel-Large & $\best{81.7}$ & $\best{24.4}$ & $\best{2.40}$ & $\best{85.3}$ & $\best{66.4}$ & $\scnd{82.5}$ \\
\bottomrule
\end{tabular}%
}
\end{table}

\subsection{Per-dataset Evaluation}
\label{sec:supp:quant:atomic}

We present results for each of the validation datasets independently in \Cref{tab:results:atomic_nyuv2depth} (NYUv2), \Cref{tab:results:atomic_kitti} (KITTI), \Cref{tab:results:atomic_ibims} (IBims-1), \Cref{tab:results:atomic_eth3d} (ETH3D), \Cref{tab:results:atomic_diodeindoor} (Diode Indoor), \Cref{tab:results:atomic_nuscenes} (nuScenes), \Cref{tab:results:atomic_ibims_f} (IBims-1\textsubscript{Dist}), \Cref{tab:results:atomic_eth3d_f} (ETH3D\textsubscript{Dist}),  \Cref{tab:results:atomic_diodeindoor_f} (Diode Indoor\textsubscript{Dist}), \Cref{tab:results:atomic_scannetpp_f} (ScanNet++ DSLR), \Cref{tab:results:atomic_adt} (ADT), and \Cref{tab:results:atomic_kitti360} (KITTI360). 
Note that we do not report results for the ``Pano'' group, as it only consists of a single dataset, Stanford-2D3D.
Our results show that performance on pinhole camera models has reached a saturation point, yet \ourmodel achieves the highest average metric overall, even though it does not always rank first on every individual dataset.
This demonstrates the strong generalization ability of \ourmodel, attributed to its flexible design and large-scale training, which enables robust performance across diverse domains without overfitting to any specific one.
We report additional and more typical metrics such as absolute relative error as $\mathrm{A.Rel}$ as a percentage and root-means-squared error $\mathrm{RSME}$ using meter as unit.

\begin{table}[t]
\centering
\caption{\textbf{Comparison with UniDepth.}
All models use ViT-S backbone and the same training data. Test set grouping as in the main paper. Best viewed on a screen and zoomed in.
}
\vspace{-10pt}
\label{tab:results:unidepth_comparison}
\resizebox{1\linewidth}{!}{%
    \begin{tabular}{l|ccc|ccc|ccc|ccc}
    \toprule
    \multirow{2}{*}{\textbf{Method}} & \multicolumn{3}{c|}{Small FoV} & \multicolumn{3}{c|}{Small FoV\textsubscript{Dist}} & \multicolumn{3}{c|}{Large FoV} & \multicolumn{3}{c}{Panoramic} \\
     & $\mathrm{\delta_1^{SSI}}\uparrow$ & $\mathrm{F_A}\uparrow$ & $\mathrm{\rho_A}\uparrow$ & $\mathrm{\delta_1^{SSI}}\uparrow$ & $\mathrm{F_A}\uparrow$ & $\mathrm{\rho_A}\uparrow$ & $\mathrm{\delta_1^{SSI}}\uparrow$ & $\mathrm{F_A}\uparrow$ & $\mathrm{\rho_A}\uparrow$ & $\mathrm{\delta_1^{SSI}}\uparrow$ & $\mathrm{F_A}\uparrow$ & $\mathrm{\rho_A}\uparrow$ \\
    \midrule
    UniDepth~\cite{piccinelli2024unidepth} & $89.0$ & $54.7$ & $77.8$ & $92.7$ & $35.4$ & $45.6$ & $71.8$ & $41.9$ & $48.8$ & $34.9$ & $1.5$ & $1.2$ \\
    \ourmodel & $\best{89.1}$ & $\best{57.3}$ & $\best{79.8}$ & $\best{93.1}$ & $\best{44.6}$ & $\best{59.3}$ & $\best{79.8}$ & $\best{53.5}$ & $\best{64.8}$ & $\best{64.3}$ & $\best{58.6}$ & $\best{26.3}$ \\
    \bottomrule
    \end{tabular}%
}
\end{table}

\section{Q\&A}
\label{sec:supp:faq}

Here we list possible questions that might arise after reading the paper.
We structure the section in a discursive question-and-answer fashion.

\begin{itemize}
    \item \textbf{What is the importance of data for generalization \wrt scene scale?} \\
    Data diversity is crucial for generalizing depth estimation, especially for monocular methods that heavily rely on semantic cues and are sensitive to domain gaps.
    Scale prediction in monocular metric depth estimation is inherently ill-posed, making it highly dependent on the training domain and its distribution coverage.
    Excessive diversity can hurt performance in narrow, specialized domains like KITTI, where models trained on large, diverse datasets often underperform compared to those trained on domain-specific data.
    Conversely, these models perform better in broader domains like NYU.
    Scale prediction is typically noisy and sensitive to domain shifts, but this issue can be mitigated through in-domain fine-tuning.
    For example, a few hundred optimization steps can largely resolve the ``scale gap'' when fine-tuning on KITTI. 
    \item \textbf{The camera representation is superior to pinhole or fully non-parametric camera model, but you did not compare it to some common camera models, why so?} \\
    We initially experimented with explicit parametric camera models but encountered significant drawbacks.
    Most standard camera models rely on backprojection operations which are not differentiable and, thus cannot be used in a standard deep learning pipeline.
    Addressing this limitation requires either (i) using differentiable parametric models, such as EUCM~\cite{Khomutenko2016EUCM} or DoubleSphere~\cite{Usenko2018DS}, (ii) approximating polynomial inversions with differentiable functions, or (iii) supervising only the model parameters without direct camera supervision.
    All these approaches suffer from the inherent instability of parametric models, where parameter variations need to be considered jointly on their actual output, namely the pencil of rays.
    This compounding effect, where small compounded changes lead to large output variations, often leads to unstable optimization.
    Furthermore, parametric models limit the expressiveness of the backprojection operation and constrain applicability to only those cameras the model can represent.
    In contrast, our representation avoids these limitations and provides greater flexibility and stability.
    \item \textbf{DUSt3R / MASt3R architecture directly predicts point maps, are they unable to work with generic cameras?} \\
    While DUSt3R and MASt3R networks can theoretically represent any camera model, our studies revealed that fully non-parametric approaches struggle when trained on diverse datasets and tested on edge cases or distribution tails.
    Additionally, the test-time point cloud global alignment technique used in DUSt3R~\cite{wang2024duster} and MASt3R~\cite{leroy2024master} explicitly requires a pinhole camera, further limiting their applicability to generic cameras.
    \item \textbf{What is the role of the confidence prediction?} \\
    Confidence prediction is included primarily for its utility in downstream tasks and also for legacy reasons.
    It is worth noting that, like most regression tasks, confidence prediction is vulnerable to domain gaps, which can render it unreliable in strong out-of-domain scenarios.
    \item \textbf{What is the rationale of camera augmentations?} \\
    Camera augmentations were employed to address the lack of diverse real-camera data.
    While our simple augmentation pipeline resulted in minor improvements, we observed that many generated cameras are unrealistic and fall outside the distribution of real-world cameras.
    However, softmax-based warping proved effective in creating realistic images.
    We hypothesize that a more sophisticated camera sampling procedure, considering the realism of the output rays instead of the singled-out parameters, could significantly enhance the robustness and generalization across real and practical camera models.
    \item \textbf{What are the differences with UniDepth?} \\
    UniDepth~\cite{piccinelli2024unidepth} and UniK3D differ in camera modeling and 3D representation, both ablated in Tabs. 3, 4, and 5.
    UniDepth relies on the \emph{pinhole model} by predicting the calibration matrix (cf.\ [60, Sec.~3.2]), thus not being able to predict \emph{any} camera.
    In addition,  \cite{piccinelli2024unidepth} represents the 3\textsuperscript{rd} dimension as \emph{depth} ($z$) [60, Sec.~3.1].
    These two aspects force \cite{piccinelli2024unidepth} to model \emph{only} pinhole and to output FoV $<180^{\circ}$.
    In contrast, UniK3D uses spherical harmonics (SH) to approximate \emph{any} camera model and it exploits \emph{radial distance} ($r$) as 3\textsuperscript{rd} dimension.
    UniDepth projects the predicted \emph{pinhole ray map} [60, Sec.~3.1] onto a high-dimensional space $\mathbf{E}$ using SH, whereas UniK3D directly \emph{predicts the SH coefficient} used to generate the ray map $\mathbf{C}$ via inverse transform (L230-262).
    This key methodological difference leads to modeling any camera.
    \Cref{tab:results:unidepth_comparison} (row 1 \vs row 3) shows its impact, as UniK3D consistently outperforms \cite{piccinelli2024unidepth} also when trained on identical data.
    \item \textbf{Has someone done something similar before?} \\ Yes, there are a few works~\cite{Kumar2020UnRectDepth, Hirose2021Depth360} which tried to remove the pinhole assumption for depth estimation. However, they are different for two important reasons: (i) those works focused on single-domain scenarios, leading to a simpler setting and (ii) the task is self-supervised depth estimation, where the camera is needed to define the warping-based photometric loss, inherently needing the camera, rather than supervised large-scale monocular 3D estimation.
    \item \textbf{}
    We provide here the $\mathrm{\delta_1^{SSI}}$ scores of row~3 and 4 of Tab.~5: 92.1 and 92.2, respectively.
    This score similarity, along with $\mathrm{F_A}$ and $\mathrm{\rho_A}$ drops (Tab.~5), spotlights angular module's role.
    In fact, radial- and SH-based model (row 4) overestimates FoV of images with lens distortions.
    Retraining with stronger distortion augmentation for small FoV leads to $(\mathrm{F_A}, \mathrm{\rho_{\mathrm{A}}}) = (43.1, 62.3)$, validating our assumption.
\end{itemize}

\section{Additional Qualitative Results}
\label{sec:supp:qual}

\begin{figure*}[h!]
    \renewcommand{\arraystretch}{1}
    \centering
    \small
    \begin{tabular}{cc|cccc}
        \multirow{1}{*}[2.9cm]{\rotatebox[origin=c]{90}{\parbox{2.7cm}{\centering ScanNet++ (DSLR)\\(Kannala-Brandt)}}}
        & \includegraphics[width=0.165\linewidth]{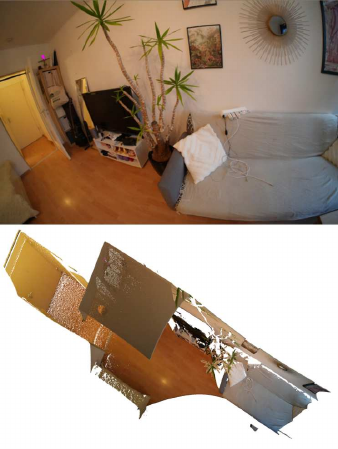}
        & \includegraphics[width=0.165\linewidth]{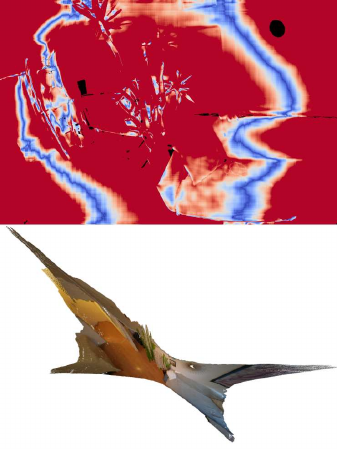}
        & \includegraphics[width=0.165\linewidth]{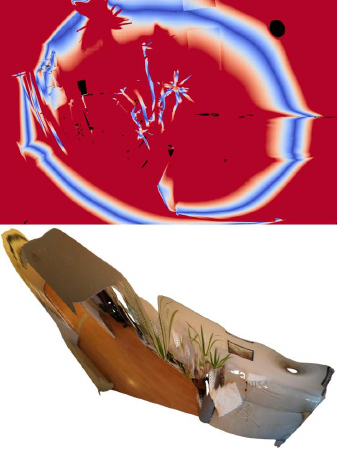}
        & \includegraphics[width=0.165\linewidth]{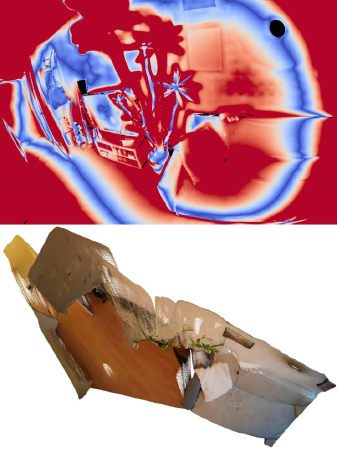}
        & \includegraphics[width=0.165\linewidth]{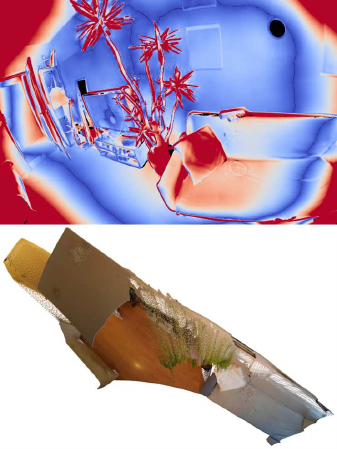}\\
        
        \multirow{1}{*}[3.4cm]{\rotatebox[origin=c]{90}{\parbox{2cm}{\centering IBims-1\textsubscript{Dist}\\(Fisheye)} }}
        & \includegraphics[width=0.165\linewidth]{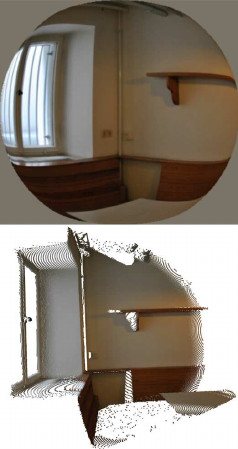}
        & \includegraphics[width=0.165\linewidth]{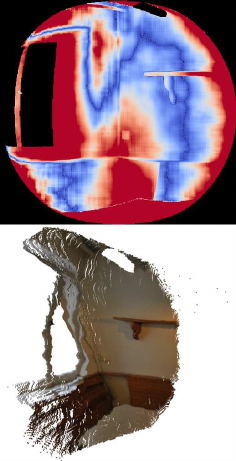}
        & \includegraphics[width=0.165\linewidth]{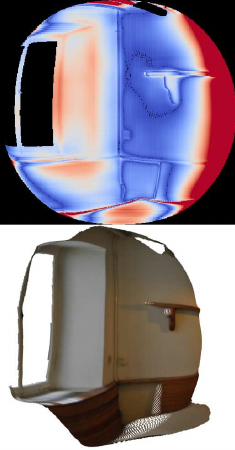}
        & \includegraphics[width=0.165\linewidth]{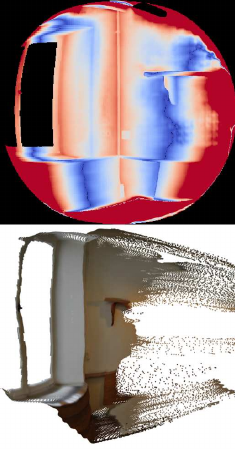}
        & \includegraphics[width=0.165\linewidth]{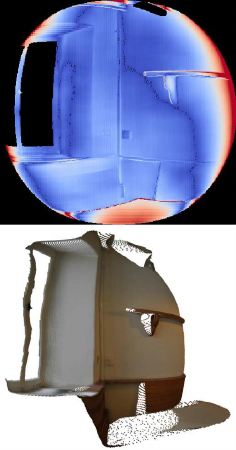}\\
        
        \multirow{1}{*}[3.2cm]{\rotatebox[origin=c]{90}{\parbox{3cm}{\centering ETH3D\textsubscript{Dist}\\(Fisheye)} }}
        & \includegraphics[width=0.165\linewidth]{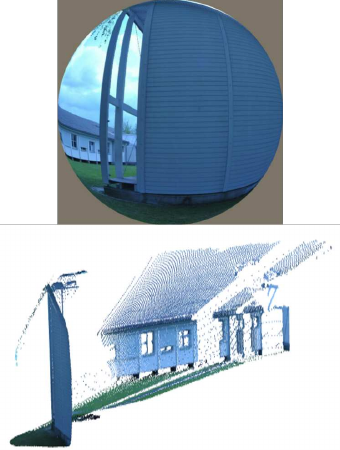}
        & \includegraphics[width=0.165\linewidth]{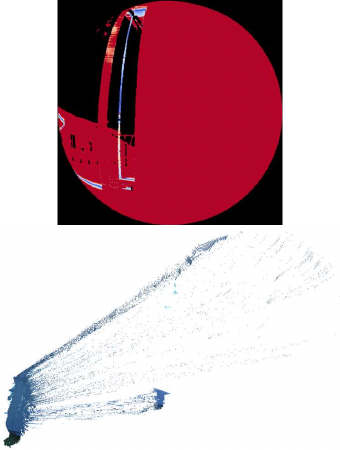}
        & \includegraphics[width=0.165\linewidth]{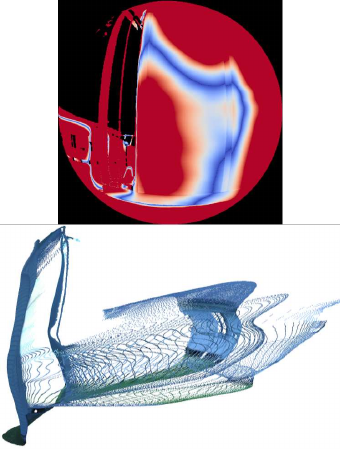}
        & \includegraphics[width=0.165\linewidth]{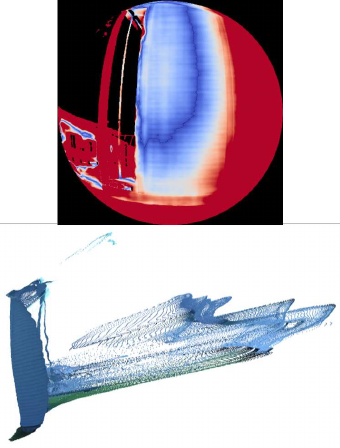}
        & \includegraphics[width=0.165\linewidth]{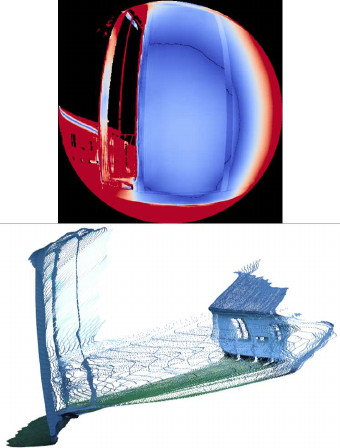}\\

        \multirow{1}{*}[3.5cm]{\rotatebox[origin=c]{90}{\parbox{2.4cm}{\centering Diode\textsubscript{Dist} (Indoor)\\(Fisheye)}}}
        & \includegraphics[width=0.165\linewidth]{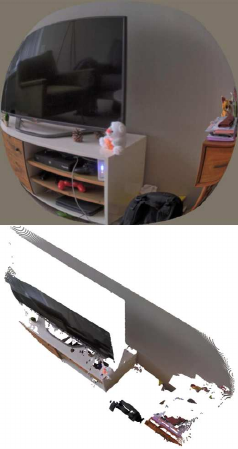}
        & \includegraphics[width=0.165\linewidth]{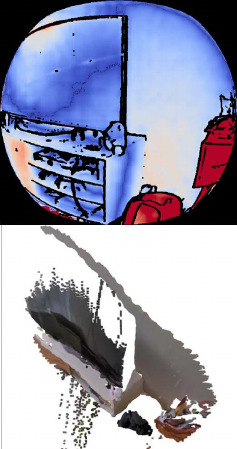}
        & \includegraphics[width=0.165\linewidth]{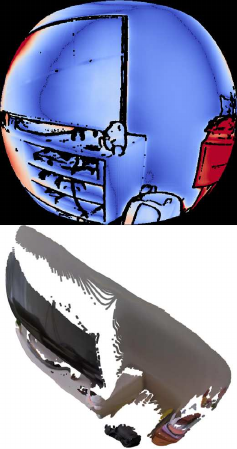}
        & \includegraphics[width=0.165\linewidth]{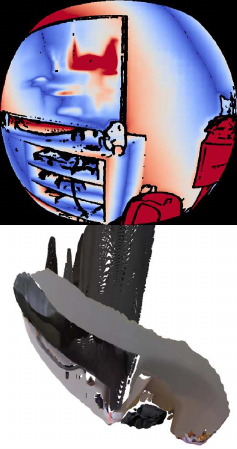}
        & \includegraphics[width=0.165\linewidth]{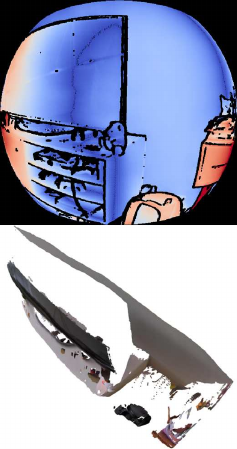}\\

        & RGB \& GT & Metric3Dv2\textsuperscript{\dag}~\cite{hu2024metric3dv2} & UniDepth~\cite{piccinelli2024unidepth} & MASt3R~\cite{leroy2024master} & \ourmodel  \\
    \end{tabular}
    \vspace{-5pt}
    \caption{\textbf{Qualitative comparisons.} Each pair of consecutive rows represents one test sample. Each odd row displays the input RGB image and the 2D error map, color-coded with the \textit{coolwarm} colormap based on absolute relative error with blue corresponding to 0\% error and red to 25\%.
    To ensure a fair comparison, errors are calculated on GT-based shifted and scaled outputs for all models.
    Each even row shows the ground truth and predictions of the 3D point cloud. All samples are randomly selected and not picked. \dag: GT-camera unprojection.}
    \vspace{-5pt}
    \label{fig:supp:main_vis}
\end{figure*}

\begin{figure*}[h!]
    \centering
    \footnotesize
    \includegraphics[width=0.98\linewidth]{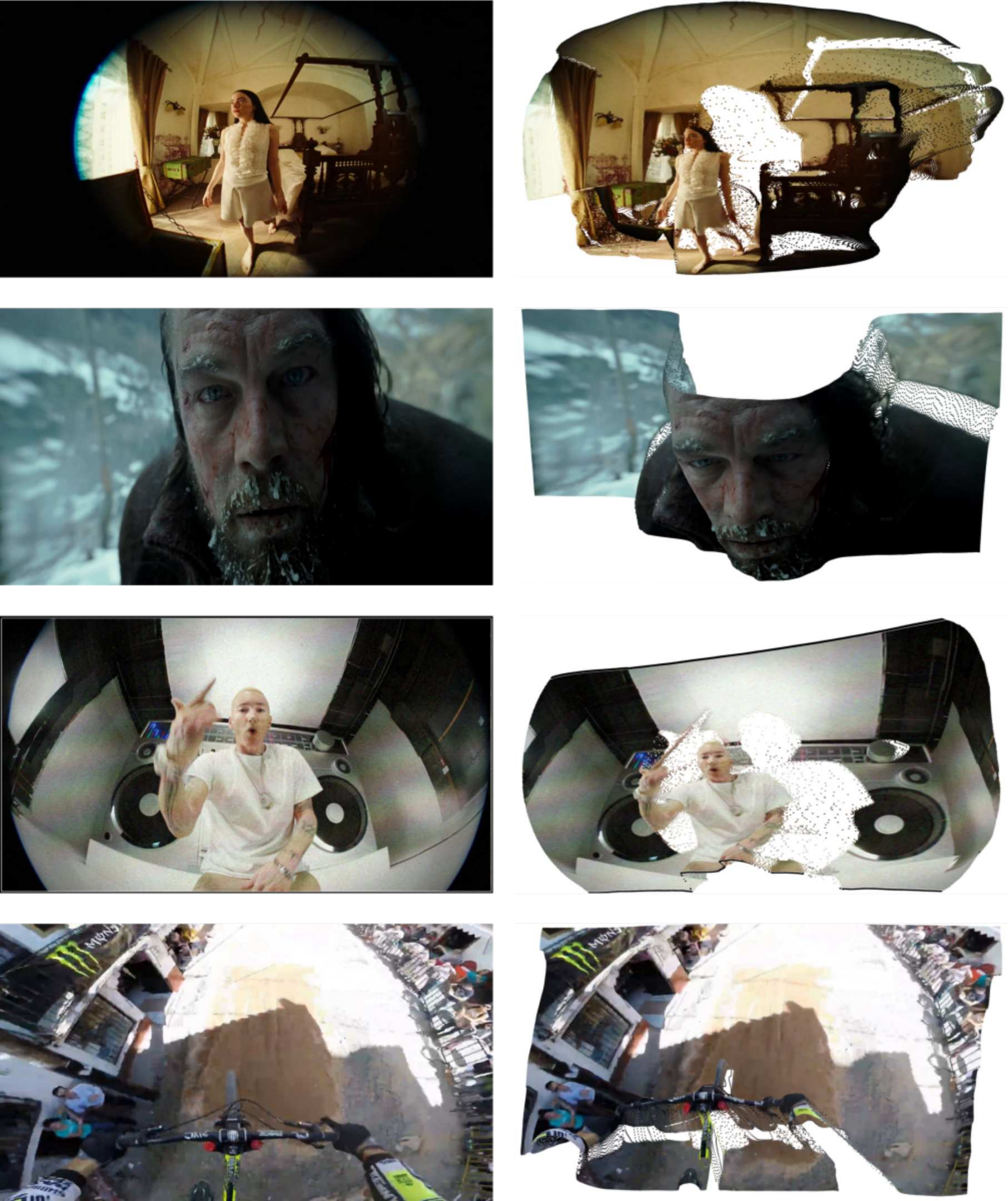}
    \vspace{-5pt}
    \caption{\textbf{Qualitative in-the-wild 3D results.}\ourmodel is fed solely each single image in the left column and it outputs the corresponding point cloud in the right column, the point of view is slightly tilted to better appreciate the 3D. The images are video frames respectively from Poor Things (movie), The Revenant (movie), Eminem (music video), and YouTube (egocentric GoPro). The frames present a variety of camera types and unusual viewpoints.}
    \label{fig:supp:quali1}
\end{figure*}

\begin{figure*}[h!]
    \centering
    \footnotesize
    \includegraphics[width=0.98\linewidth]{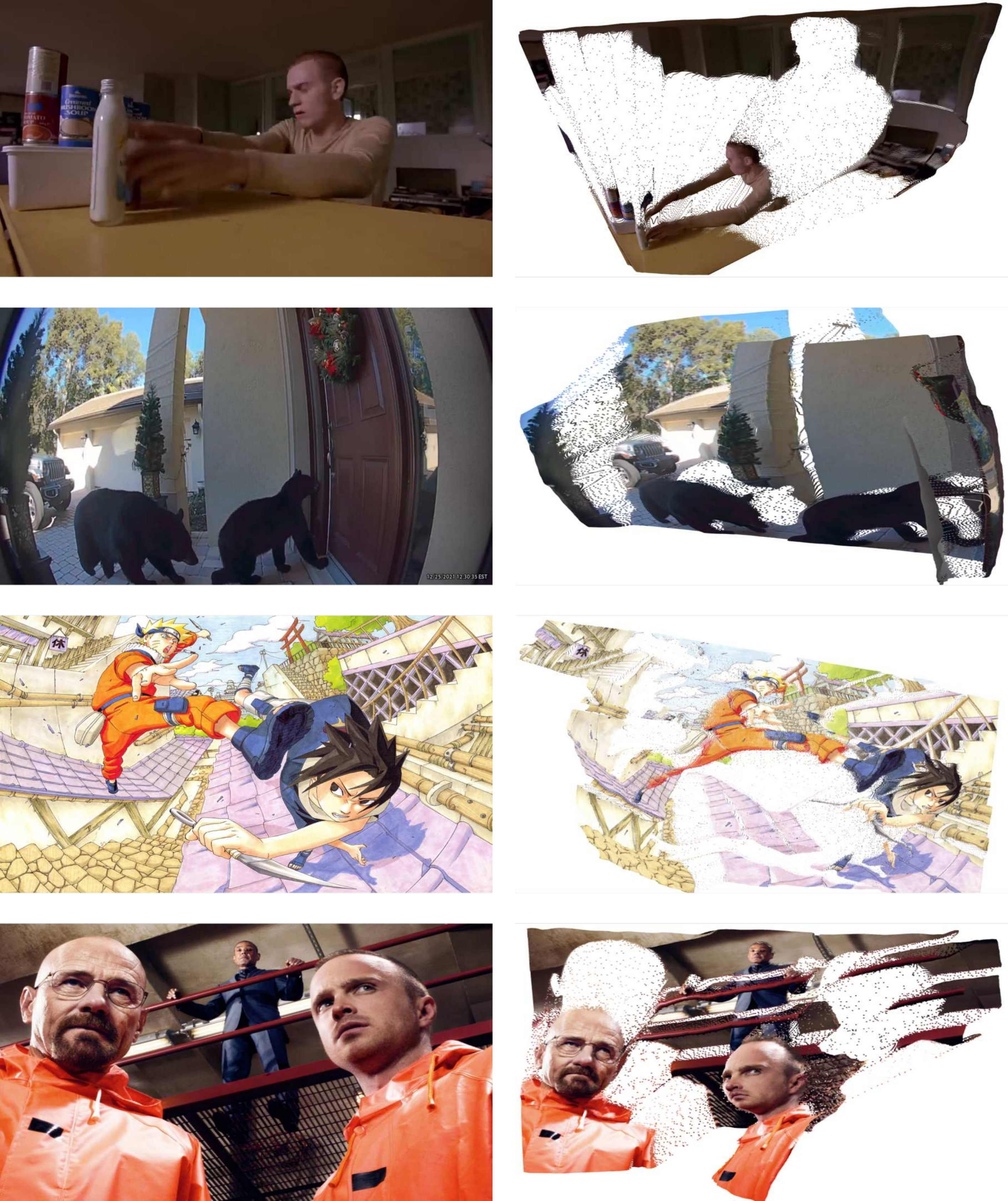}
    \vspace{-5pt}
    \caption{\textbf{Qualitative in-the-wild 3D results.} \ourmodel is fed solely each single image in the left column and it outputs the corresponding point cloud in the right column, the point of view is slightly tilted to better appreciate the 3D. The images are video frames respectively from Trainspotting (movie), YouTube (doorbell camera), Naruto (anime), and Breaking Bad (TV series). The frames present a variety of camera types and unusual viewpoints.}
    \label{fig:supp:quali2}
\end{figure*}

We provide here more qualitative comparisons, in particular from validation domains not presented in the main paper and with distorted cameras, namely ScanNet++ (DSLR), IBims-1\textsubscript{Dist}, and Diode\textsubscript{Dist} (Indoor), in \cref{fig:supp:main_vis}.
In addition, we test our model on complete in-the-wild scenarios, for instance, frames from movies, TV series, YouTube, or animes.
All images depicted in \cref{fig:supp:quali1} and \cref{fig:supp:quali2} present deformed cameras or unusual points of view.
The visualization here presented, both from the validation sets and the in-the-wild ones are casually selected and not cherry-picked.

\end{document}